\documentclass{article}
    \usepackage[nonatbib,preprint]{neurips_data_2023}
    \usepackage[square,numbers]{natbib}
\usepackage{listings}
\usepackage{graphicx}
\usepackage[utf8]{inputenc} % allow utf-8 input
\usepackage[T1]{fontenc}    % use 8-bit T1 fonts
\usepackage{hyperref}       % hyperlinks
\usepackage{url}            % simple URL typesetting
\usepackage{booktabs}       % professional-quality tables
\usepackage{amsfonts}       % blackboard math symbols
\usepackage{nicefrac}       % compact symbols for 1/2, etc.
\usepackage{microtype}      % microtypography
\usepackage{xcolor}         % colors
\usepackage{enumitem}
\usepackage{xspace}
\usepackage{graphicx}
\usepackage{multicol}
\usepackage{multirow}
\usepackage{wrapfig}
\usepackage{dblfloatfix}
\usepackage{subcaption}
\usepackage{devanagari}
\usepackage{upquote}
\usepackage[utf8]{inputenc}
\usepackage{scalerel}
\usepackage{longtable}

\usepackage{multirow}
\usepackage{multicol}
\usepackage{makecell}

\usepackage{setspace}
\usepackage{changepage} % to adjust margins
\usepackage{float} % for tables inside tcolorbox https://tex.stackexchange.com/a/274342
\usepackage{enumitem}

\usepackage{tabularx}

\usepackage{mdframed}

\hypersetup{
    colorlinks=true,
    linkcolor=blue,
    filecolor=magenta,      
    urlcolor=cyan,
}

\newcommand{\data}{{\fontfamily{qcr}\selectfont MADLAD-400}\xspace}

\title{\data: A Multilingual And Document-Level Large Audited Dataset}

\author{%
Sneha Kudugunta$^\dagger$ \quad Isaac Caswell$^\diamond$ \quad Biao Zhang$^\dagger$ \quad Xavier Garcia$^\dagger$\\ \textbf{Christopher A. Choquette-Choo}$^\dagger$ \quad \textbf{Katherine Lee}$^\dagger$ \quad \textbf{Derrick Xin}$^\dagger$ \quad \textbf{Aditya Kusupati}$^\diamond$ \\ \textbf{Romi Stella}$^\dagger$ \quad \textbf{Ankur Bapna}$^\dagger$ \quad \textbf{Orhan Firat}$^\dagger$\\
$^\dagger$Google DeepMind \quad
$^\diamond$Google Research
}

\begin{document}

% \doparttoc % Tell to minitoc to generate a toc for the parts
% \faketableofcontents % Run a fake tableofcontents command for the partocs

\maketitle
\begin{abstract}

We introduce \data, a manually audited, general domain 3T token monolingual dataset based on CommonCrawl, spanning 419 languages. We discuss the limitations revealed by self-auditing \data, and the role data auditing had in the dataset creation process. We then train and release a 10.7B-parameter multilingual machine translation model on 250 billion tokens covering over 450 languages using publicly available data, and find that it is competitive with models that are significantly larger, and report the results on different domains. In addition, we train a 8B-parameter language model, and assess the results on few-shot translation. We make the baseline models \footnote{\url{https://github.com/google-research/google-research/tree/master/madlad_400}} available to the research community. 

\end{abstract}
\section{Introduction}
\label{sec:intro}

The availability of large multilingual corpora  has accelerated the progress of multilingual natural language processing (NLP) models~\citep{xue-etal-2021-mt5,conneau2019unsupervised,lin2021few,bapna-etal-2022-building,nllb2022}. However, most publicly available general-domain multilingual corpora contain 100-200 languages~\citep{xue-etal-2021-mt5,nllb2022,oscar}, with some datasets containing more languages in specific domains such as religious content~\cite{agic2019jw300}, children's books~\citep{leong-etal-2022-bloom} or dialects~\citep{adebara2022serengeti}.

A common approach to creating such datasets is to mine language specific data from general web crawls such as CommonCrawl~\citep{raffel2020exploring,laurenccon2022bigscience,xue2020mt5} to create datasets.  We simply take this approach and scale it. We train a document-level LangID model on 498 languages to obtain CommonCrawl annotations at a document level and obtain a 5-trillion token, document-level monolingual dataset.

However, such web-scale corpora are known to be noisy and contain undesirable content~\citep{paullada2021data,luccioni2021s,dodge2021documenting}, with their multilingual partitions often having their own specific issues such as unusable text, misaligned and mislabeled/ambiguously labeled data ~\citep{kreutzer-etal-2022-quality}. To mitigate this, we manually audit our data. Based on our findings, we discard 79 of the languages from our preliminary dataset, rename or combine several languages and apply additional preprocessing steps. Finally, to validate the efficacy of our dataset, we train multilingual machine translation models of various sizes up to 10.7B parameters, as well as an 8B decoder-only model, and then evaluate these models on highly multilingual translation evaluation sets.

In Section~\ref{sec:data}, we describe the creation and composition of \data, and discuss the results of the audit. Then, in Section~\ref{sec:parallel}, we describe the parallel data we collect using publicly available sources to train the multilingual machine translation models described in Section~\ref{sec:mt-model}. In Section~\ref{sec:exp}, we describe the training process of the multilingual machine translation models and 8B decoder-only model, and then evaluate these models on highly multilingual translation datasets. In Section~\ref{sec:memorization} we describe our tests for memorization in the multilingual models that we release and discuss preliminary results. Finally, we discuss the limitations of this work and directions for future work.
\begin{figure}[t!]
    \centering
    \includegraphics[width=\linewidth]{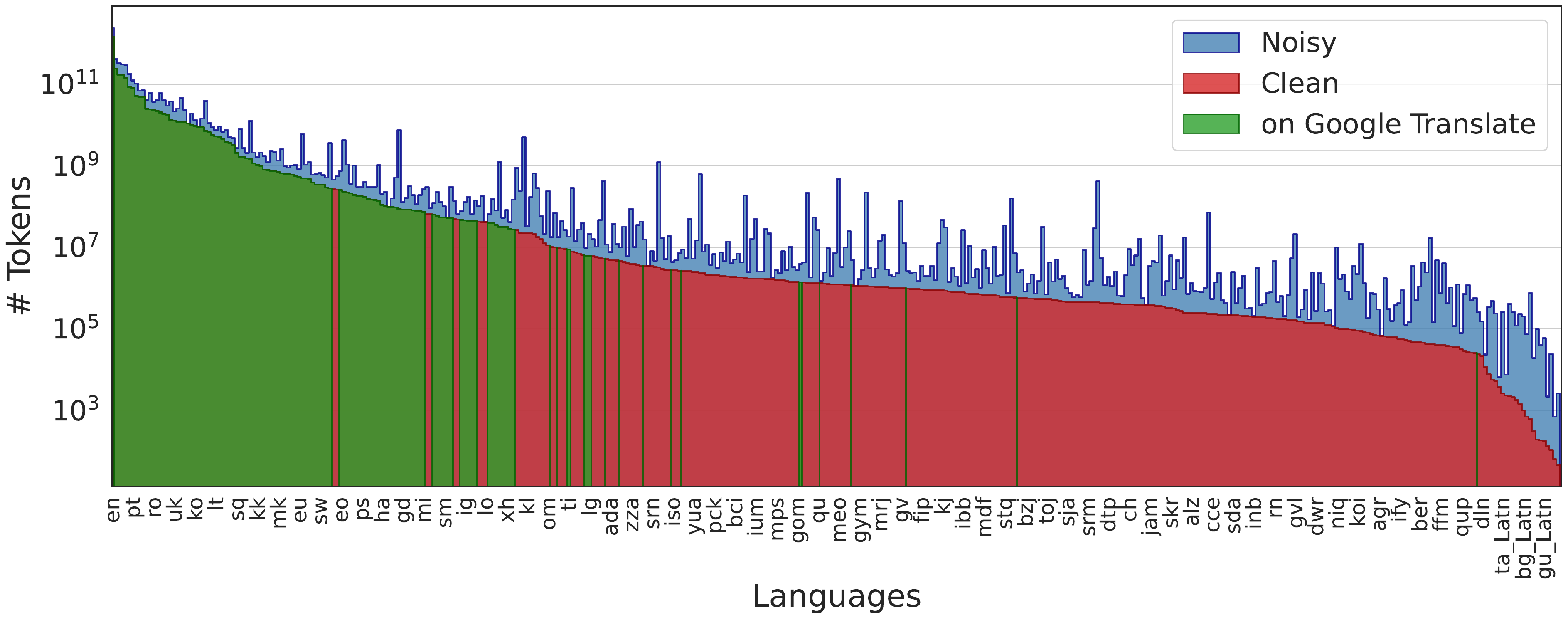}
    \caption{\textbf{Comparing the size of the noisy and clean monolingual datasets in \data}. The difference is more noticeable on lower-resource languages, where noise effects are especially severe. For reference, languages supported by Google Translate are shaded in green. Note that, since this chart is in log scale, the difference in size is much greater than it may appear; for instance, for the lower-resource half of the dataset, the ratio is about 4$\times$ on median.}
    \label{fig:madlad-chart}
\end{figure}
\section{\data}\label{sec:data}
\label{sec:madlad}

The process we follow to create \data is similar to that of other large-scale web corpora \cite{caswell2020language,xue2020mt5,oscar,nllb2022}. First, we collect as large a dataset of unlabeled web text as possible. More specifically, we use all available snapshots of CommonCrawl\footnote{https://commoncrawl.org/} as of August 20, 2022. After some preliminary data cleaning, we use a highly multilingual LangID model to provide document-level annotations (Section \ref{sub:langid}). Finally, we conduct a self-audit (Section \ref{sub:audit}), or quality review, of this preliminary dataset partitioned by language, and design filters to remove noisy content. When appropriate, we correct language names and remove languages from the preliminary dataset. We note that building \data was an iterative process, and that while we describe one major quality review in depth, we conducted several stages of filtering. To reflect this, we describe the preprocessing steps and improvements made in chronological order.

\setlength{\textfloatsep}{5pt}
\begin{wrapfigure}{r}{0.4\columnwidth}\vspace{-4mm}
 \centering
 \captionof{table}{\small{Geographic distribution of languages in \data.}}
\begin{tabular}{lc}
\toprule
{Continent} & {\# Languages} \\ \midrule
{Asia} & {149} \\
{Americas} & {66} \\
{Africa} & {87} \\
{Europe} & {89} \\
{Oceania} & {26} \\
{Constructed} & {2} \\ \bottomrule
\end{tabular}%
\label{tab:distr} 
  % \vspace{-3mm}
\end{wrapfigure}

We release two version of this dataset: a 5 trillion token \texttt{noisy} dataset, which is the dataset obtained before applying document-level LangID and the final filters, and a 3 trillion token  \texttt{clean} dataset, which has a variety of filters applied based on our self-audit, though it naturally has a fair amount of noise itself. Each dataset is released in both a document-level form and a sentence-level form. Some overall statistics for these dataset versions are given in Table \ref{tab:madlad-stats}, with a graph visualizing the distribution of sizes (number of tokens) across languages  in Figure~\ref{fig:madlad-chart}. The final version of \data has 419 languages, with a varied geographic distribution, as seen in Table \ref{tab:distr}.

% % Please add the following required packages to your document preamble:
% % \usepackage{graphicx}
% % \usepackage[table,xcdraw]{xcolor}
% % If you use beamer only pass "xcolor=table" option, i.e. \documentclass[xcolor=table]{beamer}
% \begin{table}[]
% \centering
% \resizebox{\textwidth}{!}{%
% \begin{tabular}{l|cr|cr|cr}
% \multicolumn{1}{c|}{\textbf{Dataset Version}} & \multicolumn{2}{c|}{\textbf{N(Documents)}} & \multicolumn{2}{c|}{\textbf{N(Sentences)}} & \multicolumn{2}{c}{\textbf{N(Tokens)}} \\
% \multicolumn{1}{c|}{\textbf{}} & \textbf{Total} & \multicolumn{1}{c|}{\textbf{Median}} & \textbf{Total} & \multicolumn{1}{c|}{\textbf{Median}} & \textbf{Total} & \multicolumn{1}{c}{\textbf{Median}} \\ \hline
% \data-\texttt{noisy} & \multicolumn{1}{r}{7.8B} & 27K & \multicolumn{1}{r}{150B} & 235K & \multicolumn{1}{r}{5.0T} & 7.1M \\
% \data-\texttt{clean} & \multicolumn{1}{r}{4.0B} & 1.7K & \multicolumn{1}{r}{105B} & 73K & \multicolumn{1}{r}{{\color[HTML]{1F1F1F} 2.8T}} & 1.2M
% \end{tabular}%
% }
% \caption{\label{tab:madlad-stats} Overall statistics of both the \texttt{noisy} and \texttt{clean} partitions of \data.}
% \end{table}

% \usepackage{booktabs}
% \usepackage{multirow}
\begin{table}[h!]
\centering
\caption{\label{tab:madlad-stats} Overall statistics of both the \texttt{noisy} and \texttt{clean} partitions of \data.}
\resizebox{0.8\columnwidth}{!}{
\begin{tabular}{@{}lcccccc@{}}
\toprule
\multirow{2}{*}{Dataset Version} & \multicolumn{2}{c}{\# Documents} & \multicolumn{2}{c}{\# Sentences} & \multicolumn{2}{c}{\# Tokens} \\ \cmidrule(l){2-7} 
                                 & Total           & Median          & Total           & Median          & Total         & Median        \\ \midrule
\data-\texttt{noisy}                            & 7.8B            & 27K            & 150B            & 240K           & 5.0T         & 7.1M          \\
\data-\texttt{clean}                            & 4.0B            & 1.7K           & 100B            & 73K          & 2.8T         & 1.2M         \\ \bottomrule
\end{tabular}
}
\end{table}
\subsection{Preliminary Filters}

We carry out a few preliminary preprocessing steps on the web-crawled corpus: first, we deduplicate lines across documents \citep{lee2021deduplicating}. Then, we filter out all pages that do not contain at least 3 lines of 200 or more characters (as done by \citet{xue2020mt5}). We also use other commonly used filtering heuristics such as removing lines containing the word ``Javascript'' and removing pages that contain ``lorem ipsum'' and curly brackets ``\{'' (as done by \citet{raffel2020exploring}).

\subsection{Language Identification (LangID)}\label{sub:langid}

We train a Semi-Supervised LangID model (SSLID) on 500 languages, following the recipe introduced by \citet{caswell2020language}. We then filter the corpus on document-level LangID, which was taken to be the majority sentence-level LangID prediction. The resulting dataset is \data-\texttt{noisy}. For the Additional details on these LangID models is in Appendix~\ref{app:langid}.

\subsection{Filtering Out Questionable Content}

To assess the quality of this preliminary dataset, we inspected 20 sentences each from a subset of 30 languages in our dataset. Based on our observations, we introduced a score, \texttt{pct\_questionable}. The \texttt{pct\_questionable} score is simply the percentage of sentences in the input document that were ``questionable''. A sentence was considered questionable if any of the following were true:

 \begin{enumerate}[nolistsep] %[leftmargin=*]
    \item \textbf{Document consistency:} Sentence-level LangID does not match the document-level LangID.
    \item \textbf{List Case:} Over 50\% percent of the tokens began in a capital letter (we apply this filter only if the sentence has at least 12 tokens.)
    \item \textbf{Abnormal Lengths:} The sentence has under 20 characters or over 500 characters. We note that this is a bad heuristic for ideographic languages\footnote{\url{http://www.grcdi.nl/dqglossary/ideographic\%20language.html}}).
    \item \textbf{Technical Characters:} Over 20\% of the characters in the sentence match  \texttt{[0-9\{\}+/()>]}.
    \item \textbf{Cursed Regexes:} The sentence matched a ``cursed regex''. These are a heuristic set of substrings and regexes that we found accounted for a significant amount of questionable content in the data samples we observed.  They are described in depth in Appendix~\ref{app:filtering}.
\end{enumerate}

We removed all documents with a \texttt{percent\_questionable} score greater than 20\%. Furthermore, we removed any document with under 5 sentences.

\subsection{Self-Audit (Quality Review)}\label{sub:audit}

After filtering out generally lower-quality content with the approach described above, we performed a self-audit of every corpus in this dataset, following ~\citet{kreutzer-etal-2022-quality}. The aim of our self-audit was to correct any remaining systematic issues by either applying additional filters, renaming/merging language codes, or completely removing the language from the dataset. Although we do not speak most of the 498 languages, we were able to give high-level comments on the general quality. For each language, we inspected a sample of 20 documents. This task was evenly divided between the first two authors based in part on which scripts they could read. We used the following guidelines:

\begin{itemize}[nolistsep]
    \item If dataset is mostly plausibly in-language text, we can keep it. For unknown languages, search the web for a few sentences and look at the website and URL for language clues.
    \item If dataset is noisy but the noise looks filterable, leave a note of how to filter it.
    \item If the dataset is very noisy and does not look possible to filter, mark it for removal.
    \item Optionally put note that may be helpful for downstream users, e.g. if dataset is 100\% Bible.
\end{itemize}

We made the decision to include languages that looked noisy, but omit any language that was majority noise, or only had 20 or fewer docs. While this is not a high quality bar, we hope it still has the potential to be useful to the research community, given that foundation models have demonstrated the potential to learn distributions for very few exammples \cite{brown2020language}. The motivation for not releasing ``nonsense'' or tiny datasets is to avoid giving a false sense of how multilingual the dataset is (``Representation washing''), as recommended by \textbf{Quality at a Glance}~\citep{kreutzer-etal-2022-quality}.

\paragraph{Overall Results.} Of the 498 languages that we obtained LangID annotations for, we decided to omit 79 languages, bringing the final number of languages in \data to 419. Based on the self-audit, we also expanded the filters (particularly the cursed regexes), and made changes as described in Sections \ref{sub:add-filters} and \ref{sub:other-issues}. We details stats for these languages in Appendix Section~\ref{app:mono}.

For transparency, we provide full results of the self-audit in Appendix \ref{app:mono}. In Table \ref{tab:audit}, we provide an overview of the issues surfaced through this self-audit. We find that a significant fraction of languages contain mostly or entirely religious documents, while other issues include misrendered text, pornographic content, and boilerplate.

% \usepackage{graphicx}
\begin{table}[h]
\centering
\caption{\label{tab:audit} Summary of results of the audit on the preliminary dataset comprising of 498 languages. Note that there may be multiple issues with data in one language.}
\resizebox{0.6\textwidth}{!}{%
\begin{tabular}{lr}
\toprule
\# Languages... & \multicolumn{1}{l}{} \\ \midrule
Audited & 498 \\
With significant amounts of Bible data & 141 \\
With significant amounts of JW data & 37 \\
With significant amounts of LDS data & 2 \\
With significant amounts of virama-based issues & 8 \\
With a significant number of short docs & 42 \\
With complaints about noise & 28 \\
With complaints about porn & 10 \\
With complaints about boilerplate & 15 \\
% With a note to ask a native speaker & 6 \\
With a note to remove from the dataset & 77 \\ \bottomrule
\end{tabular}%
}

\end{table}

\subsection{Additional Filters}\label{sub:add-filters}

Based on the results of the self-audit, we apply three additional filters.

\paragraph{Virama Filtering and Correction.}

Many languages using Brahmic Abugida (South and Southeast Asian scripts like Devanagari, Khmer, etc.) use some variant on the virama~\footnote{\url{https://en.wikipedia.org/wiki/Virama}} character. We found that such languages in \data-\texttt{noisy} had incorrectly encoded viramas: for example, \includegraphics[height=0.6cm]{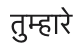} was rendered as \includegraphics[height=0.5cm]{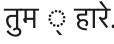}, where the middle character is a detached virama. 
Therefore, for the languages \texttt{bn, my, pa, gu, or, ta, te, kn, ml, si, th, tl, mn, lo, bo, km, hi, mr, ne, gom, as, jv, dv, bho, dz, hne, ks\_Deva, mag, mni, shn, yue, zh, ja, kjg, mnw, ksw, rki, mtr, mwr} and \texttt{xnr}, we did a special filtering/correction step --- we removed all extraneous spaces before virama characters. We provide the pseudocode and list of virama characters in Appendix \ref{app:filtering}.

\paragraph{Zawgyi Encoded Data.} We found that languages using Myanmar script like \texttt{my} and \texttt{mnw} appeared to have the same issues with virama characters that still remained after applying the virama correction. This was because a large fraction of Myanmar script data on the internet is Zawgyi encoded data, which appears to have the rendering issues described above if rendered in Unicode. Therefore, we used an open-source Zawgyi detector \footnote{\url{https://github.com/google/myanmar-tools}} to convert the encoding of documents with more than a 50\% probability of being Zawgyi encoded into standard Unicode encoding.

\paragraph{Chinese-Specific Filters.} The Mandarin (\texttt{zh}) data in CommonCrawl had a particular issue with pornographic content. We combed through the data and developed a list of strings likely to be present in pornographic content, and filtered out all documents containing the strings in the blocklist. This resulted in a 17\% reduction in the number of documents and a 56\% reduction in file size. We list these strings in Appendix \ref{app:filtering}.

\subsection{Correcting Other Systematic Issues.}\label{sub:other-issues}

Based on various specific notes from the self-audit, we made a variety of changes. Five datasets were found to be in the wrong language, and were renamed or merged into the correct dataset. Six languages that looked suspicious were run by native speakers of those or related languages, some of which were discarded, and some of which were merged into the correct dataset. Finally, we removed all languages with fewer than 20 documents. Details can be seen in Appendix \ref{app:other-issues}.

\section{Parallel Data}\label{sec:parallel}

To train the machine translation (MT) models described in Section \ref{sec:mt-model}, we also collect a dataset composed of publicly available datasets coming from various data sources. A full list of the data sources and associated language pairs are in Appendix \ref{app:parallel}. The final dataset has 156 languages across 4.1B sentence pairs and 4124 language pairs total. In the rest of the paper, we refer to the input sentence to an MT model as the ``source side" and the reference/output sentence as the ``target side".

\subsection{Filters}

We describe the data preprocessing steps taken below. We find that a significant amount of data is filtered out, with the amount of data available 396 of 4.1k language pairs reducing by more than $40\%$.

\paragraph{Deduplication.} We deduplicate sentence pairs that are an exact match on both the source and target.

\paragraph{Virama Filtering and Correction/Zawgyi Encoded Data.} We observed the same issues described in Section~\ref{sub:add-filters}, and used the same filters for sentence pairs where either the source language or target language belonged to the list of languages in Section~\ref{sub:add-filters}.

\paragraph{Unmatched Toxicity Filters.}

We use the unmatched toxicity filters described by~\citet{nllb2022}, but ultimately unusable for our purposes in most cases. For the languages \texttt{ace, am, ar, az, bg, bm, bn, bs, cs, din, en, es, fa, fr, ga, gl, ha, hi, id, it, kk, ko, ml, ms, my, nl, no, nus, prs, ru, scn, sd, so, sv, tg, th, tt, ur, uz} and \texttt{zh}, more than 3\% of documents were marked as having unmatched toxicity. On closer inspection, we found that while \texttt{zh} and \texttt{ko} had a lot of pornographic content that was removed by the filtering process, most other languages removed sentences that had homonyms of non-toxic words. Similarly, languages like \texttt{id, ur, tg, fa} and \texttt{no} had data from Tanzil (Qur'an dataset), but the toxicity word lists contained words such as \texttt{kafir}, \texttt{mercy} and \texttt{purity}, that are not normally considered toxic content for our purpose of filtering the dataset using wordlists.

\paragraph{Source-Target Filters.}

We removed all sentences that have more than 75\% overlap between the source and target side. To avoid filtering out valid entity translations, we only applied this filter on sentences longer than 5 tokens. In addition, we remove sentence pairs whose source length to target length ratio falls outside of $0.66-1.5$. We omitted this filter for the following, which are mainly non-whitespace languages: \texttt{zh, ja, ko, km, my, lo, th, wuu, shn, zh\_tw, zh\_cn,iu, simple, dz, kr\_Arab, din, nus} and \texttt{mi}.

\paragraph{Script Filters.}

We removed all sentences that are less than 50\% in-script for both the source and target language. For instance, if the sentence was supposed to be in \texttt{kaa} (Cyrillic script) but was 70\% in the Latin script, we removed it.

\subsection{Self-Audit (Quality Review)}

Similar to the self-audit done for \data, we conducted a review of the data sources that compose the parallel data we collected to verify the quality of this data. We collected 20 source-target pairs from each language, and assessed the data for the presence of offensive content, porn, and whether the data seemed to be of the correct language pair and whether the target sentence seemed to be a plausible translation. Since we did not have access to native speakers of all 157 languages, the latter was primarily based on guesses. In Appendix~\ref{app:parallel} we provide full details of the instructions we provided to auditors, the results of the self-audit and any changes made the dataset.

\subsection{A Note on Language Codes}

As observed by \citet{kreutzer-etal-2022-quality}, the datasets used to create the parallel data (and \data) use a variety of different language codes. We use the BCP-47 standard, which specifies the 2-letter ISO-693-1 code when applicable, and otherwise the ISO-693-3 code. Script tags and region tags are omitted when they are defined as the default value by CLDR \footnote{\url{https://cldr.unicode.org/}}, and otherwise included. For example, \texttt{ks} refers to Kashmiri in Nastaliq/Arabic script (CLDR default), whereas \texttt{ks\_Deva} refers to Kashmiri in Devanagari. A detailed investigation of codes in \data can be found in Appendix \ref{app:other-issues}.

\subsection{Multiway Data}\label{sub:multiway}

We create additional multiway data by applying the $n$-gram matching method ($n=8$) from~\citet{multiway} to the processed dataset. Using this, and the publicly available data, we obtain 11.9B sentences across a total of 20742 language pairs. Full details may be found in  Appendix~\ref{app:multiway}.
\section{Experiments}\label{sec:exp}

We validate our data by training encoder-decoder machine translation models in Section~\ref{sec:mt-model} and decoder-only language models in Section~\ref{sec:lm}, and test them on several translation benchmarks.

\subsection{MT Models}\label{sec:mt-model}

We train models of various sizes: a 3B, 32-layer parameter model,\footnote{Here and elsewhere, `X-layer' means X encoder layers and also X decoder layers, for a total of 2X layers.} a 7.2B 48-layer parameter model and a 10.7B 32-layer parameter model. We share all parameters of the model across language pairs, and use a Sentence Piece Model~\citep{kudo2018sentencepiece} with 256k tokens shared on both the encoder and decoder side. Each input sentence has a \texttt{<2xx>} token prepended to the source sentence to indicate the target language~\citep{johnson2017google}.

% \srkcomment{APPENDIX: optimizer, dimensions, SPM hparams}

 We use both supervised parallel data with a machine translation objective and the monolingual \data dataset with a MASS-style~\citep{song2019mass} objective to train this model. Each of these objectives is sampled with a 50\% probability. Within each task, we use the recently introduced UniMax~\citep{chung2023unimax} sampling strategy to sample languages from our imbalanced dataset with a threshold of $N=10$ epochs for any particular language. We also explored back-translation by randomly sampling 2M monolingual samples (or the total number of samples for that given language) for each language and translating them to/from English using the 3B model. Following~\citet{ bapna-etal-2022-building} (\S 3.5), we filter the back-translated data in a variety of ways. For a natural target and a back-translated source, we filter by round-trip ChrF to discourage hallucinations (threshold of 0.32), by ChrF between source and target to discourage copying (threshold of 0.30), by the length ratio of source to target (asymmetric bounds of (0.45, 1.6), and by LangID prediction of the source. We then finetune the 7.2B model for a $10,000$ steps by randomly mixing the original and the back-translated data with a combining ratio of 1:1. We list specific architecture and training details of these models in Appendix~\ref{app:model}.

\subsection{Zero-shot Translation with Language Models}\label{sec:lm}

Given recent interest in the efficacy of unsupervised translation using large language models, we explore training language models solely on the monolingual data. We follow the same training schedule and model configurations from~\citet{garcia2023unreasonable}. In particular, we consider 8B decoder-only models, following the same model hyperparameters as previous work~\citep{chowdhery2022palm, garcia2023unreasonable}. We train these models using a variant of the UL2 objective~\citep{tay2022unifying} adapted for decoder-only models, and use the same configuration as previous work~\citep{garcia2023unreasonable, orlanski2023measuring}. We provide additional details in Appendix~\ref{app:model}.

\subsection{Evaluation}\label{sub:eval}

We use the sacreBLEU~\citep{post-2018-call} implementation of bleu\footnote{\texttt{ BLEU+case.mixed+lang.<sl>-<tl>+
numrefs.1+smooth.exp+tok.<tok>+version.1.3.0}, \texttt{tok=zh} if \texttt{tl=zh} and \texttt{13a} otherwise.} and chrf\footnote{ \texttt{nrefs:1|case:mixed|eff:yes|nc:6|nw:0|space:no|version:2.3.1}} as metrics. We evaluate our trained models on the following datasets:

\paragraph{WMT.} We use the 15 WMT languages frequently used to evaluate multilingual machine translation models by \citet{siddhant-etal-2022-towards,kim2021scalable,kudugunta-etal-2021-beyond-distillation,nllb2022}: \texttt{cs, de, es, fi, fr, gu, hi, kk, lv, lt, ro, rs, es, tr} and \texttt{zh}.

\paragraph{Flores-200.} We evaluate on the languages in the Flores-200 dataset \cite{nllb2022} that overlap with the languages available in either \data or the parallel data described in Section~\ref{sec:parallel}. We list these languages in Appendix~\ref{app:overlaps}. For non-English-centric pairs, we evaluate on a 272 language pair subset of the 40k language pairs possible due to computational constraints. We evaluate on all language pairs possible using the following languages as either source or target language: \texttt{en, fr, cs, zh, et, mr, eu, cy, so, ckb, or, yo, ny, ti, ln, fon} and \texttt{ss}. We obtained this set of languages by selecting every $10^{th}$ language by number of tokens in MADLAD-400 (clean), starting with French (\texttt{fr}). Noticing that this had no Indian languages, we shifted \texttt{af} and \texttt{fo} (both close dialects of HRLS) down one index to \texttt{mr} and \texttt{or}, respectively. Finally, we noticed that this initial list had supervised and unsupervised languages, but didn't have a good representative of a ``slightly supervised language'', that is, one with a small but extant amount of parallel data. Therefore, we added \texttt{yo} to the list, which has the least parallel data of any supervised language. This resulting subset of languages also contains a nice variety of scripts: Latin, Chinese, Devanagari, Arabic, Odia, and Ethiopic scripts.

\paragraph{NTREX.} We evaluate on the languages in the recently introduced NTREX dataset~\citep{federmann-etal-2022-ntrex}.

\paragraph{Gatones.} Finally, we evaluate on the languages in \textsc{Gatones}, the in-house, 38-language eval set used in~\citep{bapna-etal-2022-building} and the \textsc{Gatitos} paper~\citep{jones2023bilex}. Again, we take the subset of languages overlapping with the languages available in either \data or the parallel training data. 

\begin{table}[t!]
\centering
\caption{\label{tab:wmt} Evaluation scores on WMT (depicted as \texttt{<bleu> / <chrf>}) for the MT models and language models described in Section \ref{sec:mt-model} and Section \ref{sec:lm} compared against NLLB-54B.}
\resizebox{\columnwidth}{!}{
\begin{tabular}{@{}lcccccccc@{}}
\toprule
\multirow{2}{*}{} & \multicolumn{1}{c}{\multirow{2}{*}{\textbf{NLLB}}} & \multicolumn{1}{c}{\multirow{2}{*}{\textbf{MT-3B}}} & \multicolumn{1}{c}{\multirow{2}{*}{\textbf{MT-7.2B}}} & \multicolumn{1}{c}{\multirow{2}{*}{\textbf{MT-10.7B}}} & \multicolumn{4}{c}{\textbf{LM-8B}}                                                                                          \\ \cmidrule(l){6-9} 
                  & \multicolumn{1}{c}{}                      & \multicolumn{1}{c}{}                       & \multicolumn{1}{c}{}                         & \multicolumn{1}{c}{}                        & \multicolumn{1}{c}{0-shot} & \multicolumn{1}{c}{1-shot} & \multicolumn{1}{c}{5-shot} & \multicolumn{1}{c}{10-shot} \\ \midrule
 \textbf{xx2en} & 34.2 / 60.4 & 33.4 / 60.0 & 34.9 / 60.6 & \textbf{34.6 / 60.8} & 2.3 / 17.3 & 25.1 / 51.4 & 26.2 / 52.9 & 26.2 / 53.4 \\
 \textbf{en2xx} & \textbf{31.1 / 58.0} & 28.2 / 55.4 & 29.3 / 56.2 & 29.0 / 56.2 & 1.0 / 10.3 & 18.7 / 43.5 & 18.8 / 44.5 & 19.3 / 45.5 \\
 \textbf{Average} & \textbf{32.7 / 59.2} & 30.8 / 57.7 & 32.1 / 58.4 & 31.8 / 58.5 & 1.6 / 13.8 & 21.9 / 47.4 & 22.5 / 48.7 & 22.8 / 49.4  \\ \bottomrule
\end{tabular}
}\vspace{3mm}
\end{table}

\begin{table}[h!]
\centering
\caption{\label{tab:flores} Evaluation scores on Flores-200 (depicted as \texttt{<bleu> / <chrf>}) for the MT models and language models described in Section \ref{sec:mt-model} and Section \ref{sec:lm} compared against NLLB-54B. All metrics are computed with the sacrebleu reference implementation.}
\resizebox{\columnwidth}{!}{
\begin{tabular}{@{}lcccccccc@{}}
\toprule
\multirow{2}{*}{} & \multicolumn{1}{c}{\multirow{2}{*}{\textbf{NLLB}}} & \multicolumn{1}{c}{\multirow{2}{*}{\textbf{MT-3B}}} & \multicolumn{1}{c}{\multirow{2}{*}{\textbf{MT-7.2B}}} & \multicolumn{1}{c}{\multirow{2}{*}{\textbf{MT-10.7B}}} & \multicolumn{4}{c}{\textbf{LM-8B}}                                                                                          \\ \cmidrule(l){6-9} 
                  &                                &                                              &                                                &                                       & 0-shot        & 1-shot        & 5-shot        & 10-shot       \\ \midrule

 \textbf{xx2en} & \textbf{35.5 / 59.6} & 29.7 / 54.4 & 30.9 / 55.4 & 31.9 / 56.4 & 2.0 / 13.3 & 20.5 / 44.1 & 22.3 / 46.9 & 22.4 / 47.6 \\
 \textbf{en2xx} & \textbf{20.7 / 50.}1 & 17.3 / 44.1 & 17.8 / 44.7 & 18.6 / 45.7 & 0.4 / 5.7 & 8.1 / 26.7 & 8.7 / 29.0 & 8.7 / 28.8 \\
 \textbf{Mean} & \textbf{28.2 / 54.9} & 23.5 / 49.2 & 24.4 / 50.0 & 25.3 / 51.1 & 1.2 / 9.6 & 14.3 / 35.5 & 15.6 / 38.0 & 15.6 / 38.2 \\
 \textbf{xx2yy} & \textbf{13.7 / 40.5} & 8.8 / 31.2 & 8.4 / 30.9 & 10.1 / 34.0 & 0.3 / 4.1 & 4.0 / 16.1 & 4.4 / 17.3 & 4.2 / 17.1 \\ \bottomrule
\end{tabular}
}
\end{table}

\subsubsection{Few-shot evaluation for language modeling}\label{sub:few-shot}

% \srkcomment{Biao TODO - also mention NTREX split}

We perform few-shot prompting to evaluate the language model with the following prompt:

\texttt{[sl]:$X_1$\textbackslash n[tl]:$Y_1$\textbackslash  n\textbackslash n[sl]:$X_2$\textbackslash  n[tl]:$Y_2$\textbackslash n\textbackslash n\ldots[sl]:$X$\textbackslash n[tl]:}

where \texttt{[sl]} and \texttt{[tl]} denote the source and target language name (expressed in English. For example, when translating a sentence from \texttt{en} to \texttt{te}, we use \texttt{[sl]=English} and \texttt{[tl]=Telugu}), respectively. $X_\star$ and $Y_\star$ are demonstration examples used for prompting, and $X$ is the test input.

For each test example, we randomly sample demonstration examples, which is simple yet performs competitively with more complicated strategies~\cite{vilar2022prompting,zhang2023prompting}. In particular, we randomly select examples from the dev split of each dataset. Since NTREX does not have a dev split, we randomly sample 1000 examples as the dev set and use the rest for test evaluation.

% \usepackage{booktabs}
% \usepackage{multirow}
\begin{table}[t!]
\centering
\caption{\label{tab:ntrex} Evaluation scores on the recently introduced NTREX test set (depicted as \texttt{<bleu> / <chrf>}) for the MT models and language models described in Section \ref{sec:mt-model} and Section \ref{sec:lm} compared against unsupervised baselines \cite{baziotis2023does}. Note that LM-8B is evaluated on a 50\% split of the NTREX data and is not comparable to the MT-model evaluations.}
\resizebox{\columnwidth}{!}{
% Original results
% \begin{tabular}{@{}lcccccccc@{}}
% \toprule
% \multirow{2}{*}{} & \multicolumn{1}{c}{\multirow{2}{*}{\textbf{\citet{baziotis2023does}}}} & \multicolumn{1}{c}{\multirow{2}{*}{\textbf{MT-3B}}} & \multicolumn{1}{c}{\multirow{2}{*}{\textbf{MT-7.2B}}} & \multicolumn{1}{c}{\multirow{2}{*}{\textbf{MT-10.7B}}} & \multicolumn{4}{c}{\textbf{LM-8B}}                                                                                          \\ \cmidrule(l){6-9} 
%  & & & & & 0-shot & 1-shot & 5-shot & 10-shot \\ \midrule
%  \textbf{xx2en} & 23.7 / 51.7 & 30.6 / 54.5 & 32.7 / 56.2 & \textbf{33.6 / 57.6} & 3.2 / 17.3 & 20.4 / 43.8 & 23.8 / 48.2 & 24.4 / 49.0 \\
%  \textbf{en2xx} & 15.9 / \textbf{44.9} & 16.5 / 39.6 & 17.6 / {41.9} & {17.9 / 41.9} & 0.8 / 7.3 & 11.7 / 31.2 & 12.6 / 32.4 & 12.3 / 32.3 \\
%  \textbf{Average} & 19.8 / 48.3 & 23.5 / 47.0 & 25.1 / 49.0 & 2\textbf{5.7 / 49.7} & 2.0 / 12.3 & 16.0 / 37.4 & 18.1 / 40.2 & 18.3 / 40.6 \\ \bottomrule
% \end{tabular}
\begin{tabular}{@{}lcccccccc@{}}
\toprule
\multirow{2}{*}{} & \multicolumn{1}{c}{\multirow{2}{*}{\textbf{\citet{baziotis2023does}}}} & \multicolumn{1}{c}{\multirow{2}{*}{\textbf{MT-3B}}} & \multicolumn{1}{c}{\multirow{2}{*}{\textbf{MT-7.2B}}} & \multicolumn{1}{c}{\multirow{2}{*}{\textbf{MT-10.7B}}} & \multicolumn{4}{c}{\textbf{LM-8B}}                                                                                          \\ \cmidrule(l){6-9} 
 & & & & & 0-shot & 1-shot & 5-shot & 10-shot \\ \midrule
 \multicolumn{8}{@{}l}{\bf Results on the subset of \citet{baziotis2023does}} \vspace{3pt} \\
 \textbf{xx2en} & 23.6 / 51.7  & 34.3 / 59.9 & \textbf{36.1} / 61.0 & 35.9 / \textbf{61.1} & 4.0 / 18.9 & 23.4 / 48.8 & 26.8 / 52.8 & 27.6 / 53.7 \\
 \textbf{en2xx} & 15.9 / 44.8 & 22.3 / 50.2 & \textbf{22.8} / 50.6 & \textbf{22.8} / \textbf{51.0} & 1.0 / 8.8 & 15.2 / 40.1 & 16.5 / 42.4 & 15.9 / 42.3 \\
 \textbf{Average} & 19.8 / 51.7 & 28.3 / 55.1 & \textbf{29.4} / 55.8 & \textbf{29.4} / \textbf{56.1} & 2.5 / 13.9 & 19.3 / 44.5 & 21.6 / 47.6 & 21.8 / 48.0 \\
 \midrule
 \multicolumn{8}{@{}l}{\bf Results on full test sets} \vspace{3pt} \\
 \textbf{xx2en} & - & 30.6 / 54.5 & 32.7 / 56.2 & \textbf{33.6 / 57.6} & 3.2 / 17.3 & 20.4 / 43.8 & 23.8 / 48.2 & 24.4 / 49.0 \\
 \textbf{en2xx} & - & 16.5 / 39.6 & 17.6 / \textbf{41.9} & \textbf{17.9 / 41.9} & 0.8 / 7.3 & 11.7 / 31.2 & 12.6 / 32.4 & 12.3 / 32.3 \\
 \textbf{Average} & - & 23.5 / 47.0 & 25.1 / 49.0 & \textbf{25.7 / 49.7} & 2.0 / 12.3 & 16.0 / 37.4 & 18.1 / 40.2 & 18.3 / 40.6 \\ \bottomrule
\end{tabular}
}
\vspace{3mm}
\end{table}

\subsection{Results}\label{sec:res}

In Tables~\ref{tab:wmt} and~\ref{tab:ntrex} we present evaluation scores on the WMT datasets and NTREX datasets, which are evaluation sets in the news domain. We find that both the 7.2B parameter model and the 10B parameter model is competitive with the significantly larger NLLB-54B model~\citep{nllb2022} on WMT. For the recent NTREX dataset, the only published results are small-scale results by \citet{baziotis2023does}.

In Table~\ref{tab:flores} we find that on Flores-200, our model is within 3.8 chrf of the 54B parameter NLLB model, while on \textbf{xxyy} pairs the 10.7B model is behind by 6.5 chrf. This is likely due to a combination of factors, including using a significantly smaller model (5x smaller), domain differences~\citep{baziotis2023does, bapna-etal-2022-building}, and back-translated data ~\citep{sennrich-etal-2016-improving}. Similarly, in Table~\ref{tab:ntl}, we find that the 10.7B parameter model is within 5.7 chrf of the scores reported by \citet{bapna-etal-2022-building}. Again, it is very difficult to compare their results to ours; their two largest advantages are 1) iterative back-translation, and 2) access to a much larger in-house text data. In Table \ref{tab:bt_for_mt}, we display the results for when we finetune the 7.2B parameter model on backtranslated data. While this setup is very likely sub-optimal, we see that back-translation greatly improves \textbf{en2xx} translation (by 3.0 chrf, in the case of Flores-200)  in most cases. We note that the results we present are merely baselines to demonstrate the utility of \data, and hope that future work builds upon these experiments by applying improved modeling techniques.

Finally, across all evaluation datasets, we find that while results on few-shot translation using the 8B language model increase with an increasing number of demonstrations, these results are still significantly weaker than the results of models trained on supervised data. We present per-language pair results on all datasets in Appendix~\ref{app:res}.
% \usepackage{booktabs}
% \usepackage{multirow}
\begin{table}[h!]
\centering
\caption{\label{tab:ntl}  Evaluation scores on the \textsc{Gatones} test set used by \citet{bapna-etal-2022-building} (depicted as \texttt{<bleu> / <chrf>}) for the MT models and language models described in Section \ref{sec:mt-model} and Section \ref{sec:lm}.}
\resizebox{\columnwidth}{!}{
\begin{tabular}{@{}lccccccccc@{}}
\toprule
\multirow{2}{*}{} & \multicolumn{2}{c}{{\textbf{NTL (\citet{bapna-etal-2022-building})}}} & \multicolumn{1}{c}{\multirow{2}{*}{\textbf{MT-3B}}} & \multicolumn{1}{c}{\multirow{2}{*}{\textbf{MT-7.2B}}} & \multicolumn{1}{c}{\multirow{2}{*}{\textbf{MT-10.7B}}} & \multicolumn{4}{c}{\textbf{LM-8B}}                                                                                          \\ \cmidrule(l){2-3}\cmidrule(l){7-10} 
 & 1.6B & 6.4B & & & & 0-shot & 1-shot & 5-shot & 10-shot \\ \midrule
% \textbf{xx2en} & {-} / 37.2 & {-} / \textbf{41.2} & 13.3 / 34.6 & \textbf{15.5} / 36.5 & 14.8 / 36.0 & 0.3 / 6.5 & 6.6 / 25.4 & 8.3 / 28.1 & 8.4 / 28.4 \\
% \textbf{en2xx} & {-} / 28.5 & {-} / \textbf{33.1} & 4.5 / 23.9 & \textbf{5.5} / 26.4 & 5.4 / 26.2 & 0.2 / 4.2 & 1.7 / 10.5 & 1.7 / 9.9 & 1.8 / 9.4 \\
% \textbf{Average} & {-} / 32.9 & {-} / \textbf{37.2} & 8.9 / 29.3 & \textbf{10.5} / 31.5 & 10.1 / 31.1 & 0.3 / 5.4 & 4.2 / 18.0 & 5.0 / 19.0 & 5.1 / 18.9 \\
\textbf{xx2en} & {-} / 37.2 & {-} / \textbf{41.2} & 13.3 / 34.6 & 14.8 / 36.0 & \textbf{15.4} / 37.0 & 0.3 / 6.5 & 6.6 / 25.4 & 8.3 / 28.1 & 8.4 / 28.4 \\
\textbf{en2xx} & {-} / 28.5 & {-} / \textbf{33.1} & 4.5 / 23.9 & \textbf{5.4} / 26.2 & \textbf{5.4} / 26.5 & 0.2 / 4.2 & 1.7 / 10.5 & 1.7 / 9.9 & 1.8 / 9.4 \\
\textbf{Average} & {-} / 32.9 & {-} / \textbf{37.2} & 8.9 / 29.3 & 10.1 / 31.1 & \textbf{10.4} / 31.8 & 0.3 / 5.4 & 4.2 / 18.0 & 5.0 / 19.0 & 5.1 / 18.9 \\

\bottomrule
\end{tabular}
}
\end{table}

% \usepackage{booktabs}
% \usepackage{multirow}
\begin{table}[h!]
\centering
\caption{\label{tab:bt_for_mt} Evaluation scores on different test sets (depicted as \texttt{<bleu> / <chrf>}) for {MT-7.2B} trained with back-translated data ({+BT}).}
\resizebox{\columnwidth}{!}{
% \begin{tabular}{@{}lcccccccc@{}}
% \toprule
%  & \multicolumn{2}{c}{\textbf{WMT}} & \multicolumn{2}{c}{\textbf{Flores-200}} & \multicolumn{2}{c}{\textbf{NTREX}} & \multicolumn{2}{c}{\textbf{\textsc{Gatones}}} \\
% \cmidrule(lr){2-3}
% \cmidrule(lr){4-5}
% \cmidrule(lr){6-7}
% \cmidrule(lr){8-9}
%  & \textbf{MT-7.2B} & \textbf{+BT} & \textbf{MT-7.2B} & \textbf{+BT} & \textbf{MT-7.2B} & \textbf{+BT} & \textbf{MT-7.2B} & \textbf{+BT} \\
% \midrule
% \textbf{xx2en}   & \textbf{60.6} & 60.4 & \textbf{55.4} & 53.9 & 56.2 & \textbf{56.5} & \textbf{36.0} & 33.6 \\
% \textbf{en2xx}   & 56.2 & \textbf{56.9} & 44.7 & \textbf{47.7} & 41.9 & \textbf{44.4} & \textbf{26.2} & 25.4 \\
% \textbf{average} & 58.4 & \textbf{58.6} & 50.0 & \textbf{50.8} & 49.0 & \textbf{50.4} & \textbf{31.1} & 29.7 \\ 
% \textbf{xx2yy} & - & - & 30.9 & \textbf{31.9} & - & - & - & - \\
% \bottomrule
% \end{tabular}
\begin{tabular}{@{}lcccccccc@{}}
\toprule
 & \multicolumn{2}{c}{\textbf{WMT}} & \multicolumn{2}{c}{\textbf{Flores-200}} & \multicolumn{2}{c}{\textbf{NTREX}} & \multicolumn{2}{c}{\textbf{\textsc{Gatones}}} \\
\cmidrule(lr){2-3}
\cmidrule(lr){4-5}
\cmidrule(lr){6-7}
\cmidrule(lr){8-9}
 & \textbf{MT-7.2B} & \textbf{+BT} & \textbf{MT-7.2B} & \textbf{+BT} & \textbf{MT-7.2B} & \textbf{+BT} & \textbf{MT-7.2B} & \textbf{+BT} \\
\midrule
\textbf{xx2en}   & \textbf{34.9} / \textbf{60.6} & 33.8 / 60.4 & \textbf{30.9} / \textbf{55.4} & 27.2 / 53.9 & \textbf{32.7} / 56.2 & 31.0 / \textbf{56.5} & \textbf{14.8} / \textbf{36.0} & 10.2 / 34.5 \\
\textbf{en2xx}   & 29.3 / 56.2 & \textbf{29.8} / \textbf{56.9} & 17.8 / 44.7 & \textbf{18.5} / \textbf{47.7} & 17.6 / 41.9 & \textbf{18.4} / \textbf{44.4} & \textbf{5.4} / \textbf{26.2} & 3.5 / 26.1 \\
\textbf{average} & \textbf{32.1} / 58.4 & 31.8 / \textbf{58.6} & \textbf{24.4} / 50.0 & 22.8 / \textbf{50.8} & \textbf{25.1} / 49.0 & 24.7 / \textbf{50.4} & \textbf{10.1} / \textbf{31.1} & 6.9 / 30.3 \\ 
\textbf{xx2yy} & - & - & 8.4 / 30.9 & 8.4 / \textbf{31.9} & - & - & - & - \\
\bottomrule
\end{tabular}
}
\end{table}

\section{Training Data Extraction and Memorization}\label{sec:memorization} 

Generative models have been shown to regurgitate training data~\cite{carlini2021extracting} that may plagiarize, violate copyright assumptions, or infringe privacy. It can be difficult to assess and prevent these cases because such information may be paraphrased in ways that are difficult for automated systems to detect~\cite{ippolito2022preventing}. Instead, existing literature measures memorization in generative models to estimate the propensity for disallowed outputs. Typically, this means prompting a language model with some prefix of length $P$ and comparing generated outputs of length $S$ with the training data to see if they are `novel' or if the generation is simply a regurgitation of its training data~\cite{carlini2021extracting,anil2023palm,ippolito2022preventing,jagielski2022measuring,carlini2022quantifying}. In the multilingual setting this may present new risks because tail languages may be more vulnerable to memorization~\cite{anil2023palm}.

\paragraph{The Difficulty of Assessing Memorization in Translation Settings.} While memorization has been well-studied for language models, assessing the extent of memorization is difficult within translation settings. This is primarily because translation has a significantly smaller space of valid outputs, as opposed to many possible continuations for language modeling. This presents some difficulty in extending common memorization tests for language generation to translation. As an illustrative example, consider the case of translating to the same target language as the source ("\texttt{translate\_copy}"). Performing a standard training data extraction attack would test if the generation matches the continuation. However, success would not indicate training data extraction as the adversary would have already had access to it.\footnote{Though membership inference may be possible.} Thus, we modify the standard framework for testing memorization to better identify \emph{additional} leaked data.

\paragraph{Memorization in Translation Settings} We define memorization in \texttt{translate\_copy} to be when the model outputs any generation with length $S>P$ that matches the continuation; then, $S-P$ captures the additional bits. In cases where the source and target language are different ("\texttt{translate\_diff}"), performing a similar test would require knowledge of which part of the continuation exactly corresponded to the prompt.
% - the exact problem of having a perfect translation model. 
Given that such an alignment is not easily obtained, we instead use the relative token lengths between the continuation and the prompt to choose an appropriate size of $S$. For example, if at training time the continuation for the target language was $1.5\times$ larger, we set $S=P \cdot 1.5 + \delta$ where $\delta$ captures the additional bits. For each of \texttt{translate\_copy} and \texttt{translate\_diff}, we sample $2,000$ sequences for each language and choose $P=50$. We then perform both a verbatim match of the generation with the continuation and an approximate match requiring $90\%$ Levenshtein similarity similar to~\cite{ippolito2022preventing}.

\begin{figure}[t]
    \centering
    \begin{subfigure}{0.45\textwidth}
    \centering
        \includegraphics[width=\linewidth]{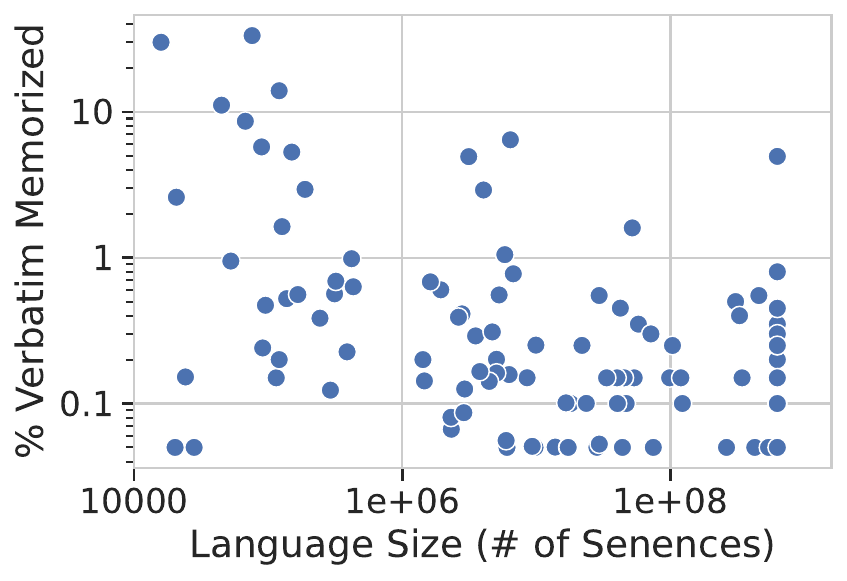}
        \label{fig:mono_verbatim_mem}
    \end{subfigure}
    \begin{subfigure}{0.45\textwidth}
    \centering
        \includegraphics[width=\linewidth]{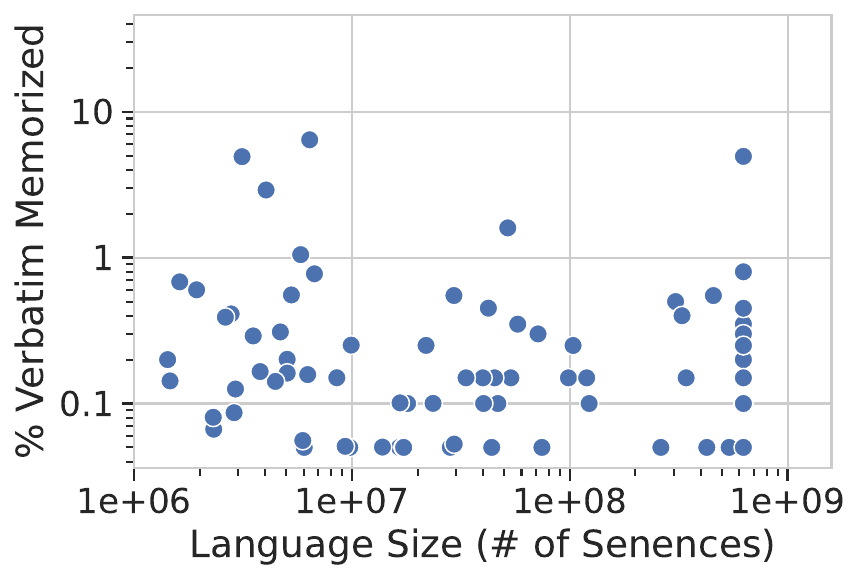}
        \label{fig:multi_verbatim_mem}
    \end{subfigure}
    \caption{\textbf{Monolingual (translate\_copy) data used in translation is more likely to be memorized.} Verbatim training data extraction rates for both \texttt{translate\_copy} \textbf{(left)} and  \texttt{translate\_diff} \textbf{(right)}  data. Extraction performed on the 3B parameter model using a $S=P+50$. In monoway, 257/370 languages exhibited no memorization in testing and 130/146 for multiway.}
    \label{fig:verbatim_mem}
\end{figure}

\paragraph{Results.} We show the per-language and average training data extraction rates, for both the \texttt{translate\_copy} and \texttt{translate\_diff}  
~settings in Figure~\ref{fig:verbatim_mem}, with $S$ set to test for $50$ tokens of additional information leakage. We find that translate models can memorize and regurgitate their training data, even beyond what is contained in the prompt. We also observe that some lower resource languages may exhibit higher memorization rates, however we observe no strong correlation between the resource level and the level of memorization. In the \texttt{translate\_diff}  tests, we observe  much lower memorization - we hypothesize this may be due to the higher difficulty of the task. Even though many languages have nontrivial memorization, we found that many languages exhibited no memorization across the samples tested (257/370 for \texttt{translate\_copy}  and 130/146 for \texttt{translate\_diff} ). We also present results for approximate memorization in Appendix~\ref{sec:memorization-details}, which show that translate models may also paraphrase memorizations leading to even higher memorization rates.

\paragraph{Discussion} Our preliminary experiments show that memorization can exist in the translation setting. However, capturing when memorization is intended or beneficial versus undesired is still an open question.
% However, it also begs to question how to accurately capture 
%  We hope that by studying the propensity for memorization of neural machine translation models, these questions may begin to be addressed.
To aid future research in this direction, we design and include ``canaries''---carefully crafted data designed to be outliers to the natural training distribution that can be used to analyze memorization. Canaries enable studying memorization in the multilingual and machine translation settings by measuring the capability to extract canaries added to the training set~\cite{anil2023palm,jagielski2022measuring}. As with \citet{anil2023palm}, our canaries are designed to share characteristics with the natural training data so as to better ground memorization evaluation in practical risks. The canaries are also designed tosl be outliers to assess varying degrees of risk. To ensure similarity with natural data, canaries are generated by sampling and then randomly modifying real data in a manner similar to~\cite{anil2023palm}, where each source of randomness defines the canary type. In total, we generate $1,945,631$ canaries across both the monolingual \data dataset and the parallel data ($\approx0.0026\%$ of the training data).  The methodology for each canary type and the exact distribution of canaries are detailed in Appendix~\ref{sec:canary-details}.

\section{Related Work}
\label{sec:rw}
Extensive work has been done to mine general purpose datasets for multilingual machine translation and language modeling. \citet{xue2020mt5} introduce mC4, a general web domain corpus on 101 languages to train mT5, a pretrained language model for downstream NLP tasks. Similarly, \citet{conneau2019unsupervised} introduce CC-100, later extended to CC100-XL by \citet{lin2021few}. The OSCAR corpus~\citep{oscar} is also a mined dataset that supports 166 languages and the ROOTS corpus is a compiled dataset that contains 46 natural languages. Glot500-C~\citep{imanigooghari2023glot500} covers 511 languages: however, it is not clear how many of these languages comprise solely of religious texts. \citet{bapna-etal-2022-building} create an internal dataset on 1500+ languages, while \citet{nllb2022} mine a dataset from CommonCrawl and ParaCrawl~\citep{espla2019paracrawl}. Recently, \citet{leong-etal-2022-bloom} created a 350+ language dataset from children's books.

In addition, there have been efforts to get better represented corpora and models for languages often underrepresented in general multilingual corpora: Serengeti~\citep{adebara2022serengeti} introduces a dataset and associated model trained on 517 African languages and language varieties, while IndicTrans2~\citep{gala2023indictrans2} introduces a machine translated model for the 22 scheduled languages in India. 
\section{Limitations}
\label{sec:limit}
While we used thorough self-audits to guide the creation of \data, we note that most audits were conducted by non-speakers of the languages in \data; as a result, many types of noise, like machine-generated or disfluent content, could not be detected. Moreover, toxicity detectors, classifiers and filters that work reliably for all the 419 languages in \data do not exist, limiting the extent to which we can clean and document~\citep{dodge2021documenting,bandy2021addressing} the dataset. It is possible that issues still remain, so we encourage users to report issues that will be listed on the project Github page\footnote{\url{https://github.com/google-research/google-research/tree/master/madlad_400}}.
This paucity extends to the availability of multilingual evaluation sets for these languages - we could only evaluate our models on 204 of the languages in \data. Additionally, even though decoder-only models are often evaluated on NLP tasks that are not necessarily machine translation \cite{hu2020xtreme,asai2023buffet,ahuja2023mega}, we did not conduct such evaluations - most available benchmarks cover only 30-50 languages of which most are not tail languages (which forms the focus of \data). We instead leave this to future work. Finally, during our self-audit we noted the skew of data on the long tail towards specific domains such as religious texts. We hope that these limitations motivate the creation of more language-specific corpora not captured by web crawls, and the development of language-specific data cleaning tools and practices.

\section{Conclusion}
\label{sec:disc}
Through \data, we introduce a highly multilingual, general web-domain, document-level text dataset. We perform a self-audit of this dataset for quality on samples of all 498 languages, develop filters, and remove spurious datasets, for a total of 419 languages in the release. We carefully describe the dataset creation process, laying out the iterations of audits and improvements upon the preliminary dataset along with observations that guided our decisions. We hope that this encourages creators of large-scale pretraining datasets both to put in their due diligence for manually inspecting and dealing with data, and also to describe and publicize the process in a level of detail that is reproducible and insightful for downstream users. This increased visibility into the dataset creation cycle can in turn improve model development and enable responsible data use \cite{sambasivan2021everyone}. Using \data, we train and release large machine translation and general NLP models and evaluate them thoroughly. We hope that this further motivates work towards language technologies that are more inclusive of the rich language diversity housed by humanity.

\section{Ethics Statement}
\label{sec:ethics}
Innovation in NLP technologies in English has been accelerated by training large scale deep learning models~\citep{devlin2018bert,brown2020language} on massive web corpora~\citep{chelba2013one,zhu2015aligning,raffel2020exploring}. However, on the long tail of written languages in the world there is a lack of high quality general data sources~\citep{joshi2020state} that impede the progress of NLP tools for other languages. We hope that making an audited and cleaned corpus such as \data available mitigates this issue. While we extensively cleaned \data, the extent to which we can preprocess this data is limited by how not all languages have available tools for removing problematic content such as porn, toxic content, PII, copyrighted content or noise. We urge practitioners to carefully consider their target usecase before using \data.

\section*{Acknowledgements}

We would like to thank Wolfgang Macherey, Zoubin Ghahramani and Orevaoghene Ahia for their helpful comments on the draft. We would also like to thank Subramanian Venkateswaran for debugging the virama rendering issues, and Ali Dabirmoghaddam for his insight on data samples of various languages in \data. 
\bibliography{neurips_data_2023,anthology}

\begin{thebibliography}{74}
\providecommand{\natexlab}[1]{#1}
\providecommand{\url}[1]{\texttt{#1}}
\expandafter\ifx\csname urlstyle\endcsname\relax
  \providecommand{\doi}[1]{doi: #1}\else
  \providecommand{\doi}{doi: \begingroup \urlstyle{rm}\Url}\fi

\bibitem[sta()]{statmt}
{StatMT}.
\newblock \url{https://www.statmt.org/}.
\newblock Accessed: 2022-05-03.

\bibitem[Abadji et~al.(2022)Abadji, Suarez, Romary, and Sagot]{oscar}
J.~Abadji, P.~O. Suarez, L.~Romary, and B.~Sagot.
\newblock Towards a cleaner document-oriented multilingual crawled corpus.
\newblock \emph{arXiv preprint arXiv:2201.06642}, 2022.

\bibitem[Adebara et~al.(2022)Adebara, Elmadany, Abdul-Mageed, and
  Inciarte]{adebara2022serengeti}
I.~Adebara, A.~Elmadany, M.~Abdul-Mageed, and A.~A. Inciarte.
\newblock Serengeti: Massively multilingual language models for africa.
\newblock \emph{arXiv preprint arXiv:2212.10785}, 2022.

\bibitem[Agic and Vulic(2019)]{agic2019jw300}
{\v{Z}}.~Agic and I.~Vulic.
\newblock Jw300: A wide-coverage parallel corpus for low-resource languages.
\newblock Association for Computational Linguistics, 2019.

\bibitem[Ahuja et~al.(2023)Ahuja, Hada, Ochieng, Jain, Diddee, Maina, Ganu,
  Segal, Axmed, Bali, et~al.]{ahuja2023mega}
K.~Ahuja, R.~Hada, M.~Ochieng, P.~Jain, H.~Diddee, S.~Maina, T.~Ganu, S.~Segal,
  M.~Axmed, K.~Bali, et~al.
\newblock Mega: Multilingual evaluation of generative ai.
\newblock \emph{arXiv preprint arXiv:2303.12528}, 2023.

\bibitem[Anil et~al.(2023)Anil, Dai, Firat, Johnson, Lepikhin, Passos, Shakeri,
  Taropa, Bailey, Chen, et~al.]{anil2023palm}
R.~Anil, A.~M. Dai, O.~Firat, M.~Johnson, D.~Lepikhin, A.~Passos, S.~Shakeri,
  E.~Taropa, P.~Bailey, Z.~Chen, et~al.
\newblock Palm 2 technical report.
\newblock \emph{arXiv preprint arXiv:2305.10403}, 2023.

\bibitem[Asai et~al.(2023)Asai, Kudugunta, Yu, Blevins, Gonen, Reid, Tsvetkov,
  Ruder, and Hajishirzi]{asai2023buffet}
A.~Asai, S.~Kudugunta, X.~V. Yu, T.~Blevins, H.~Gonen, M.~Reid, Y.~Tsvetkov,
  S.~Ruder, and H.~Hajishirzi.
\newblock Buffet: Benchmarking large language models for few-shot cross-lingual
  transfer.
\newblock \emph{arXiv preprint arXiv:2305.14857}, 2023.

\bibitem[Bandy and Vincent(2021)]{bandy2021addressing}
J.~Bandy and N.~Vincent.
\newblock Addressing" documentation debt" in machine learning research: A
  retrospective datasheet for bookcorpus.
\newblock \emph{arXiv preprint arXiv:2105.05241}, 2021.

\bibitem[{Bapna} et~al.(2022){Bapna}, {Caswell}, {Kreutzer}, {Firat}, {van
  Esch}, {Siddhant}, {Niu}, {Baljekar}, {Garcia}, {Macherey}, {Breiner},
  {Axelrod}, {Riesa}, {Cao}, {Chen}, {Macherey}, {Krikun}, {Wang}, {Gutkin},
  {Shah}, {Huang}, {Chen}, {Wu}, and {Hughes}]{bapna-etal-2022-building}
A.~{Bapna}, I.~{Caswell}, J.~{Kreutzer}, O.~{Firat}, D.~{van Esch},
  A.~{Siddhant}, M.~{Niu}, P.~{Baljekar}, X.~{Garcia}, W.~{Macherey},
  T.~{Breiner}, V.~{Axelrod}, J.~{Riesa}, Y.~{Cao}, M.~X. {Chen},
  K.~{Macherey}, M.~{Krikun}, P.~{Wang}, A.~{Gutkin}, A.~{Shah}, Y.~{Huang},
  Z.~{Chen}, Y.~{Wu}, and M.~{Hughes}.
\newblock {Building Machine Translation Systems for the Next Thousand
  Languages}.
\newblock \emph{arXiv e-prints}, art. arXiv:2205.03983, May 2022.

\bibitem[Baziotis et~al.(2023)Baziotis, Zhang, Birch, and
  Haddow]{baziotis2023does}
C.~Baziotis, B.~Zhang, A.~Birch, and B.~Haddow.
\newblock When does monolingual data help multilingual translation: The role of
  domain and model scale.
\newblock \emph{arXiv preprint arXiv:2305.14124}, 2023.

\bibitem[Bommasani et~al.(2021)Bommasani, Hudson, Adeli, Altman, Arora, von
  Arx, Bernstein, Bohg, Bosselut, Brunskill,
  et~al.]{bommasani2021opportunities}
R.~Bommasani, D.~A. Hudson, E.~Adeli, R.~Altman, S.~Arora, S.~von Arx, M.~S.
  Bernstein, J.~Bohg, A.~Bosselut, E.~Brunskill, et~al.
\newblock On the opportunities and risks of foundation models.
\newblock \emph{arXiv preprint arXiv:2108.07258}, 2021.

\bibitem[Brown et~al.(2020)Brown, Mann, Ryder, Subbiah, Kaplan, Dhariwal,
  Neelakantan, Shyam, Sastry, Askell, et~al.]{brown2020language}
T.~Brown, B.~Mann, N.~Ryder, M.~Subbiah, J.~D. Kaplan, P.~Dhariwal,
  A.~Neelakantan, P.~Shyam, G.~Sastry, A.~Askell, et~al.
\newblock Language models are few-shot learners.
\newblock \emph{Advances in neural information processing systems},
  33:\penalty0 1877--1901, 2020.

\bibitem[Carlini et~al.(2021)Carlini, Tramer, Wallace, Jagielski, Herbert-Voss,
  Lee, Roberts, Brown, Song, Erlingsson, et~al.]{carlini2021extracting}
N.~Carlini, F.~Tramer, E.~Wallace, M.~Jagielski, A.~Herbert-Voss, K.~Lee,
  A.~Roberts, T.~Brown, D.~Song, U.~Erlingsson, et~al.
\newblock Extracting training data from large language models.
\newblock In \emph{30th USENIX Security Symposium (USENIX Security 21)}, pages
  2633--2650, 2021.

\bibitem[Carlini et~al.(2022)Carlini, Ippolito, Jagielski, Lee, Tramer, and
  Zhang]{carlini2022quantifying}
N.~Carlini, D.~Ippolito, M.~Jagielski, K.~Lee, F.~Tramer, and C.~Zhang.
\newblock Quantifying memorization across neural language models.
\newblock \emph{arXiv preprint arXiv:2202.07646}, 2022.

\bibitem[Caswell et~al.(2020)Caswell, Breiner, van Esch, and
  Bapna]{caswell2020language}
I.~Caswell, T.~Breiner, D.~van Esch, and A.~Bapna.
\newblock Language id in the wild: Unexpected challenges on the path to a
  thousand-language web text corpus, 2020.
\newblock URL \url{https://arxiv.org/abs/2010.14571}.

\bibitem[Chelba et~al.(2013)Chelba, Mikolov, Schuster, Ge, Brants, Koehn, and
  Robinson]{chelba2013one}
C.~Chelba, T.~Mikolov, M.~Schuster, Q.~Ge, T.~Brants, P.~Koehn, and
  T.~Robinson.
\newblock One billion word benchmark for measuring progress in statistical
  language modeling.
\newblock \emph{arXiv preprint arXiv:1312.3005}, 2013.

\bibitem[Chowdhery et~al.(2022)Chowdhery, Narang, Devlin, Bosma, Mishra,
  Roberts, Barham, Chung, Sutton, Gehrmann, et~al.]{chowdhery2022palm}
A.~Chowdhery, S.~Narang, J.~Devlin, M.~Bosma, G.~Mishra, A.~Roberts, P.~Barham,
  H.~W. Chung, C.~Sutton, S.~Gehrmann, et~al.
\newblock Palm: Scaling language modeling with pathways.
\newblock \emph{arXiv preprint arXiv:2204.02311}, 2022.

\bibitem[Chung et~al.(2023)Chung, Constant, Garcia, Roberts, Tay, Narang, and
  Firat]{chung2023unimax}
H.~W. Chung, N.~Constant, X.~Garcia, A.~Roberts, Y.~Tay, S.~Narang, and
  O.~Firat.
\newblock Unimax: Fairer and more effective language sampling for large-scale
  multilingual pretraining.
\newblock \emph{arXiv preprint arXiv:2304.09151}, 2023.

\bibitem[Conneau et~al.(2019)Conneau, Khandelwal, Goyal, Chaudhary, Wenzek,
  Guzm{\'a}n, Grave, Ott, Zettlemoyer, and Stoyanov]{conneau2019unsupervised}
A.~Conneau, K.~Khandelwal, N.~Goyal, V.~Chaudhary, G.~Wenzek, F.~Guzm{\'a}n,
  E.~Grave, M.~Ott, L.~Zettlemoyer, and V.~Stoyanov.
\newblock Unsupervised cross-lingual representation learning at scale.
\newblock \emph{arXiv preprint arXiv:1911.02116}, 2019.

\bibitem[Devlin et~al.(2018)Devlin, Chang, Lee, and Toutanova]{devlin2018bert}
J.~Devlin, M.-W. Chang, K.~Lee, and K.~Toutanova.
\newblock Bert: Pre-training of deep bidirectional transformers for language
  understanding.
\newblock \emph{arXiv preprint arXiv:1810.04805}, 2018.

\bibitem[Dodge et~al.(2021)Dodge, Sap, Marasovi{\'c}, Agnew, Ilharco,
  Groeneveld, Mitchell, and Gardner]{dodge2021documenting}
J.~Dodge, M.~Sap, A.~Marasovi{\'c}, W.~Agnew, G.~Ilharco, D.~Groeneveld,
  M.~Mitchell, and M.~Gardner.
\newblock Documenting large webtext corpora: A case study on the colossal clean
  crawled corpus.
\newblock \emph{arXiv preprint arXiv:2104.08758}, 2021.

\bibitem[Espl{\`a}-Gomis et~al.(2019)Espl{\`a}-Gomis, Forcada,
  Ram{\'\i}rez-S{\'a}nchez, and Hoang]{espla2019paracrawl}
M.~Espl{\`a}-Gomis, M.~L. Forcada, G.~Ram{\'\i}rez-S{\'a}nchez, and H.~Hoang.
\newblock Paracrawl: Web-scale parallel corpora for the languages of the eu.
\newblock In \emph{Proceedings of Machine Translation Summit XVII: Translator,
  Project and User Tracks}, pages 118--119, 2019.

\bibitem[Federmann et~al.(2022)Federmann, Kocmi, and
  Xin]{federmann-etal-2022-ntrex}
C.~Federmann, T.~Kocmi, and Y.~Xin.
\newblock {NTREX}-128 {--} news test references for {MT} evaluation of 128
  languages.
\newblock In \emph{Proceedings of the First Workshop on Scaling Up Multilingual
  Evaluation}, pages 21--24, Online, Nov. 2022. Association for Computational
  Linguistics.
\newblock URL \url{https://aclanthology.org/2022.sumeval-1.4}.

\bibitem[Fernando et~al.(2020)Fernando, Ranathunga, and Dias]{fernando2020data}
A.~Fernando, S.~Ranathunga, and G.~Dias.
\newblock Data augmentation and terminology integration for domain-specific
  sinhala-english-tamil statistical machine translation.
\newblock \emph{arXiv preprint arXiv:2011.02821}, 2020.

\bibitem[Freitag and Firat(2020)]{multiway}
M.~Freitag and O.~Firat.
\newblock Complete multilingual neural machine translation.
\newblock \emph{CoRR}, abs/2010.10239, 2020.
\newblock URL \url{https://arxiv.org/abs/2010.10239}.

\bibitem[Gala et~al.(2023)Gala, Chitale, AK, Doddapaneni, Gumma, Kumar, Nawale,
  Sujatha, Puduppully, Raghavan, et~al.]{gala2023indictrans2}
J.~Gala, P.~A. Chitale, R.~AK, S.~Doddapaneni, V.~Gumma, A.~Kumar, J.~Nawale,
  A.~Sujatha, R.~Puduppully, V.~Raghavan, et~al.
\newblock Indictrans2: Towards high-quality and accessible machine translation
  models for all 22 scheduled indian languages.
\newblock \emph{arXiv preprint arXiv:2305.16307}, 2023.

\bibitem[Garcia et~al.(2023)Garcia, Bansal, Cherry, Foster, Krikun, Feng,
  Johnson, and Firat]{garcia2023unreasonable}
X.~Garcia, Y.~Bansal, C.~Cherry, G.~Foster, M.~Krikun, F.~Feng, M.~Johnson, and
  O.~Firat.
\newblock The unreasonable effectiveness of few-shot learning for machine
  translation.
\newblock \emph{arXiv preprint arXiv:2302.01398}, 2023.

\bibitem[Groenewald and Fourie(2009)]{groenewald2009introducing}
H.~J. Groenewald and W.~Fourie.
\newblock Introducing the autshumato integrated translation environment.
\newblock In \emph{Proceedings of the 13th Annual conference of the European
  Association for Machine Translation}, 2009.

\bibitem[Haddow and Kirefu(2020)]{haddow2020pmindia}
B.~Haddow and F.~Kirefu.
\newblock Pmindia--a collection of parallel corpora of languages of india.
\newblock \emph{arXiv preprint arXiv:2001.09907}, 2020.

\bibitem[Hu et~al.(2020)Hu, Ruder, Siddhant, Neubig, Firat, and
  Johnson]{hu2020xtreme}
J.~Hu, S.~Ruder, A.~Siddhant, G.~Neubig, O.~Firat, and M.~Johnson.
\newblock Xtreme: A massively multilingual multi-task benchmark for evaluating
  cross-lingual generalisation.
\newblock In \emph{International Conference on Machine Learning}, pages
  4411--4421. PMLR, 2020.

\bibitem[ImaniGooghari et~al.(2023)ImaniGooghari, Lin, Kargaran, Severini,
  Sabet, Kassner, Ma, Schmid, Martins, Yvon, et~al.]{imanigooghari2023glot500}
A.~ImaniGooghari, P.~Lin, A.~H. Kargaran, S.~Severini, M.~J. Sabet, N.~Kassner,
  C.~Ma, H.~Schmid, A.~F. Martins, F.~Yvon, et~al.
\newblock Glot500: Scaling multilingual corpora and language models to 500
  languages.
\newblock \emph{arXiv preprint arXiv:2305.12182}, 2023.

\bibitem[Ippolito et~al.(2022)Ippolito, Tram{\`e}r, Nasr, Zhang, Jagielski,
  Lee, Choquette-Choo, and Carlini]{ippolito2022preventing}
D.~Ippolito, F.~Tram{\`e}r, M.~Nasr, C.~Zhang, M.~Jagielski, K.~Lee, C.~A.
  Choquette-Choo, and N.~Carlini.
\newblock Preventing verbatim memorization in language models gives a false
  sense of privacy.
\newblock \emph{arXiv preprint arXiv:2210.17546}, 2022.

\bibitem[Jagielski et~al.(2022)Jagielski, Thakkar, Tramer, Ippolito, Lee,
  Carlini, Wallace, Song, Thakurta, Papernot, et~al.]{jagielski2022measuring}
M.~Jagielski, O.~Thakkar, F.~Tramer, D.~Ippolito, K.~Lee, N.~Carlini,
  E.~Wallace, S.~Song, A.~Thakurta, N.~Papernot, et~al.
\newblock Measuring forgetting of memorized training examples.
\newblock \emph{arXiv preprint arXiv:2207.00099}, 2022.

\bibitem[Joanis et~al.(2020)Joanis, Knowles, Kuhn, Larkin, Littell, Lo,
  Stewart, and Micher]{joanis2020nunavut}
E.~Joanis, R.~Knowles, R.~Kuhn, S.~Larkin, P.~Littell, C.-k. Lo, D.~Stewart,
  and J.~Micher.
\newblock The nunavut hansard inuktitut--english parallel corpus 3.0 with
  preliminary machine translation results.
\newblock In \emph{Proceedings of The 12th Language Resources and Evaluation
  Conference}, pages 2562--2572, 2020.

\bibitem[Johnson et~al.(2017)Johnson, Schuster, Le, Krikun, Wu, Chen, Thorat,
  Vi{\'e}gas, Wattenberg, Corrado, et~al.]{johnson2017google}
M.~Johnson, M.~Schuster, Q.~V. Le, M.~Krikun, Y.~Wu, Z.~Chen, N.~Thorat,
  F.~Vi{\'e}gas, M.~Wattenberg, G.~Corrado, et~al.
\newblock Google’s multilingual neural machine translation system: Enabling
  zero-shot translation.
\newblock \emph{Transactions of the Association for Computational Linguistics},
  5:\penalty0 339--351, 2017.

\bibitem[Jones et~al.(2023)Jones, Caswell, Saxena, and Firat]{jones2023bilex}
A.~Jones, I.~Caswell, I.~Saxena, and O.~Firat.
\newblock Bilex rx: Lexical data augmentation for massively multilingual
  machine translation, 2023.

\bibitem[Joshi et~al.(2020)Joshi, Santy, Budhiraja, Bali, and
  Choudhury]{joshi2020state}
P.~Joshi, S.~Santy, A.~Budhiraja, K.~Bali, and M.~Choudhury.
\newblock The state and fate of linguistic diversity and inclusion in the nlp
  world.
\newblock \emph{arXiv preprint arXiv:2004.09095}, 2020.

\bibitem[Kim et~al.(2021)Kim, Awan, Muzio, Salinas, Lu, Hendy, Rajbhandari, He,
  and Awadalla]{kim2021scalable}
Y.~J. Kim, A.~A. Awan, A.~Muzio, A.~F.~C. Salinas, L.~Lu, A.~Hendy,
  S.~Rajbhandari, Y.~He, and H.~H. Awadalla.
\newblock Scalable and efficient moe training for multitask multilingual
  models.
\newblock \emph{arXiv preprint arXiv:2109.10465}, 2021.

\bibitem[Koehn(2005)]{koehn2005europarl}
P.~Koehn.
\newblock Europarl: A parallel corpus for statistical machine translation.
\newblock In \emph{Proceedings of machine translation summit x: papers}, pages
  79--86, 2005.

\bibitem[Kreutzer et~al.(2022)Kreutzer, Caswell, Wang, Wahab, van Esch,
  Ulzii-Orshikh, Tapo, Subramani, Sokolov, Sikasote, Setyawan, Sarin, Samb,
  Sagot, Rivera, Rios, Papadimitriou, Osei, Suarez, Orife, Ogueji, Rubungo,
  Nguyen, M{\"u}ller, M{\"u}ller, Muhammad, Muhammad, Mnyakeni, Mirzakhalov,
  Matangira, Leong, Lawson, Kudugunta, Jernite, Jenny, Firat, Dossou, Dlamini,
  de~Silva, {\c{C}}abuk~Ball{\i}, Biderman, Battisti, Baruwa, Bapna, Baljekar,
  Azime, Awokoya, Ataman, Ahia, Ahia, Agrawal, and
  Adeyemi]{kreutzer-etal-2022-quality}
J.~Kreutzer, I.~Caswell, L.~Wang, A.~Wahab, D.~van Esch, N.~Ulzii-Orshikh,
  A.~Tapo, N.~Subramani, A.~Sokolov, C.~Sikasote, M.~Setyawan, S.~Sarin,
  S.~Samb, B.~Sagot, C.~Rivera, A.~Rios, I.~Papadimitriou, S.~Osei, P.~O.
  Suarez, I.~Orife, K.~Ogueji, A.~N. Rubungo, T.~Q. Nguyen, M.~M{\"u}ller,
  A.~M{\"u}ller, S.~H. Muhammad, N.~Muhammad, A.~Mnyakeni, J.~Mirzakhalov,
  T.~Matangira, C.~Leong, N.~Lawson, S.~Kudugunta, Y.~Jernite, M.~Jenny,
  O.~Firat, B.~F.~P. Dossou, S.~Dlamini, N.~de~Silva, S.~{\c{C}}abuk~Ball{\i},
  S.~Biderman, A.~Battisti, A.~Baruwa, A.~Bapna, P.~Baljekar, I.~A. Azime,
  A.~Awokoya, D.~Ataman, O.~Ahia, O.~Ahia, S.~Agrawal, and M.~Adeyemi.
\newblock Quality at a glance: An audit of web-crawled multilingual datasets.
\newblock \emph{Transactions of the Association for Computational Linguistics},
  10:\penalty0 50--72, 2022.
\newblock \doi{10.1162/tacl_a_00447}.
\newblock URL \url{https://aclanthology.org/2022.tacl-1.4}.

\bibitem[Kudo and Richardson(2018)]{kudo2018sentencepiece}
T.~Kudo and J.~Richardson.
\newblock Sentencepiece: A simple and language independent subword tokenizer
  and detokenizer for neural text processing.
\newblock \emph{arXiv preprint arXiv:1808.06226}, 2018.

\bibitem[Kudugunta et~al.(2021)Kudugunta, Huang, Bapna, Krikun, Lepikhin,
  Luong, and Firat]{kudugunta-etal-2021-beyond-distillation}
S.~Kudugunta, Y.~Huang, A.~Bapna, M.~Krikun, D.~Lepikhin, M.-T. Luong, and
  O.~Firat.
\newblock Beyond distillation: Task-level mixture-of-experts for efficient
  inference.
\newblock In \emph{Findings of the Association for Computational Linguistics:
  EMNLP 2021}, pages 3577--3599, Punta Cana, Dominican Republic, Nov. 2021.
  Association for Computational Linguistics.
\newblock \doi{10.18653/v1/2021.findings-emnlp.304}.
\newblock URL \url{https://aclanthology.org/2021.findings-emnlp.304}.

\bibitem[Lauren{\c{c}}on et~al.(2022)Lauren{\c{c}}on, Saulnier, Wang, Akiki,
  Villanova~del Moral, Le~Scao, Von~Werra, Mou, Gonz{\'a}lez~Ponferrada,
  Nguyen, et~al.]{laurenccon2022bigscience}
H.~Lauren{\c{c}}on, L.~Saulnier, T.~Wang, C.~Akiki, A.~Villanova~del Moral,
  T.~Le~Scao, L.~Von~Werra, C.~Mou, E.~Gonz{\'a}lez~Ponferrada, H.~Nguyen,
  et~al.
\newblock The bigscience roots corpus: A 1.6 tb composite multilingual dataset.
\newblock \emph{Advances in Neural Information Processing Systems},
  35:\penalty0 31809--31826, 2022.

\bibitem[Lee et~al.(2021)Lee, Ippolito, Nystrom, Zhang, Eck, Callison-Burch,
  and Carlini]{lee2021deduplicating}
K.~Lee, D.~Ippolito, A.~Nystrom, C.~Zhang, D.~Eck, C.~Callison-Burch, and
  N.~Carlini.
\newblock Deduplicating training data makes language models better.
\newblock \emph{arXiv preprint arXiv:2107.06499}, 2021.

\bibitem[Leong et~al.(2022)Leong, Nemecek, Mansdorfer, Filighera, Owodunni, and
  Whitenack]{leong-etal-2022-bloom}
C.~Leong, J.~Nemecek, J.~Mansdorfer, A.~Filighera, A.~Owodunni, and
  D.~Whitenack.
\newblock Bloom library: Multimodal datasets in 300+ languages for a variety of
  downstream tasks.
\newblock In \emph{Proceedings of the 2022 Conference on Empirical Methods in
  Natural Language Processing}, pages 8608--8621, Abu Dhabi, United Arab
  Emirates, Dec. 2022. Association for Computational Linguistics.
\newblock URL \url{https://aclanthology.org/2022.emnlp-main.590}.

\bibitem[Liebling et~al.(2022)Liebling, Heller, Robertson, and
  Deng]{liebling2022opportunities}
D.~Liebling, K.~Heller, S.~Robertson, and W.~Deng.
\newblock Opportunities for human-centered evaluation of machine translation
  systems.
\newblock In \emph{Findings of the Association for Computational Linguistics:
  NAACL 2022}, pages 229--240, 2022.

\bibitem[Lin et~al.(2021)Lin, Mihaylov, Artetxe, Wang, Chen, Simig, Ott, Goyal,
  Bhosale, Du, et~al.]{lin2021few}
X.~V. Lin, T.~Mihaylov, M.~Artetxe, T.~Wang, S.~Chen, D.~Simig, M.~Ott,
  N.~Goyal, S.~Bhosale, J.~Du, et~al.
\newblock Few-shot learning with multilingual language models.
\newblock \emph{arXiv preprint arXiv:2112.10668}, 2021.

\bibitem[Luccioni and Viviano(2021)]{luccioni2021s}
A.~S. Luccioni and J.~D. Viviano.
\newblock What's in the box? a preliminary analysis of undesirable content in
  the common crawl corpus.
\newblock \emph{arXiv preprint arXiv:2105.02732}, 2021.

\bibitem[Nakazawa et~al.(2016)Nakazawa, Yaguchi, Uchimoto, Utiyama, Sumita,
  Kurohashi, and Isahara]{nakazawa2016aspec}
T.~Nakazawa, M.~Yaguchi, K.~Uchimoto, M.~Utiyama, E.~Sumita, S.~Kurohashi, and
  H.~Isahara.
\newblock Aspec: Asian scientific paper excerpt corpus.
\newblock In \emph{Proceedings of the Tenth International Conference on
  Language Resources and Evaluation (LREC'16)}, pages 2204--2208, 2016.

\bibitem[Neubig(2011)]{neubig11kftt}
G.~Neubig.
\newblock The {Kyoto} free translation task.
\newblock http://www.phontron.com/kftt, 2011.

\bibitem[NLLBTeam et~al.(2022)NLLBTeam, Costa-jussà, Cross, Çelebi, Elbayad,
  Heafield, Heffernan, Kalbassi, Lam, Licht, Maillard, Sun, Wang, Wenzek,
  Youngblood, Akula, Barrault, Gonzalez, Hansanti, Hoffman, Jarrett, Sadagopan,
  Rowe, Spruit, Tran, Andrews, Ayan, Bhosale, Edunov, Fan, Gao, Goswami,
  Guzmán, Koehn, Mourachko, Ropers, Saleem, Schwenk, and Wang]{nllb2022}
NLLBTeam, M.~R. Costa-jussà, J.~Cross, O.~Çelebi, M.~Elbayad, K.~Heafield,
  K.~Heffernan, E.~Kalbassi, J.~Lam, D.~Licht, J.~Maillard, A.~Sun, S.~Wang,
  G.~Wenzek, A.~Youngblood, B.~Akula, L.~Barrault, G.~M. Gonzalez, P.~Hansanti,
  J.~Hoffman, S.~Jarrett, K.~R. Sadagopan, D.~Rowe, S.~Spruit, C.~Tran,
  P.~Andrews, N.~F. Ayan, S.~Bhosale, S.~Edunov, A.~Fan, C.~Gao, V.~Goswami,
  F.~Guzmán, P.~Koehn, A.~Mourachko, C.~Ropers, S.~Saleem, H.~Schwenk, and
  J.~Wang.
\newblock No language left behind: Scaling human-centered machine translation.
\newblock 2022.

\bibitem[Orlanski et~al.(2023)Orlanski, Xiao, Garcia, Hui, Howland, Malmaud,
  Austin, Singh, and Catasta]{orlanski2023measuring}
G.~Orlanski, K.~Xiao, X.~Garcia, J.~Hui, J.~Howland, J.~Malmaud, J.~Austin,
  R.~Singh, and M.~Catasta.
\newblock Measuring the impact of programming language distribution.
\newblock \emph{arXiv preprint arXiv:2302.01973}, 2023.

\bibitem[Paullada et~al.(2021)Paullada, Raji, Bender, Denton, and
  Hanna]{paullada2021data}
A.~Paullada, I.~D. Raji, E.~M. Bender, E.~Denton, and A.~Hanna.
\newblock Data and its (dis) contents: A survey of dataset development and use
  in machine learning research.
\newblock \emph{Patterns}, 2\penalty0 (11):\penalty0 100336, 2021.

\bibitem[Philip et~al.(2019)Philip, Namboodiri, and
  Jawahar]{philip2019baseline}
J.~Philip, V.~P. Namboodiri, and C.~Jawahar.
\newblock A baseline neural machine translation system for indian languages.
\newblock \emph{arXiv preprint arXiv:1907.12437}, 2019.

\bibitem[Post(2018)]{post-2018-call}
M.~Post.
\newblock A call for clarity in reporting {BLEU} scores.
\newblock In \emph{Proceedings of the Third Conference on Machine Translation:
  Research Papers}, pages 186--191, Brussels, Belgium, Oct. 2018. Association
  for Computational Linguistics.
\newblock \doi{10.18653/v1/W18-6319}.
\newblock URL \url{https://aclanthology.org/W18-6319}.

\bibitem[Pryzant et~al.(2017)Pryzant, Chung, Jurafsky, and
  Britz]{pryzant2017jesc}
R.~Pryzant, Y.~Chung, D.~Jurafsky, and D.~Britz.
\newblock Jesc: Japanese-english subtitle corpus.
\newblock \emph{arXiv preprint arXiv:1710.10639}, 2017.

\bibitem[Raffel et~al.(2020)Raffel, Shazeer, Roberts, Lee, Narang, Matena,
  Zhou, Li, and Liu]{raffel2020exploring}
C.~Raffel, N.~Shazeer, A.~Roberts, K.~Lee, S.~Narang, M.~Matena, Y.~Zhou,
  W.~Li, and P.~J. Liu.
\newblock Exploring the limits of transfer learning with a unified text-to-text
  transformer.
\newblock \emph{The Journal of Machine Learning Research}, 21\penalty0
  (1):\penalty0 5485--5551, 2020.

\bibitem[Sambasivan et~al.(2021)Sambasivan, Kapania, Highfill, Akrong,
  Paritosh, and Aroyo]{sambasivan2021everyone}
N.~Sambasivan, S.~Kapania, H.~Highfill, D.~Akrong, P.~Paritosh, and L.~M.
  Aroyo.
\newblock “everyone wants to do the model work, not the data work”: Data
  cascades in high-stakes ai.
\newblock In \emph{proceedings of the 2021 CHI Conference on Human Factors in
  Computing Systems}, pages 1--15, 2021.

\bibitem[Schwenk et~al.(2019)Schwenk, Chaudhary, Sun, Gong, and
  Guzm{\'a}n]{schwenk2019wikimatrix}
H.~Schwenk, V.~Chaudhary, S.~Sun, H.~Gong, and F.~Guzm{\'a}n.
\newblock Wikimatrix: Mining 135m parallel sentences in 1620 language pairs
  from wikipedia.
\newblock \emph{arXiv preprint arXiv:1907.05791}, 2019.

\bibitem[Sennrich et~al.(2016)Sennrich, Haddow, and
  Birch]{sennrich-etal-2016-improving}
R.~Sennrich, B.~Haddow, and A.~Birch.
\newblock Improving neural machine translation models with monolingual data.
\newblock In \emph{Proceedings of the 54th Annual Meeting of the Association
  for Computational Linguistics (Volume 1: Long Papers)}, pages 86--96, Berlin,
  Germany, Aug. 2016. Association for Computational Linguistics.
\newblock \doi{10.18653/v1/P16-1009}.
\newblock URL \url{https://aclanthology.org/P16-1009}.

\bibitem[Siddhant et~al.(2022)Siddhant, Bapna, Firat, Cao, Chen, Caswell, and
  Garcia]{siddhant-etal-2022-towards}
A.~Siddhant, A.~Bapna, O.~Firat, Y.~Cao, M.~X. Chen, I.~Caswell, and X.~Garcia.
\newblock Towards the next 1000 languages in multilingual machine translation:
  Exploring the synergy between supervised and self-supervised learning.
\newblock \emph{CoRR}, abs/2201.03110, 2022.
\newblock URL \url{https://arxiv.org/abs/2201.03110}.

\bibitem[Song et~al.(2019)Song, Tan, Qin, Lu, and Liu]{song2019mass}
K.~Song, X.~Tan, T.~Qin, J.~Lu, and T.-Y. Liu.
\newblock Mass: Masked sequence to sequence pre-training for language
  generation.
\newblock \emph{arXiv preprint arXiv:1905.02450}, 2019.

\bibitem[Tay et~al.(2022)Tay, Dehghani, Tran, Garcia, Bahri, Schuster, Zheng,
  Houlsby, and Metzler]{tay2022unifying}
Y.~Tay, M.~Dehghani, V.~Q. Tran, X.~Garcia, D.~Bahri, T.~Schuster, H.~S. Zheng,
  N.~Houlsby, and D.~Metzler.
\newblock Unifying language learning paradigms.
\newblock \emph{arXiv preprint arXiv:2205.05131}, 2022.

\bibitem[Tiedemann(2012)]{tiedemann2012parallel}
J.~Tiedemann.
\newblock Parallel data, tools and interfaces in opus.
\newblock In \emph{Lrec}, volume 2012, pages 2214--2218. Citeseer, 2012.

\bibitem[Vaswani et~al.(2017)Vaswani, Shazeer, Parmar, Uszkoreit, Jones, Gomez,
  Kaiser, and Polosukhin]{vaswani-etal-2017-attention}
A.~Vaswani, N.~Shazeer, N.~Parmar, J.~Uszkoreit, L.~Jones, A.~N. Gomez, L.~u.
  Kaiser, and I.~Polosukhin.
\newblock Attention is all you need.
\newblock In I.~Guyon, U.~V. Luxburg, S.~Bengio, H.~Wallach, R.~Fergus,
  S.~Vishwanathan, and R.~Garnett, editors, \emph{Advances in Neural
  Information Processing Systems}, volume~30. Curran Associates, Inc., 2017.
\newblock URL
  \url{https://proceedings.neurips.cc/paper/2017/file/3f5ee243547dee91fbd053c1c4a845aa-Paper.pdf}.

\bibitem[Vilar et~al.(2022)Vilar, Freitag, Cherry, Luo, Ratnakar, and
  Foster]{vilar2022prompting}
D.~Vilar, M.~Freitag, C.~Cherry, J.~Luo, V.~Ratnakar, and G.~Foster.
\newblock Prompting palm for translation: Assessing strategies and performance.
\newblock \emph{arXiv preprint arXiv:2211.09102}, 2022.

\bibitem[Weidinger et~al.(2021)Weidinger, Mellor, Rauh, Griffin, Uesato, Huang,
  Cheng, Glaese, Balle, Kasirzadeh, et~al.]{weidinger2021ethical}
L.~Weidinger, J.~Mellor, M.~Rauh, C.~Griffin, J.~Uesato, P.-S. Huang, M.~Cheng,
  M.~Glaese, B.~Balle, A.~Kasirzadeh, et~al.
\newblock Ethical and social risks of harm from language models.
\newblock \emph{arXiv preprint arXiv:2112.04359}, 2021.

\bibitem[Xue et~al.(2020)Xue, Constant, Roberts, Kale, Al-Rfou, Siddhant,
  Barua, and Raffel]{xue2020mt5}
L.~Xue, N.~Constant, A.~Roberts, M.~Kale, R.~Al-Rfou, A.~Siddhant, A.~Barua,
  and C.~Raffel.
\newblock mt5: A massively multilingual pre-trained text-to-text transformer.
\newblock \emph{arXiv preprint arXiv:2010.11934}, 2020.

\bibitem[Xue et~al.(2021)Xue, Constant, Roberts, Kale, Al-Rfou, Siddhant,
  Barua, and Raffel]{xue-etal-2021-mt5}
L.~Xue, N.~Constant, A.~Roberts, M.~Kale, R.~Al-Rfou, A.~Siddhant, A.~Barua,
  and C.~Raffel.
\newblock m{T}5: A massively multilingual pre-trained text-to-text transformer.
\newblock In \emph{Proceedings of the 2021 Conference of the North American
  Chapter of the Association for Computational Linguistics: Human Language
  Technologies}, pages 483--498, Online, June 2021. Association for
  Computational Linguistics.
\newblock \doi{10.18653/v1/2021.naacl-main.41}.
\newblock URL \url{https://aclanthology.org/2021.naacl-main.41}.

\bibitem[Ye et~al.(2018)Ye, Devendra, Matthieu, Sarguna, and
  Graham]{Ye2018WordEmbeddings}
Q.~Ye, S.~Devendra, F.~Matthieu, P.~Sarguna, and N.~Graham.
\newblock When and why are pre-trained word embeddings useful for neural
  machine translation.
\newblock In \emph{HLT-NAACL}, 2018.

\bibitem[Zhang et~al.(2020)Zhang, Williams, Titov, and
  Sennrich]{zhang2020improving}
B.~Zhang, P.~Williams, I.~Titov, and R.~Sennrich.
\newblock Improving massively multilingual neural machine translation and
  zero-shot translation.
\newblock \emph{arXiv preprint arXiv:2004.11867}, 2020.

\bibitem[Zhang et~al.(2023)Zhang, Haddow, and Birch]{zhang2023prompting}
B.~Zhang, B.~Haddow, and A.~Birch.
\newblock Prompting large language model for machine translation: A case study.
\newblock \emph{arXiv preprint arXiv:2301.07069}, 2023.

\bibitem[Zhu et~al.(2015)Zhu, Kiros, Zemel, Salakhutdinov, Urtasun, Torralba,
  and Fidler]{zhu2015aligning}
Y.~Zhu, R.~Kiros, R.~Zemel, R.~Salakhutdinov, R.~Urtasun, A.~Torralba, and
  S.~Fidler.
\newblock Aligning books and movies: Towards story-like visual explanations by
  watching movies and reading books.
\newblock In \emph{Proceedings of the IEEE international conference on computer
  vision}, pages 19--27, 2015.

\bibitem[Ziemski et~al.(2016)Ziemski, Junczys-Dowmunt, and
  Pouliquen]{ziemski2016united}
M.~Ziemski, M.~Junczys-Dowmunt, and B.~Pouliquen.
\newblock The united nations parallel corpus v1. 0.
\newblock In \emph{Proceedings of the Tenth International Conference on
  Language Resources and Evaluation (LREC'16)}, pages 3530--3534, 2016.

\end{thebibliography}
\newpage
\appendix

% \part{Appendix} % Start the appendix part
% \parttoc % Insert the appendix TOC

\section{Appendix}

\subsection{LangID Details}\label{app:langid}

Following \textit{Language Id In the Wild} \citep{caswell2020language}, we
trained a Transformer-Base \citep{vaswani-etal-2017-attention} Semi-Supervised LangId model (SSLID) on 498 languages. The training
data is as described in \textit{Language ID in the Wild}, with the differences that 1) training data
is sampled to a temperature of \texttt{T=3} to reduce over-triggering on low-resource
languages; and 2) the data is supplemented with web-crawled data from the same
paper (that has already been through the various filters described therein). The purpose of adding this data is to increase robustness to web-domain text, and possibly distill some of the filters used to create the web-crawl. The languages chosen for this model were roughly the top 498 by number of sentences in the dataset reported by \textit{Language ID in the Wild}. The complete list may be seen in Table \ref{tab:madlad-full}.

% {\fontsize{6.4}{8}\selectfont

% \newcommand{\insertbigboi}{
{
\fontsize{6.4}{8}\selectfont {
\begin{longtable}{@{}lllrrrrrr@{}}

\caption{\label{tab:madlad-full} BCP-47 codes, name, script and amount of data associated with all languages in \data. The last 79 languages, with entries of "-", were languages that the LangID model was trained to detect, but was omitted after the self-audit.} \\

\toprule \\
BCP-47 & Name & Script & docs (noisy) & docs (clean) & sents (noisy) & sents (clean) & chars (noisy) & chars (clean) \\
\midrule \\
        %  \endhead
         
% \toprule
total & - & - & 7.8B & 4B & 148.4B & 105.5B & 33.3T & 18.3T \\
median & - & - & 21.6K & 1.5K & 202.3K & 61.8K & 63.6M & 8.5M \\
\midrule \\
en & English & Latn & 3.6B & 1.8B & 87.9B & 53.4B & 15T & 9T \\
\end{longtable}

}
}
% \insertbigboi

\subsection{Filtering Details}
\label{app:filtering}

\paragraph{Cursed Substrings} Following is the list of cursed substrings that we used to filter the monolingual data. Here are a few general notes about these strings:

\begin{enumerate}
  \item low quality sentences ending in the pipe character were very common. (Note: this was not Devanagari-script text using a Danda.)
  \item The last few regexes are meant to match \texttt{A N T S P E A K}, \textit{List Case}, and
weirdly regular text (for instance, lists of shipping labels or country codes)
\end{enumerate}

Here is the complete list of cursed substrings and cursed regexes, along with the function used for filtering:
% CURSED_SUBSTRINGS = [" №", "���", "\\|\\s*$", " nr\\.$", "aute irure dolor ", " sunt in culpa qui ", "orem ipsum ", " quis nostrud ", " adipisicing ", " dolore eu ", " cupidatat ", "autem vel eum", "wisi enim ad", " sex ", " porn ", "黄色电影", "mp3", "ownload", "Vol\\.", " Ep\\.", "Episode", " г\\.\\s*$", " кг\\.\\s*$", " шт\\.", "Develop", "Facebook", " crusher ", " xxx ", " ... ... ... ... ... ... ... ... ...", " .... .... .... .... .... .... .... .... ....", " [^ ] [^ ] [^ ] [^ ] [^ ] [^ ] [^ ] [^ ] [^ ]", ", ..,,? ..,,? ..,,? ..,,?"]
\begin{verbatim}
# this implementation is for demonstration and not very efficient;
# to speed it up, use string inclusion (`in`) instead of regex for
# all but the last four, and for those use a compiled regex.
def is_cursed(s):
  return any(re.findall(curse, s) in s for curse in CURSED_SUBSTRINGS)


CURSED_SUBSTRINGS = [' \u2116', '\ufffd\ufffd\ufffd', '\\|\\s*$', ' nr\\.$', 
'aute irure dolor ', ' sunt in culpa qui ', 'orem ipsum ', ' quis nostrud ',
' adipisicing ', ' dolore eu ', ' cupidatat ', 'autem vel eum', 'wisi enim ad',
' sex ', ' porn ', '\u9ec4\u8272\u7535\u5f71', 'mp3', 'ownload',
'Vol\\.', ' Ep\\.', 'Episode', ' \u0433\\.\\s*$', ' \u043a\u0433\\.\\s*$',
' \u0448\u0442\\.', 'Develop', 'Facebook', ' crusher ', ' xxx ',
' ... ... ... ... ... ... ... ... ...',
' .... .... .... .... .... .... .... .... ....',
' [^ ] [^ ] [^ ] [^ ] [^ ] [^ ] [^ ] [^ ] [^ ]',
', ..,,? ..,,? ..,,? ..,,?',
]
\end{verbatim}

\paragraph{Virama Correction} Below is the virama substitution code: 

\begin{verbatim}
    
_VIRAMA_CHARS = (
'\u094d\u09cd\u0a4d\u0acd\u0b4d\u0bcd\u0c4d\u0ccd\u0d3b'
'\u0d3c\u0d4d\u0dca\u0e3a\u0eba\u0f84\u1039\u103a\u1714'
'\u1734\u17d2\u1a60\u1b44\u1baa\u1bab\u1bf2\u1bf3\u2d7f'
'\ua806\ua82c\ua8c4\ua953\ua9c0\uaaf6\uabed\u10a3f\u11046'
'\u1107f\u110b9\u11133\u11134\u111c0\u11235\u112ea\u1134d'
'\u11442\u114c2\u115bf\u1163f\u116b6\u1172b\u11839\u1193d'
'\u1193e\u119e0\u11a34\u11a47\u11a99\u11c3f\u11d44\u11d45'
'\u11d97\u1031\u1057\u1058\u1059\u1056\u1060\u1062\u1068'
'\u1063\u1067\u1068\u1069\u105e\u105f\u1036\u103d\u102d'
'\u102f\u102e\u102d\u1030\u1033\u1034\u1035\u102f\u1032'
'\u102c\u103c\u103d\u103e\u102b\u1037\u1038\u25cc\u25cc'
'\u000a\u1071\u1072\u1073\u1074\u1082\u1083\u1084\u1085'
'\u1086\u1087\u1088\u1089\u108a\u108b\u108c\u108d\u108f'
'\u109a\u109b\u109c\u109d\ua9e5\uaa7b\uaa7c\uaa7d'
)


def remove_viramas(x: str) -> str:
  return '%s' % regex.sub(r' ([%s]) ' % _VIRAMA_CHARS, '\\1', x)

\end{verbatim}

% zh_pornsignals = "caoporn caoprom caopron caoporen caoponrn caoponav caopom caoorn 99re dy888 caopro hezyo re99 4438x zooskool xfplay 7tav xxoo xoxo 52av freexx 91chinese anquye cao97 538porm 87fuli 91pron 91porn 26uuu 4438x 182tv kk4444 777me ae86 91av 720lu yy6080 6080yy qqchub paa97 aiai777 yy4480 videossexo 91free 一级特黄大片 偷拍久久国产视频 日本毛片免费视频观看 久久免费热在线精品 高清毛片在线看 日本毛片高清免费视频 一级黄色录像影片 亚洲男人天堂 久久精品视频在线看 自拍区偷拍亚洲视频 亚洲人成视频在线播放 色姑娘综合站 丁香五月啪啪 在线视频成人社区 亚洲人成视频在线播放 久久国产自偷拍 一本道 大香蕉无码 香港经典三级 亚洲成在人线免费视频 天天色综合网 大香蕉伊人久草 欧美一级高清片 天天鲁夜夜啪视频在线 免费黄片视频在线观看 加比勒久久综合 久草热久草在线视频 韩国三级片大全在线观看 青青草在线视频 美国一级毛片 久草在线福利资源 啪啪啪视频在线观看免费 成人福利视频在线观看 婷婷我去也 老司机在线国产 久久成人视频 手机看片福利永久国产 高清国产偷拍在线 大香蕉在线影院 日本高清免费一本视频 男人的天堂东京热 影音先锋男人资源 五月婷婷开心中文字幕 亚洲香蕉视频在线播放 天天啪久久爱视频精品 超碰久久人人摸人人搞".split()

\paragraph{Chinese Porn Filter} Below is the Chinese porn filter list:

\begin{verbatim}
zh_pornsignals = [
'caoporn', 'caoprom', 'caopron', 'caoporen', 'caoponrn', 'caoponav', 'caopom',
'caoorn', '99re', 'dy888', 'caopro', 'hezyo', 're99', '4438x', 'zooskool',
'xfplay', '7tav', 'xxoo', 'xoxo', '52av', 'freexx', '91chinese', 'anquye',
'cao97', '538porm', '87fuli', '91pron', '91porn', '26uuu', '4438x', '182tv',
'kk4444', '777me', 'ae86', '91av', '720lu', 'yy6080', '6080yy', 'qqchub',
'paa97', 'aiai777', 'yy4480', 'videossexo', '91free',
'\u4e00\u7ea7\u7279\u9ec4\u5927\u7247',
'\u5077\u62cd\u4e45\u4e45\u56fd\u4ea7\u89c6\u9891',
'\u65e5\u672c\u6bdb\u7247\u514d\u8d39\u89c6\u9891\u89c2\u770b',
'\u4e45\u4e45\u514d\u8d39\u70ed\u5728\u7ebf\u7cbe\u54c1',
'\u9ad8\u6e05\u6bdb\u7247\u5728\u7ebf\u770b',
'\u65e5\u672c\u6bdb\u7247\u9ad8\u6e05\u514d\u8d39\u89c6\u9891',
'\u4e00\u7ea7\u9ec4\u8272\u5f55\u50cf\u5f71\u7247',
'\u4e9a\u6d32\u7537\u4eba\u5929\u5802',
'\u4e45\u4e45\u7cbe\u54c1\u89c6\u9891\u5728\u7ebf\u770b',
'\u81ea\u62cd\u533a\u5077\u62cd\u4e9a\u6d32\u89c6\u9891',
'\u4e9a\u6d32\u4eba\u6210\u89c6\u9891\u5728\u7ebf\u64ad\u653e',
'\u8272\u59d1\u5a18\u7efc\u5408\u7ad9',
'\u4e01\u9999\u4e94\u6708\u556a\u556a',
'\u5728\u7ebf\u89c6\u9891\u6210\u4eba\u793e\u533a',
'\u4e9a\u6d32\u4eba\u6210\u89c6\u9891\u5728\u7ebf\u64ad\u653e',
'\u4e45\u4e45\u56fd\u4ea7\u81ea\u5077\u62cd',
'\u4e00\u672c\u9053',
'\u5927\u9999\u8549\u65e0\u7801',
'\u9999\u6e2f\u7ecf\u5178\u4e09\u7ea7',
'\u4e9a\u6d32\u6210\u5728\u4eba\u7ebf\u514d\u8d39\u89c6\u9891',
'\u5929\u5929\u8272\u7efc\u5408\u7f51',
'\u5927\u9999\u8549\u4f0a\u4eba\u4e45\u8349',
'\u6b27\u7f8e\u4e00\u7ea7\u9ad8\u6e05\u7247',
'\u5929\u5929\u9c81\u591c\u591c\u556a\u89c6\u9891\u5728\u7ebf',
'\u514d\u8d39\u9ec4\u7247\u89c6\u9891\u5728\u7ebf\u89c2\u770b',
'\u52a0\u6bd4\u52d2\u4e45\u4e45\u7efc\u5408',
'\u4e45\u8349\u70ed\u4e45\u8349\u5728\u7ebf\u89c6\u9891',
'\u97e9\u56fd\u4e09\u7ea7\u7247\u5927\u5168\u5728\u7ebf\u89c2\u770b',
'\u9752\u9752\u8349\u5728\u7ebf\u89c6\u9891',
'\u7f8e\u56fd\u4e00\u7ea7\u6bdb\u7247',
'\u4e45\u8349\u5728\u7ebf\u798f\u5229\u8d44\u6e90',
'\u556a\u556a\u556a\u89c6\u9891\u5728\u7ebf\u89c2\u770b\u514d\u8d39',
'\u6210\u4eba\u798f\u5229\u89c6\u9891\u5728\u7ebf\u89c2\u770b',
'\u5a77\u5a77\u6211\u53bb\u4e5f',
'\u8001\u53f8\u673a\u5728\u7ebf\u56fd\u4ea7',
'\u4e45\u4e45\u6210\u4eba\u89c6\u9891',
'\u624b\u673a\u770b\u7247\u798f\u5229\u6c38\u4e45\u56fd\u4ea7',
'\u9ad8\u6e05\u56fd\u4ea7\u5077\u62cd\u5728\u7ebf',
'\u5927\u9999\u8549\u5728\u7ebf\u5f71\u9662',
'\u65e5\u672c\u9ad8\u6e05\u514d\u8d39\u4e00\u672c\u89c6\u9891',
'\u7537\u4eba\u7684\u5929\u5802\u4e1c\u4eac\u70ed',
'\u5f71\u97f3\u5148\u950b\u7537\u4eba\u8d44\u6e90',
'\u4e94\u6708\u5a77\u5a77\u5f00\u5fc3\u4e2d\u6587\u5b57\u5e55',
'\u4e9a\u6d32\u9999\u8549\u89c6\u9891\u5728\u7ebf\u64ad\u653e',
'\u5929\u5929\u556a\u4e45\u4e45\u7231\u89c6\u9891\u7cbe\u54c1',
'\u8d85\u78b0\u4e45\u4e45\u4eba\u4eba\u6478\u4eba\u4eba\u641e',
]

\end{verbatim}

\subsection{Other issues fixed after the self-audit}\label{app:other-issues}

\paragraph{Consulting Language Speakers} For a few languages, we had strong suspicions that the text was noisy or spurious, but were unable to acertain the quality of the data. In these cases we asked a native speaker to audit the data. Based on their recommendations, we did the following:

\begin{enumerate}[nolistsep]%[leftmargin=*]
  \item \texttt{zh}, \texttt{zh\_Latn}: This resulted in the special filters described below.
  \item \texttt{en\_Arab}, \texttt{tly\_IR}: This data was found to boilerplate, so we removed this data.
  \item \texttt{fa}, \texttt{bho}: No changes were made.
\end{enumerate}

\paragraph{Language Renames and Merges} For several languages, we found that (mostly by checking URLs) the corpora were in languages different from the LangID predictions. This led to the following changes:

\begin{enumerate}[nolistsep]%[leftmargin=*]
  \item dty renamed to \texttt{zxx-xx-dtynoise}, aka a ``language'' of noise. This is mainly mis-rendered PDFs and may have practical applications for denoising, or for decoding such garbled PDFs.
  \item \texttt{fan} renamed to \texttt{bum}
  \item \texttt{cjk} merged into the \texttt{gil} dataset
  \item \texttt{bjj} merged into the \texttt{awa} dataset
  \item \texttt{ss-SZ} renamed to \texttt{ss} -- this was a result of inconsistent data labels.
\end{enumerate}

\subsection{Monolingual Data Details}\label{app:mono}

Notes from rounds 2 and 3 of the self-audit can be seen in Table \ref{tab:audit-full}. Some of these notes may refer to previous, less filtered versions of the data, especially those with a ``r1'' (meaning ``round 1''). Some of them however do have some useful information about quirks of problems with the current dataset. The overall statistics of \data are in Table \ref{tab:madlad-full}.

% {\fontsize{6.4}{8}\selectfont

% \newcommand{\insertauditnotes}{
{
\fontsize{6.4}{8}\selectfont {
% \resizebox{\columnwid}{}{}
\setlength\LTleft{0pt}
\setlength\LTright{0pt}
% \resizebox{\columnwidth}{!}{
\begin{longtable}{@{\extracolsep{\fill}}p{1.75cm}p{11.25cm}@{}}

\caption{\label{tab:audit-full} Notes that we made about individual samples while auditing them. Some languages have notes from the earlier round of auditing in parentheses, e.g. '(r1: get this checked by Hindi speaker)'. Notes from Round 0, which were used to find cursed substrings, were not kept.} \\
% {\fontsize{6.4}{8}\selectfont
% \begin{xltabular}[H]{\linewidth}{lp{4in}}
\toprule \\
%     \centering
BCP-47 & notes \\ 
\midrule \\
aa & some pretty bad data but also some good data. filter on "Woo" (case sensitive) (r1: ok) \\
abs & all short nonsense remove (r1: ok) \\
abt & fine; bible (r1: ok) \\
ace & good; bible (r1: ok) \\
acf & good; bible (r1: ok) \\
ach & good; bible (r1: ok) \\
ada & good; bible; likely mixed with gaa (r1: ok but odd character usage LATIN CAPITAL LETTER OPEN O when it should be lower case in the middle of words) \\
adh & good; bible (r1: ok, lots bible) \\
ady & good (r1: ok but weird boilerplate) \\
af & good (r1: ok) \\
agr & good; bible (r1: ok; some AL in Arabic script) \\
ahk & good; bible; crazy diacritics (r1: ok but weird: lots of u748 MODIFIER LETTER VOICING) \\
ak & good; much but not all bible (r1: ok) \\
akb & good; bible (r1: empty) \\
alt & WAIT THIS IS AMAZING IT IS ACTUALLY ALTAI! e.g. from urls like https://altaicholmon.ru/2020/02/28/jarashty-la-jajaltany-jarkyndu-lekeri/ (r1: ok but there are just lots of numbers...not very clean) \\
alz & good; bible (r1: ok; bible) \\
am & good (r1: ok) \\
amu & good; bible; crazy diacritics (r1: empty) \\
ang & much noise but some good Old English in there! (r1: ok; wikipedia; one document that is just "lastfootwear.com" 1M times) \\
ann & good; all from wikimedia incubator (r1: ok) \\
apd-SD & terribly questionable; probably remove (r1: maybe ok, but looks like lots of template....maybe remove) \\
ape & good; bible (r1: remove) \\
ar & good (r1: ok) \\
ar-Latn & terrible, 0pct correct, remove (r1: remove) \\
arn & good; bible (r1: ok) \\
as & good (r1: ok) \\
av & good (r1: ok) \\
awa & OK; should be used with caution and suspicion (r1: remove) \\
awa & all bible in awadhi (awa). Renamed from bjj (r1: remove) \\
ay & good; mix of bible and other news sources (r1: ok but very noisy) \\
ayl & remove. not ayl. (r1: uh this is all Arabic with "homo" in English...remove?) \\
az & good (r1: ok) \\
az-RU & good; a lot of JW (r1: ok) \\
azg & 70pct short noise; 30pct good bible (r1: empty) \\
ba & ok (r1: ok) \\
ban & ok bible (r1: ok) \\
bas & ok; has some fun blog stuff! (r1: empty) \\
bbc & ok (r1: ok) \\
bci & ok bible (r1: ok; bible) \\
be & ok (r1: ok) \\
ber & ok great! (r1: ok; Mixed in French) \\
ber-Latn & ok (r1: ok) \\
bew & mostly blogs. i have no way of knowing if this is standard indonesian or not (r1: ok; noisy) \\
bfy & very bad. remove unless it looks better after filtering short docs; remove (r1: remove) \\
bg & ok (r1: ok) \\
bg-Latn & ok (r1: ok but questionable...Slavic speaker review needed) \\
bgc & super sketch. Remove unless short doc filter leaves some. remove  (r1: very questionable....Hindi speaker review) \\
bgp & almost all ur-Latn. consider removing or renaming (r1: very questionable. Remove? mainly ur-Latn) \\
bgz & idk maybe ok but probably bad (r1: remove. Wow, this is amazing. It is in all sorts of languages -- the only thing they share is that they each have like 500 question marks) \\
bhb-Gujr & bad. remove. all junk gu. (r1: remove; great noise?) \\
bho & mostly from anjoria.com. Ankur reviews and says that it looks like valid Bhojpuri for the most part (r1: questionable but ok?) \\
bi & good! fun! (r1: ok) \\
bik & ok. keep in mind the bik vs bcl issue. (r1: ok) \\
bim & good; bible (r1: empty) \\
bm & good (r1: ok but these headers are LONG) \\
bmm & terrible. filter on short and reevaluate (r1: remove) \\
bn & ok (r1: ok) \\
bn-Latn & ok (r1: ok) \\
bo & needs some serious script filtering. but there is some ok data in there. (r1: ok) \\
bqc & ok; bible (r1: ok but too short?) \\
br & ok after shortfilter (r1: ok) \\
bru & ok; bible (r1: ok) \\
brx & quite good! (r1: ok but questionable...Hindi speaker review?) \\
bs & good (r1: ok) \\
bto & bad; remove unless short filter keeps enough (r1: empty) \\
bts & ok; mostly bible (r1: ok) \\
btx & ok probably (r1: ok) \\
bua & ok (r1: ok) \\
bum & ok bible; but technically wrong language. Data is in Bulu, not Fang, though they are closely related, so ranamed from "fan" \\
bus & ok; bible; about 50bzc (r1: ok bible) \\
bzj & ok bible (r1: ok) \\
ca & ok (r1: ok i guess....but is it actually italian?) \\
cab & ok jw (r1: ok) \\
cac & ok bible (r1: ok bible) \\
cak & ok bible (r1: ok bible) \\
cbk & ok bible; not Spanish (r1: remove; all Spanish) \\
cce & ok jw (r1: empty) \\
ce & ok (r1: ok) \\
ceb & ok (r1: ok) \\
cfm & ok mostly from chinland.co (r1: ok) \\
cgg & rather noisy but potentialy ok. not sure if WL or not (r1: ok) \\
ch & ok; not sure about WL (r1: ok) \\
chk & ok bible (r1: ok bible) \\
chm & ok; fyi watch out for yandex translationese (r1: ok) \\
chr & ok bible (r1: ok) \\
ckb & ok (r1: ok) \\
clu & ok bible (r1: ok) \\
cnh & good, some local news! not sure if WL (r1: ok) \\
cni & ok; bible; lots of mixed in content in not,cob,cpc,arl (r1: ok) \\
co & ok;l i suspect lots of MT (r1: ok i guess?) \\
cr-Latn & noise and lorem ipsom. But some ok Cree text. (r1: mostly Lorem Ipsom. remove? Or release with note? there is some plausible stuff here too.) \\
crh & ok (r1: ok but review with russian speaker as it could be russian....) \\
crs & ok (r1: ok) \\
cs & ok (r1: ok) \\
ctd-Latn & ok; from some local news? (r1: ok) \\
ctg & probably terrible probably remove (r1: very questionable....remove?) \\
ctu & ok bible (r1: ok bible) \\
cuk & ok bible (r1: ok bible) \\
cv & good (r1: ok) \\
cy & ok after shortfilter; OK (r1: ok) \\
cyo & terrifying noise; remove (r1: empty) \\
da & ok (r1: ok) \\
dcc & remove (r1: empty) \\
de & ok (r1: ok) \\
din & ok after short doc filter (r1: ok but LONG headers uh oh) \\
dje & ok; mostly but not all bible (r1: ok bible) \\
djk & ok; bible+jw (r1: empty) \\
dln & ok bible (r1: empty) \\
doi & ok actually nice! (r1: sus; review by hindi speaker) \\
dov & ok bible + jw (r1: ok) \\
dtp & ok; mostly from www.newsabahtimes.com.my (r1: ok) \\
dv & good (r1: ok) \\
dwr & ok; bible; mixed script (r1: empty) \\
dyu & ok bible (r1: empty) \\
dz & ok; hidden parallel text; maybe actually bo; mainly buddhist (r1: ok; mixed dz-Latn) \\
ee & good; mostly religious (r1: ok bible) \\
el & ok (r1: ok) \\
el-CY & bad (r1: v suspicious; mainly comma lists or boilerplate; remove) \\
el-Latn & good; a lot of old content! (r1: ok) \\
emp & ok bible (r1: ok) \\
en & ok (r1: ok) \\
en-Arab & Ali reviewed; this is not good data. remove. (r1: idk review w/Arabic reader) \\
en-Cyrl & ok ... some fr-Cyrl too and maybe others (r1: OMG LOL yes ok) \\
enq & ok bible (r1: ok bible) \\
eo & ok; likely a lot of MT (r1: ok) \\
es & good (r1: ok) \\
et & ok (r1: ok) \\
eu & ok (r1: ok; lots of poetry?) \\
fa & consulted Ali; he says it's ok (r1: ok) \\
ff & ok after shortfilter (r1: some noise but some nice stuff! ok!) \\
ff-Adlm & good (r1: ok sweet) \\
ffm & ok bible; mixed fulfulde dialects; consider mergind with ff (r1: ok but idk the dialect) \\
fi & ok (r1: ok but lotsa headers) \\
fil & ok more bible than expected for such a major language (r1: ok pls note in release that this is the same as tl) \\
fip & ok jw ; but wrong language. mostly Mambwe-Lungu and Bemba, not Fipu (mgr+bem vs fip) (r1: ok bible) \\
fj & ok (r1: ok bible lotsa noise) \\
fo & good (r1: ok TODO check that this is not icelandic review) \\
fon & ok mostly jw but not all (r1: ok bible) \\
fr & ok (r1: ok) \\
frp & fair amount from wikipedia. (r1: remove; all noise + hashtags) \\
fy & ok plausible but i bet there is a lot of Dutch in there (r1: ok) \\
ga & ok some en noise (r1: ok) \\
gag & has 1-2 cyrillic examples with small amts of arabic script noise (r1: ok) \\
gbm & ok (r1: ok) \\
gd & ok (r1: ok; but barely) \\
gil & empty; but merged in data in "cjk" (r1: empty) \\
gil & this is all in gil (Kiribati). merged into "gil" (r1: empty) \\
gjk & empty remove (r1: empty) \\
gju & remove short boilerplate (r1: empty) \\
gl & ok (r1: ok) \\
gn & ok some broken characters some bible (r1: ok) \\
gof & ok some bible (r1: empty) \\
gom & ok (r1: ok) \\
gom-Latn & filter on really short boilerplate in en; some porn; after: ok very noisy ; some ok stuff ; release with disclaimer (r1: ok) \\
gor & ok bible (r1: ok) \\
grc & warning: this is likely polyphonic greek, not ancient greek (r1: ok but idk diff between ancient and modern greek) \\
gsw & wtf is happening here; keep with disclaimer; STILL BOILERPLATE (r1: ok but idk diff between gsw and de) \\
gu & ok (r1: ok some en boilerplate) \\
gu-Latn & filter short en boillerplate and repetitive sentences (r1: lots of social media pages and some porn) \\
gub & ok bible (r1: empty) \\
guc & ok bible (r1: ok) \\
guh & ok bible (r1: ok) \\
gui & ok bible (r1: ok) \\
gv & filter short repetitive sentenecs; still same but keep (r1: ok) \\
gvl & filter short boilerplate mostly bible (r1: ok) \\
gym & ok biblle (r1: ok) \\
gyn & remove boilerplate and porn (r1: remove) \\
ha & ok (r1: ok) \\
haw & ok scam tv products (r1: ok but filter u65533 REPLACEMENT CHARACTER) \\
hi & ok some porn (r1: ok but some en boilerplate) \\
hi-Latn & filter porn this is half porn (r1: ok but some hi and en) \\
hif & ok some en noise and religious (r1: ok it is in Latin) \\
hil & ok some en boilerplate (r1: ok) \\
hmn & ok (r1: ok) \\
hne & ok (r1: ok) \\
ho & ok (r1: ok but but split between wiki boilerplate and actual content) \\
hr & ok (r1: ok) \\
ht & ok (r1: ok) \\
hu & ok (r1: ok) \\
hui & ok some bible (r1: ok bible) \\
hus & ok bible (r1: some wiki boilerplate) \\
hvn & ok religioous text (r1: ok bible) \\
hy & ok (r1: ok) \\
iba & ok jw data (r1: ok) \\
ibb & ok bible and repeated @ (r1: ok but bible and some repeated lines) \\
id & ok (r1: ok) \\
ify & ok bible (r1: empty) \\
ig & ok (r1: ok) \\
ilo & ok some bible (r1: some repetitive content) \\
inb & ok bible (r1: remove; it's a single bible doc lol) \\
is & ok (r1: ok) \\
iso & ok jw (r1: ok) \\
it & ok (r1: ok) \\
iu & filter script some is en rest is iu script (r1: ok filter latin script) \\
ium & filter out zh (r1: remove mostly en) \\
iw & ok (r1: ok has some codemixing because of boilerplate) \\
izz & ok bible (r1: empty) \\
ja & ok a little en mixed in (r1: ok but some porn) \\
ja-Latn & remove maybe low quality short and repeated (r1: ok some noise that is manga pages in english) \\
jac & ok bible (r1: remove 'home loan' repeated over and over) \\
jam & ok bible (r1: ok) \\
jax & filter mostly text.medjugorje.ws boilerplate (r1: remove) \\
jiv & ok bible \\
jv & ok (r1: ok) \\
jvn & ok bible (r1: ok) \\
ka & ok (r1: ok) \\
kaa & ok (FYI cyrllic) (r1: ok) \\
kaa-Latn & ok urls are .ru or .kz (r1: ok) \\
kac & ok (r1: ok) \\
kbd & ok many .ru (r1: ok some repetitive text and en noise) \\
kbp & not sure if right script wiki says latin (r1: ok) \\
kek & ok jw bible (r1: ok bible) \\
kfy & filter virama issue (r1: ok) \\
kg & ok bible jw (r1: ok) \\
kha & ok (r1: ok some repetitive boilerplate) \\
kj & ok (r1: filter english out) \\
kjb & ok bible (r1: empty) \\
kjg & ok bible (r1: empty) \\
kjh & ok .ru domain (r1: ok) \\
kk & ok (r1: ok) \\
kl & ok (r1: ok) \\
km & ook (r1: ok) \\
kmb & ok bible jw (r1: ok) \\
kmz-Latn & ok soome ar script noise (r1: ok) \\
kn & ok (r1: ok) \\
kn-Latn & filter en noise of karnataka govt websites (r1: filter porn there is too much porn and repetitive content) \\
knj & ok bible (r1: empty) \\
ko & ok (r1: ok) \\
koi & ok (r1: ok) \\
kos & ok lds bible (r1: ok bible) \\
krc & ok (r1: ok some repetitive content) \\
kri & ok boilerplate noise bible jw (r1: remove repetitive) \\
ks & ok shorter docs (r1: ok) \\
ksd & ok bible (r1: ok bible) \\
ksw & ok bible (r1: ok) \\
ktu & ok bible jw (r1: ok) \\
ku & ok (r1: ok) \\
kum & ok (r1: ok) \\
kv & ok a lil boilerplate vibes (r1: ok) \\
kw & ok short boilerplate bible wiki; ok some porn (r1: ok filter english) \\
kwi & ok bible (r1: ok) \\
ky & ok (r1: ok) \\
la & ok some broken chars \\
laj & ok bible \\
lb & ok shorter text; ok AFTER  \\
lg & ok lot of www.bukedde.co.ug in this \\
lhu & ok bible \\
ln & ok bible jw \\
lo & ok many entities in latin script \\
lrc & ok \\
lt & ok \\
ltg & ok mostly www.lakuga.lv \\
lu & ok jw \\
lus & ok \\
luz & terrible; remove \\
lv & ok \\
mad & remove mostly short text \\
mag & ok fix virama issue \\
mai & ok mild amounts of en noise \\
mak & ok bible \\
mam & ok bible jw \\
mas & ok some amount of bible \\
max & remove short some ru \\
maz & ok bible jw \\
mbt & ok bible \\
mdf & ok some short docs \\
mdh & filter porn short text and repetitive boilerplate \\
mdy & ok bible \\
mel & remove noisy en \\
meo & ok mostly blogs \\
meu & ok bible \\
mey & mostly short and noisy borderline \\
mfb & remove short boilerplate \\
mfe & ok mostly bible maybe some french creole short doc noise \\
mg & ok some bible jw \\
mgh & ok bible jw \\
mh & ok jw lds \\
mi & ok \\
min & ok mostly wiki and bible \\
miq & ok \\
mk & ok \\
mkn & ok bible \\
ml & ok \\
ml-Latn & ok some short docs \\
mn & ok \\
mni & ok \\
mnw & remove en noise and boilerplate \\
mps & ok bible \\
mqy & bible remove short docs \\
mr & ok fix virama \\
mr-Latn & remove mostly porn and short docs \\
mrj & remove short docs; ok \\
mrw & ok remove short docs \\
ms & ok \\
ms-Arab & ok mostly utusanmelayu website \\
ms-Arab-BN & ok not sure if same as ms-Arab \\
msb & ok bible \\
msi & ok  filter short docs \\
msm & ok bible \\
mt & ok \\
mtq & remove short doc repetitive \\
mtr & ok fix virama remove en noise \\
mui & remove short docs \\
mwr & filter short docs fix virama \\
my & filter noise and en fix virama \\
myv & maybe has .ru urls \\
nan-Latn-TW & ok \\
nd & ok \\
ndc-ZW & ok \\
ne & ok \\
new & ok \\
ng & ok \\
ngu & ok \\
nhe & ok \\
nia & ok \\
nij & ok \\
niq & ok \\
nl & ok \\
nnb & ok \\
no & ok \\
noa & ok \\
noe & ok \\
nog & ok \\
nr & ok \\
nso & ok \\
nut & ok \\
nv & ok \\
ny & ok \\
nyn & ok \\
nyo & ok \\
nyu & ok \\
nzi & ok \\
oc & ok \\
oj & ok \\
om & ok \\
or & ok \\
os & ok \\
otq & ok \\
pa & ok \\
pa-Arab & ok \\
pag & bible \\
pam & remove \\
pap & ok \\
pau & ok \\
pck & ok \\
pcm & ok \\
pis & bible \\
pl & ok \\
pmy & remove \\
pon & bible \\
ppk & bible \\
prk & ok \\
ps & ok \\
pt & ok \\
qu & ok \\
qub & bible \\
quc & bible \\
quf & bible \\
quh & bible \\
qup & bible \\
quy & bible \\
qvc & bible \\
qvi & bible \\
qvz & bible \\
qxr & bible \\
raj & ok \\
rcf & ok \\
rhg-Latn & remove \\
rki & ok \\
rkt & ok \\
rm & ok \\
rmc & ok \\
rn & bible \\
ro & ok \\
rom & bible \\
ru & ok \\
ru-Latn & ok \\
rw & ok \\
rwo & bible \\
rwr & remove  \\
sa & ok \\
sah & ok \\
sat-Latn & good! all from local news sources \\
sd & good \\
sda & ok bible \\
se & good \\
seh & ok jw \\
sg & ok jw \\
sgj & remove \\
shn & mostly English boilerplate. filter by latin text before releasing \\
shp & ok bible \\
shu & quite questionable. prob remove \\
si & good \\
sja & ok bibe \\
sjp & terible; probably remove; check again after short filter \\
sk & ok \\
skg & terrible; remove \\
skr & ok; some pnb mixed in \\
sl & ok \\
sm & ok \\
smt & ok bible but lots of different bibles! \\
sn & ok \\
so & good \\
spp & ok bible \\
sq & good \\
sr & ok \\
srm & ok; bible + jw \\
srn & ok bible + jw \\
srr & remove; englishboilerplate \\
ss & good mix of data ; renamed from "ss" \\
st & ok \\
stq & ok i think ? \\
su & good \\
sus & hella sus jk ok bible \\
suz & ok bible \\
sv & ok \\
sw & ok \\
sxn & ok bible ;  also wild diacritics \\
sxu & rvisit after shortfilter \\
syl & idk maybe ok ? \\
syl-Latn & revist or remove after shortfilter \\
syr & good; practictitioners should keep dialect in mind. \\
ta & ok \\
ta-Latn & good text .... but pornographic, like all Indic-Latn datasets \\
tab & idk plausibly ok \\
taj & ok bible \\
tbz & good mostly bible but not all \\
tca & ok bible + jw \\
tcy & good; mostly wikipedia; likely some konkani mixed in \\
tdx & ok jw \\
te & ok a lot of weirdly low quality looking content like commerce \\
te-Latn & great good text....but all pornographic stories + blogs, like all Indic-Latn text \\
teo & ok bible \\
tet & good ; actually a lot of fun data! \\
tg & good \\
th & ok \\
ti & ok; poor tigray \\
tiv & ok jw \\
tk & ok; a few weird docs \\
tks & ok bible but again i think some mixed dialects \\
tlh & ok, but why tf are there websites in klingon? all MT ? \\
tll & ok jw \\
tly-IR & deeply sus; remove after shortfilter \\
tn & good \\
to & good ; news bible government \\
toj & ok jw \\
tpi & empy \\
tr & ok \\
trp & good ; lots of random stuff \\
trw & sus; remove \\
ts & good \\
tsc & ok \\
tsg & much noise but some good data too! \\
tt & good plus some nonunicode misrendered PDF \\
tuc & ok  bible \\
tuf & ok bible \\
tvl & ok jw \\
twu & ok bible, but also i think it's lots of mixed similar dialects \\
tyv & ok fun stuff plus some russian noise i think \\
tyz & ok bible bu again i think some mixed dialects \\
tzh & ok jw \\
tzj & ok bible \\
tzo & ok bible + jw \\
ubu & ok bible \\
udm & ok \\
ug & ok \\
uk & ok \\
ur & ok \\
uz & ok some cyrllic noise \\
ve & ok mostly bible jw \\
vec & very noisy has wiki from other langs and .it websites so not sure if vec \\
vi & ok \\
vkt & 1 doc remove \\
wa & ok lots of wiki stuff \\
wal & ok bible + jw \\
war & ok but v sus. Pls filter out wikipedia \\
wo & ok; mostly bible. PS i have found that Wolof web-crawled data is often bad, so pls give an extra look if you want \\
xal & ok has .ru sites though \\
xh & ok \\
xmm & very noisy lots of dj tiktok and peppa pig repeated \\
xnr & ok maybe fix virama though it seems fine \\
xog & ok bible and stories \\
yap & ok \\
yaq & remove \\
yi & ok \\
ymm & remove \\
yo & ok \\
yua & ok \\
yue & pretty low quality; mostly not Canto \\
za & revisit after shortfilter \\
zap & ok JW. PS pls note that at least some Zapotec speakers view it as one language, not as a million dialects like ISO does \\
zh & mixed simplified and trad; also much porn \\
zh-Latn & revisit after shortfilter \\
zne & ok jw \\
zu & good \\
zxx-xx-dtynoise & BEAUTIFUL NOISE rename but keep as beautiful xample. (was called "dty") \\
zyj & deeply bad data .. revisit after shortfilter \\
zza & good ; also pls note that the Zazaki community is often super into NLP for Zazaki \\
        \bottomrule \\
% \end{xltabular}
% }
% }

\end{longtable}
% }
}
}
% \insertauditnotes

\subsection{Parallel Data Details}\label{app:parallel}

To create the dataset described in Section \ref{sec:parallel}, we use the data sources described in Table \ref{tab:sources}. After preprocessing, we obtain a dataset with a total of 157 different languages and 4.1B sentence pairs that vary from \texttt{en-es} with 280.3M sentence pairs to \texttt{zu-en} with 7959 sentence pairs. The list of language pairs along with the associated data count is available along with the model checkpoints.

% visualize the 4.1k language pairs and their associated data counts in Figure.
% \ref{tab:parallel-dist}. 

\begin{table}[!ht]
    \centering
    \caption{\label{tab:sources} The various data sources used to create the parallel data used to train our MT models with the number of available languages and language pairs. (*for NewsCommentary v14 we only use Kazakh (\texttt{kk}) data)}
    \begin{tabular}{llll}
    \toprule
        Dataset & Version \# Language Pairs & \# Languages \\ 
    \midrule
        Europarl \cite{koehn2005europarl} & v7 & 40 & 21 \\ 
         & v9 & 12 & 8 \\ 
        Paracrawl \cite{espla2019paracrawl} & - & 64 & 41 \\ 
        TED57 \cite{Ye2018WordEmbeddings} & - & 114 & 58 \\ 
        Tanzil \cite{tiedemann2012parallel} & - & 1560 & 40 \\ 
        NewsCommentary \cite{statmt} & v10 & 6 & 3 \\ 
         & v14* & 2 & 1 \\ 
         & v16 & 200 & 15 \\ 
        Wikimatrix \cite{schwenk2019wikimatrix} & - & 1620 & 85 \\ 
        Wikititles \cite{statmt} & v2 & 12 & 14 \\ 
        OPUS100 \cite{zhang2020improving} & - & 198 & 100 \\ 
        SETimes \cite{tiedemann2012parallel} & - & 90 & 10 \\ 
        UNv1.0 \cite{ziemski2016united} & - & 20 & 5 \\ 
        Autshumato \cite{groenewald2009introducing} & - & 10 & 5 \\ 
        PMIndia \cite{haddow2020pmindia} & v1 & 13 & 13 \\ 
        CVIT \cite{philip2019baseline} & MKB (v0), PIB (v1.3)  & 110 & 11 \\ 
        Inuktitut \cite{joanis2020nunavut} & - & 2 & 1 \\ 
        NLPC \cite{fernando2020data} & - & 2 & 1 \\ 
        JESC \cite{pryzant2017jesc} & - & 2 & 1 \\ 
        KFTT \cite{neubig11kftt} & v1.0 & 2 & 1 \\ 
        ASPEC \cite{nakazawa2016aspec} & - & 2 & 1 \\ 
        \bottomrule
    \end{tabular}
\end{table}

% \input{TabsNFigs/ParallelData}

% \usepackage{booktabs}
\begin{table}[]
\centering
\caption{Preprocessed multiway data divided by target language in decreasing order of number of \# Sentence Pairs.}
\label{tab:multiway-dist} 
\resizebox{0.85\columnwidth}{!}{
% \usepackage{booktabs}
\begin{tabular}{@{}lr|lr@{}}
\toprule
\textbf{Language (\texttt{Code})}        & \multicolumn{1}{l}{\textbf{\# Sentence Pairs}} & \textbf{Language (\texttt{Code})}          & \multicolumn{1}{l}{\textbf{\# Sentence Pairs}} \\ \midrule
English (\texttt{en})                    & 6825073150                                     & Malagasy (\texttt{mg})                     & 1190715                                        \\
Spanish (\texttt{es})                    & 630962081                                      & Punjabi (\texttt{pa})                      & 1122414                                        \\
German (\texttt{de})                     & 542301841                                      & Hausa (\texttt{ha})                        & 952163                                         \\
French (\texttt{fr})                     & 540506439                                      & Latin (\texttt{la})                        & 882038                                         \\
Portuguese (\texttt{pt})                 & 355008897                                      & Inuktitut (\texttt{iu})                    & 841459                                         \\
Italian (\texttt{it})                    & 283297447                                      & Myanmar (Burmese) (\texttt{my})            & 759196                                         \\
Dutch (\texttt{nl})                      & 272264696                                      & Walloon (\texttt{wa})                      & 706400                                         \\
Swedish (\texttt{sv})                    & 183450145                                      & Uzbek (\texttt{uz})                        & 695915                                         \\
Czech (\texttt{cs})                      & 166419080                                      & Luxembourgish (\texttt{lb})                & 645610                                         \\
Polish (\texttt{pl})                     & 154989834                                      & Assamese (\texttt{as})                     & 623321                                         \\
Mandarin Chinese (\texttt{zh})           & 147268699                                      & Pashto (\texttt{ps})                       & 606852                                         \\
Danish (\texttt{da})                     & 145764833                                      & Armenian (\texttt{hy})                     & 603397                                         \\
Hungarian (\texttt{hu})                  & 136980377                                      & Sindhi (\texttt{sd})                       & 597211                                         \\
Russian (\texttt{ru})                    & 134568973                                      & Northern Sami (\texttt{se})                & 585304                                         \\
Romanian (\texttt{ro})                   & 117227121                                      & Bashkir (\texttt{ba})                      & 551617                                         \\
Slovak (\texttt{sk})                     & 102889505                                      & Amharic (\texttt{am})                      & 548828                                         \\
Finnish (\texttt{fi})                    & 100468712                                      & Somali (\texttt{so})                       & 543643                                         \\
Greek (\texttt{el})                      & 97855862                                       & Dhivehi (\texttt{dv})                      & 543145                                         \\
Arabic (\texttt{ar})                     & 83364997                                       & Kurdish (Kurmanji) (\texttt{ku})           & 483477                                         \\
Bulgarian (\texttt{bg})                  & 68424658                                       & French (Canada) (\texttt{fr\_CA})           & 472160                                         \\
Lithuanian (\texttt{lt})                 & 66562681                                       & Odia (Oriya) (\texttt{or})                 & 465494                                         \\
Slovenian (\texttt{sl})                  & 65876962                                       & Faroese (\texttt{fo})                      & 434288                                         \\
Latvian (\texttt{lv})                    & 59061871                                       & Kannada (\texttt{kn})                      & 417255                                         \\
Norwegian (\texttt{no})                  & 57783339                                       & Kinyarwanda (\texttt{rw})                  & 386133                                         \\
Estonian (\texttt{et})                   & 47403866                                       & Wu Chinese (\texttt{wuu})                  & 369891                                         \\
Japanese (\texttt{ja})                   & 42480849                                       & Lombard (\texttt{lmo})                     & 366147                                         \\
Korean (\texttt{ko})                     & 33987656                                       & Egyptian Arabic (\texttt{arz})             & 353218                                         \\
Catalan (\texttt{ca})                    & 28627098                                       & Uyghur (\texttt{ug})                       & 337582                                         \\
Croatian (\texttt{hr})                   & 28475191                                       & Limburgan (\texttt{li})                    & 304279                                         \\
Turkish (\texttt{tr})                    & 26508188                                       & Aragonese (\texttt{an})                    & 251004                                         \\
Ukrainian (\texttt{uk})                  & 26258530                                       & Sicilian (\texttt{scn})                    & 249682                                         \\
Icelandic (\texttt{is})                  & 24073502                                       & Dzongkha (\texttt{dz})                     & 248172                                         \\
Persian (\texttt{fa})                    & 23237370                                       & Meiteilon (Manipuri) (\texttt{mni})        & 242426                                         \\
Indonesian (\texttt{id})                 & 19860783                                       & Maori (\texttt{mi})                        & 234833                                         \\
Vietnamese (\texttt{vi})                 & 18803243                                       & Nuer (\texttt{nus})                        & 231194                                         \\
Hebrew (\texttt{he})                     & 17171755                                       & Magahi (\texttt{mag})                      & 230839                                         \\
Macedonian (\texttt{mk})                 & 15900226                                       & Bhojpuri (\texttt{bho})                    & 230378                                         \\
Irish (\texttt{ga})                      & 14939080                                       & Shan (\texttt{shn})                        & 229967                                         \\
Galician (\texttt{gl})                   & 14305858                                       & Friulian (\texttt{fur})                    & 228928                                         \\
Serbian (\texttt{sr})                    & 13989182                                       & Kashmiri (\texttt{ks})                     & 228580                                         \\
Albanian (\texttt{sq})                   & 13904626                                       & Sardinian (\texttt{sc})                    & 227585                                         \\
Basque (\texttt{eu})                     & 13755514                                       & Kanuri (\texttt{kr\_Arab})                  & 227549                                         \\
Maltese (\texttt{mt})                    & 12583626                                       & Dinka (\texttt{din})                       & 227233                                         \\
Esperanto (\texttt{eo})                  & 11431440                                       & Venetian (\texttt{vec})                    & 225065                                         \\
Hindi (\texttt{hi})                      & 11044239                                       & Chhattisgarhi (\texttt{hne})               & 224085                                         \\
Bosnian (\texttt{bs})                    & 10906114                                       & Ligurian (\texttt{lij})                    & 222513                                         \\
Serbo-Croatian (\texttt{sh})             & 10592606                                       & Central Atlas Tamazight (\texttt{tzm})     & 222365                                         \\
Bengali (\texttt{bn})                    & 8394639                                        & Tamasheq (\texttt{taq\_Tfng})               & 220907                                         \\
Sinhala (\texttt{si})                    & 7363378                                        & Dari (\texttt{prs})                        & 220899                                         \\
Thai (\texttt{th})                       & 6623168                                        & Banjar (\texttt{bjn})                      & 220753                                         \\
Malayalam (\texttt{ml})                  & 5995504                                        & Achinese (\texttt{ace\_Arab})               & 220651                                         \\
Tamil (\texttt{ta})                      & 5552495                                        & Tamasheq (\texttt{taq})                    & 219708                                         \\
Marathi (\texttt{mr})                    & 5517596                                        & Banjar (\texttt{bjn\_Arab})                 & 219020                                         \\
Telugu (\texttt{te})                     & 5379834                                        & Achinese (\texttt{ace})                    & 217062                                         \\
Malay (\texttt{ms})                      & 5241831                                        & Bambara (\texttt{bm})                      & 216698                                         \\
Filipino (\texttt{fil})                  & 4532364                                        & Balinese (\texttt{ban})                    & 215965                                         \\
Urdu (\texttt{ur})                       & 4526509                                        & Moroccan Arabic (\texttt{ary})             & 214226                                         \\
Taiwanese Mandarin (\texttt{zh\_Hant})    & 3782804                                        & Nigerian Fulfulde (\texttt{fuv})           & 209659                                         \\
Swahili (\texttt{sw})                    & 3220515                                        & Silesian (\texttt{szl})                    & 209224                                         \\
Nepali (\texttt{ne})                     & 2833740                                        & Kashmiri (\texttt{ks\_Deva})                & 208575                                         \\
Norwegian Nynorsk (\texttt{nn})          & 2726466                                        & Buginese (\texttt{bug})                    & 207209                                         \\
Azerbaijani (\texttt{az})                & 2353731                                        & Guarani (\texttt{gn})                      & 201241                                         \\
Kazakh (\texttt{kk})                     & 2153356                                        & Latgalian (\texttt{ltg})                   & 200302                                         \\
Tajik (\texttt{tg})                      & 2005810                                        & Crimean Tatar (\texttt{crh\_Latn})          & 192342                                         \\
Low German (\texttt{nds})                & 1942746                                        & Kanuri (\texttt{kr})                       & 191037                                         \\
Occitan (\texttt{oc})                    & 1934958                                        & Scottish Gaelic (\texttt{gd})              & 162882                                         \\
Georgian (\texttt{ka})                   & 1882623                                        & Bavarian (\texttt{bar})                    & 159419                                         \\
Belarusian (\texttt{be})                 & 1808663                                        & Javanese (\texttt{jv})                     & 144974                                         \\
Tatar (\texttt{tt})                      & 1763087                                        & Mongolian (\texttt{mn})                    & 132599                                         \\
Western Frisian (\texttt{fy})            & 1500410                                        & Zulu (\texttt{zu})                         & 130485                                         \\
Afrikaans (\texttt{af})                  & 1459139                                        & Kyrghyz (\texttt{ky})                      & 128832                                         \\
Breton (\texttt{br})                     & 1431191                                        & Low German (Netherlands) (\texttt{ndsNL}) & 113101                                         \\
English (simple) (\texttt{en\_xx\_simple}) & 1413074                                        & Mirandese (\texttt{mwl})                   & 96829                                          \\
Khmer (\texttt{km})                      & 1330590                                        & Ido (\texttt{io})                          & 72811                                          \\
Gujarati (\texttt{gu})                   & 1309569                                        & Igbo (\texttt{ig})                         & 41296                                          \\
Cebuano (\texttt{ceb})                   & 1255728                                        & Turkmen (\texttt{tk})                      & 40650                                          \\
Welsh (\texttt{cy})                      & 1229401                                        & Yiddish (\texttt{yi})                      & 37128                                          \\
Xhosa (\texttt{xh})                      & 1219701                                        & Yoruba (\texttt{yo})                       & 26589                                          \\\midrule
 & & \textbf{Total \# Sentence Pairs} &  11944961985  \\
\bottomrule
\end{tabular}
}
\end{table}

\subsection{Language Codes}

The specifics of the language code changes described in Section \ref{sec:mt-model} that we made are as follows:

\begin{enumerate}[leftmargin=*]
  \item We use \texttt{fil} for Filipino/Tagalog, not \texttt{tl}
  \item We use \texttt{ak} for Twi/Akan, rather than \texttt{tw}. This includes Fante.
  \item Unfortunately, we use the macro code \texttt{chm} for Meadow Mari (instead of the correct \texttt{mhr}), and \texttt{mrj} for Hill Mari
  \item By convention, we use \texttt{no} for Norwegian Bokmål, whereas some resources use \texttt{nb}
  \item By convention we use \texttt{ps} for Pashto instead of \texttt{pbt} (Southern Pashto)
  \item By convention, we use \texttt{ms} for Standard Malay, not \texttt{zlm}
  \item By convention, we use \texttt{sq} for Albanian, and don't distinguish dialects like
Gheg (\texttt{aln}) and Tosk (\texttt{als})
  \item We use \texttt{ber} as the code for Tamazight, after consultation with Tamazight
 speakers opining that the dialect distinctions are not significant. Other resources use the individual codes like \texttt{tzm} and \texttt{kab}.
  \item We use the macrocode \texttt{qu} for Quechua. In practice, this seems usually to be a mix of the Ayacucho and Cusco dialects. Other resources, like NLLB, may  use the dialect code, e.g. \texttt{quy} for Ayacucho Chanka. The same is true for a few other macro codes, like \texttt{ff} (Macro code for Fulfulde, whereas other sources may use e.g. \texttt{fuv}.)
  \item Really, there are notes that can be made about almost any code, from the well-accepted conventions like \texttt{zh} for Mandarin, to many dialectical notes, like which variant of Hmong really is the \texttt{hmn} data? But The above ones are made specifically for ones where we are aware of other datasources floating out there that use different conventions.
\end{enumerate}

\subsection{Multiway Data Details}\label{app:multiway}

On creating the multiway data described in Section \ref{sub:multiway}, we obtain a dataset with 11.9B sentence pairs across 19.7k language pairs. In Table \ref{tab:multiway-dist}, we list the combined number of sentence pairs for each target language.

\subsection{Model Training Details}\label{app:model}

\paragraph{MT Model Training} We train models of various sizes: a 3B, 32-layer parameter model,\footnote{Here and elsewhere, `X-layer' means X encoder layers and also X decoder layers, for a total of 2X layers.} a 7.2B 48-layer parameter model and a 10.7B 32-layer parameter model. We describe the specifics of the model architecture in Table \ref{tab:model-arch}.

We share all parameters of the model across language pairs, and use a Sentence Piece Model (SPM)~\citep{kudo2018sentencepiece} with 256k tokens shared on both the encoder and decoder side. We train the SPM model on upto 1M sentence samples of the sentences in the sentence-level version of \data, supplemented by data from the languages in the parallel data used to train MT models when not available in \data with a temperature of $T=100$ and a character coverage of $99.9995\%$.

Each input sentence has a \texttt{<2xx>} token prepended to the source sentence to indicate the target language~\citep{johnson2017google}. We use both supervised parallel data with a machine translation objective and the monolingual \data dataset with a MASS-style~\citep{song2019mass} objective to train this model. Each of these objectives is sampled with a 50\% probability. Within each task, we use the recently introduced UniMax~\citep{chung2023unimax} sampling strategy to sample languages from our imbalanced dataset with a threshold of $N=10$ epochs for any particular language.

We used a square root learning rate decay schedule over the total number of training steps, starting at 0.01 and ending at X, as well as the AdaFactor optimizer with \texttt{factorized=False} and 10k warmup steps. We note that for the 10.7B model we use a dropout probability of $p=0.2$ instead of $p=0.1$ in order to mitigate overfitting to low-resource languages. 

\paragraph{Language Model Training} We follow the same training schedule and model configurations from~\citet{garcia2023unreasonable}. In particular, we consider 8B decoder-only models, following the same model hyperparameters as previous work~\citep{chowdhery2022palm, garcia2023unreasonable}. We train these models using a variant of the UL2 objective~\citep{tay2022unifying} adapted for decoder-only models, and use the same configuration as previous work~\citep{garcia2023unreasonable, orlanski2023measuring}. We point the reader to these papers for a detailed overview of the training process, and include basic architectural details in Table \ref{tab:model-arch}. We use the same SPM trained for the MT models.

We use a square root learning rate decay schedule over 500k steps, starting at 0.01 and ending at 0.001, as well as the AdaFactor optimizer with \texttt{factorized=False} with 10k warmup steps. We describe the evaluation setup in Section \ref{sub:few-shot}.

\begin{table}[]
\centering
\caption{\label{tab:model-arch} Architecture and training details for the various models we train in this work using \data.}
\resizebox{\textwidth}{!}{%
\begin{tabular}{lcccccccccc}
\toprule
\textbf{Model} & \multicolumn{1}{l}{SPM Size} & \multicolumn{1}{l}{Training Objective} & \multicolumn{1}{l}{\# Attn Heads} & \multicolumn{1}{l}{\# Enc. Layers} & \multicolumn{1}{l}{\# Dec. Layers} & \multicolumn{1}{l}{Model Dim} & \multicolumn{1}{l}{Hidden Dim} & \multicolumn{1}{l}{Batch Size} & \multicolumn{1}{l}{Max. Seq. Length} & Training Steps \\
\midrule
3B MT Model & \multirow{4}{*}{256k} & MT+MASS & 16 & 32 & 32 & 1024 & 8192 & 4096 & 256 & 1M \\
7.2B MT Model &  & MT+MASS & 16 & 32 & 32 & 2048 & 8192 & 4096 & 256 & 500k \\
10.7B MT Model &  & MT+MASS & 32 & 48 & 48 & 2048 & 16384 & 4096 & 256 & 250k \\
8B LM &  & UL2 & 16 & N/A & 32 & 4096 & 16384 & 1024 & 1024 & 500k \\
\bottomrule
\end{tabular}%
}
\end{table}

\subsection{Languages Evaluated}\label{app:overlaps}

In Table \ref{tab:overlaps}, we list the languages for which we evaluate the models trained as described in Sections \ref{sec:mt-model} and \ref{sec:lm}.

\begin{table}
\centering
\caption{\label{tab:overlaps} Languages on which we evaluate the trained models for each multilingual evaluation set.} 
\begin{tabular}{lp{11cm}}
\toprule
\textbf{Datasets}  &  \textbf{Languages Evaluated } \\
\midrule
WMT  &  \texttt{cs, de, es, fi, fr, gu, hi, kk, lv, lt, ro, rs, es, tr, zh}  \\\midrule
Flores-200  &  \texttt{ac\_Arab, ace, af, am, ar, arz, as, awa, ay, az, ba, ban, be, ber, bg, bho, bjn\_Arab, bjn, bm, bn, bo, bs, bug, ca, ceb, ckb, crh\_Latn, cs, cy, da, de, din, dyu, dz, ee, el, eo, es, et, eu, fa, ff, fi, fil, fj, fo, fon, fr, fur, ga, gd, gl, gn, gu, ha, hi, hne, hr, ht, hu, hy, id, ig, ilo, is, it, he, ja, jv, ka, kac, kbp, kg, kk, km, kmb, kn, ko, kr\_Arab, kr, ks\_Deva, ks, ky, lb, lg, li, lij, lmo, ln, lo, lt, ltg, lus, lv, mag, mai, mg, mi, min, mk, ml, mn, mni, mr, ms, mt, my, ne, nl, nn, no, nso, nus, ny, oc, om, or, pa, pag, pap, pl, ps, pt, quy, rn, ro, ru, rw, sa, sc, scn, sd, sg, shn, si, sk, sl, sm, sn, so, sq, sr, ss, st, su, sv, sw, szl, ta, taq\_Tfng, taq, te, tg, th, ti, tk, tn, tr, ts, tt, ug, uk, ur, vec, vi, war, wo, xh, yi, yo, zh\_Hant, zh, zu} \\\midrule
NTREX  &  \texttt{af, am, ar, az, ba, be, bem, bg, bn, bo, bs, ca, ckb, cs, cy, da, de, dv, dz, ee, el, en\_GB, en\_IN, es, es\_MX, et, eu, fa, fa\_AF, ff, fi, fil, fj, fo, fr, fr\_CA, ga, gl, gu, ha, hi, hmn, hr, hu, hy, id, ig, is, it, iw, ja, ka, kk, km, kn, ko, ku, ky, lb, lo, lt, lv, mey, mg, mi, mk, ml, mn, mr, ms, mt, my, nd, ne, nl, nn, no, nso, ny, om, pa, pl, ps, pt, pt\_PT, ro, ru, rw, sd, shi, si, sk, sl, sm, sn, so, sq, sr, sr\_Latn, ss, sv, sw, ta, te, tg, th, ti, tk, tn, to, tr, tt, ty, ug, uk, ur, uz, ve, vi, wo, xh, yo, yue, zh, zh\_Hant, zu}  \\\midrule
Gatones  &  \texttt{ady, ak, as, av, ay, ba, ban, bbc, bci, ber\_Latn, bew, bho, bm, bo, ce, chr, ckb, cv, doi, dv, dyu, dz, ee, ff, gn, gom, ilo, iso, kl, kri, lg, ln, lus, mad, mai, meo, min, mni, nso, om, or, qu, quc, rw, sa, sg, skr, ti, tiv, tk, ts, tt, ug, wo, yua, zza}  \\
 \bottomrule
 \end{tabular}
\end{table}

\subsection{Results Details}\label{app:res}
In Tables \ref{tab:wmt-full}, \ref{tab:ntrex-full}, \ref{tab:ntl-full}, \ref{tab:flores-full} and \ref{tab:flores-direct} we list the WMT, NTREX, Gatones, Flores-200 and Flores-200 (direct pairs) chrf and SacreBLEU scores respectively by language pair along with the model checkpoints.
\newpage

\begin{table}[t!]
\centering
\caption{\label{tab:wmt-full} Evaluation scores on WMT (depicted as \texttt{<bleu> / <chrf>}) for the MT models and language models described in Section \ref{sec:mt-model} and Section \ref{sec:lm} compared against NLLB-54B.}
\resizebox{\columnwidth}{!}{
\begin{tabular}{@{}lcccccccc@{}}
\toprule
\multirow{2}{*}{} & \multicolumn{1}{c}{\multirow{2}{*}{\textbf{NLLB}}} & \multicolumn{1}{c}{\multirow{2}{*}{\textbf{MT-3B}}} & \multicolumn{1}{c}{\multirow{2}{*}{\textbf{MT-7.2B}}} & \multicolumn{1}{c}{\multirow{2}{*}{\textbf{MT-10.7B}}} & \multicolumn{4}{c}{\textbf{LM-8B}}                                                                                          \\ \cmidrule(l){6-9} 
                  & \multicolumn{1}{c}{}                      & \multicolumn{1}{c}{}                       & \multicolumn{1}{c}{}                         & \multicolumn{1}{c}{}                        & \multicolumn{1}{c}{0-shot} & \multicolumn{1}{c}{1-shot} & \multicolumn{1}{c}{5-shot} & \multicolumn{1}{c}{10-shot} \\ \midrule
csen & 33.6 / 58.8 & 34.8 / 59.9 & 35.7 / 60.5 & 35.3 / 60.4 & 3 / 22.2 & 27.6 / 53.7 & 26.7 / 51.8 & 28.2 / 54 \\
deen & 37.7 / 62.4 & 37.3 / 62.5 & 38.1 / 63 & 38 / 63 & 3.3 / 23.6 & 26 / 51.9 & 29.7 / 55.2 & 30.1 / 55.5 \\
esen & 37.6 / 61.7 & 38.1 / 62.4 & 38.5 / 62.6 & 38.5 / 62.7 & 2.2 / 19 & 30 / 55.2 & 31.5 / 57.4 & 32.3 / 58.4 \\
eten & 34.7 / 61.1 & 36.2 / 61.9 & 37.4 / 62.8 & 37.4 / 62.9 & 4.7 / 24.7 & 26.4 / 52.8 & 27.9 / 55 & 27.7 / 54.5 \\
fien & 28.8 / 55.7 & 32.8 / 59.1 & 33.5 / 59.4 & 33.3 / 59.6 & 1.6 / 16.8 & 27.1 / 53.5 & 26.2 / 52 & 26.3 / 52.6 \\
fren & 41.9 / 65.7 & 42.2 / 65.6 & 42.7 / 65.9 & 42.4 / 65.8 & 2 / 18.6 & 28.9 / 54.8 & 35.1 / 60.7 & 35.6 / 61.1 \\
\bottomrule
\end{tabular}
}\vspace{3mm}
\end{table}

{
\fontsize{6.4}{8}\selectfont {
\begin{longtable}{@{}lccccccccc@{}}
% % \centering
% \caption{\label{tab:ntrex-full} Evaluation scores on the recently introduced NTREX test set (depicted as \texttt{<bleu> / <chrf>}) for the MT models and language models described in Section \ref{sec:mt-model} and Section \ref{sec:lm} compared against unsupervised baselines \cite{baziotis2023does}. Note that LM-8B is evaluated on a 50\% split of the NTREX data and is not comparable to the MT-model evaluations.} \\

% \begin{table}[h!]
% \centering
\caption{\label{tab:ntl-full}  Evaluation scores on the \textsc{Gatones} test set used by \citet{bapna-etal-2022-building} (depicted as \texttt{<bleu> / <chrf>}) for the MT models and language models described in Section \ref{sec:mt-model} and Section \ref{sec:lm}.}\\
% \resizebox{\columnwidth}{!}{
% \begin{tabular}{@{}lccccccccc@{}}
\toprule\\
\multirow{2}{*}{} & \multicolumn{2}{c}{{\textbf{NTL (\citet{bapna-etal-2022-building})}}} & \multicolumn{1}{c}{\multirow{2}{*}{\textbf{MT-3B}}} & \multicolumn{1}{c}{\multirow{2}{*}{\textbf{MT-7.2B}}} & \multicolumn{1}{c}{\multirow{2}{*}{\textbf{MT-10.7B}}} & \multicolumn{4}{c}{\textbf{LM-8B}}                                                                                          \\ \cmidrule(l){2-3}\cmidrule(l){7-10} 
 & 1.6B & 6.4B & & & & 0-shot & 1-shot & 5-shot & 10-shot \\ \midrule

ady\_en & - / 53.7 & - / 54.8 & 32.7 / 58.8 & 34.2 / 59.8 & 33.9 / 60.1 & 1.5 / 12.4 & 18.9 / 50.2 & 25 / 54.2 & 21.1 / 52.1 \\
ak\_en & - / 31.7 & - / 36.3 & 6.9 / 27 & 10.4 / 29.9 & 9.3 / 29 & 0.1 / 5.5 & 3.6 / 15.5 & 1.8 / 18.6 & 1.9 / 16.6 \\
as\_en & - / 52.9 & - / 58.6 & 30.6 / 55.6 & 32.1 / 56.8 & 33.8 / 58 & 1 / 9.7 & 17.4 / 44.5 & 19.4 / 47.4 & 21.1 / 48.6 \\
av\_en & - / 48.1 & - / 48.1 & 19.7 / 47.8 & 21.3 / 50.6 & 21.5 / 50.9 & 0.3 / 6.1 & 11.6 / 32.5 & 10.3 / 40.2 & 11.1 / 40.6 \\
\end{longtable} }}

\newpage

% \usepackage{booktabs}
% \usepackage{multirow}
% \begin{table}[h!]
% \centering
{
\fontsize{6.4}{8}\selectfont {

\begin{longtable}{@{}lcccccccc@{}}
\caption{\label{tab:flores-full} Evaluation scores on en-centric Flores-200 pairs (depicted as \texttt{<bleu> / <chrf>}) for the MT models and language models described in Section \ref{sec:mt-model} and Section \ref{sec:lm} compared against NLLB-54B. All metrics are computed with the sacrebleu reference implementation.} \\
% \resizebox{\columnwidth}{!}{
% \begin{tabular}{@{}lcccccccc@{}}
\toprule
\multirow{2}{*}{} & \multicolumn{1}{c}{\multirow{2}{*}{\textbf{NLLB}}} & \multicolumn{1}{c}{\multirow{2}{*}{\textbf{MT-3B}}} & \multicolumn{1}{c}{\multirow{2}{*}{\textbf{MT-7.2B}}} & \multicolumn{1}{c}{\multirow{2}{*}{\textbf{MT-10.7B}}} & \multicolumn{4}{c}{\textbf{LM-8B}}                                                                                          \\ \cmidrule(l){6-9} 
                  &                                &                                              &                                                &                                       & 0-shot        & 1-shot        & 5-shot        & 10-shot       \\ \midrule
en\_ace\_Arab & 0.4 / 21.1 & 1.4 / 26.6 & 1.4 / 25.3 & 1.3 / 26.9 & 0 / 0.1 & 0 / 0.5 & 0 / 2.9 & 0 / 2.2 \\
en\_ace & 9.3 / 41.4 & 8.3 / 39.6 & 6.5 / 36.6 & 7.6 / 39.1 & 0.1 / 3.7 & 0.9 / 15.3 & 0.8 / 21.1 & 0.7 / 22.4 \\
en\_acm & 10.3 / 35.4 & 0.6 / 0.8 & 0.4 / 0.8 & 0.6 / 0.8 & 0 / 0.4 & 4.8 / 34.6 & 8.3 / 43.3 & 6.1 / 41.1 \\
en\_acq & 13.6 / 46.8 & 0.6 / 0.8 & 0.4 / 0.7 & 0.6 / 0.8 & 0.1 / 1.2 & 5.9 / 36.1 & 10.1 / 43.5 & 9.7 / 44 \\
en\_aeb & 12 / 42.1 & 0.6 / 0.8 & 0.6 / 0.8 & 0.6 / 0.8 & 0.1 / 1.2 & 1.4 / 20.3 & 4.1 / 35.4 & 4.9 / 37.2 \\
\end{longtable} }}


% \usepackage{booktabs}
% \usepackage{multirow}

{
\fontsize{6.4}{8}\selectfont {

\begin{longtable}{@{}lcccccccc@{}}

\caption{\label{tab:flores-direct} Evaluation scores on Flores-200 direct pairs (depicted as \texttt{<bleu> / <chrf>}) for the MT models and language models described in Section \ref{sec:mt-model} and Section \ref{sec:lm} compared against NLLB-54B. All metrics are computed with the sacrebleu reference implementation.}\\
\toprule \\
\multirow{2}{*}{} & \multicolumn{1}{c}{\multirow{2}{*}{\textbf{NLLB}}} & \multicolumn{1}{c}{\multirow{2}{*}{\textbf{MT-3B}}} & \multicolumn{1}{c}{\multirow{2}{*}{\textbf{MT-7.2B}}} & \multicolumn{1}{c}{\multirow{2}{*}{\textbf{MT-10.7B}}} & \multicolumn{4}{c}{\textbf{LM-8B}}                                                                                          \\ \cmidrule(l){6-9} 
                  &                                &                                              &                                                &                                       & 0-shot        & 1-shot        & 5-shot        & 10-shot       \\ \midrule

ckb\_cs & 20.7 / 47.3 & 15.4 / 39.4 & 18.5 / 44.9 & 19.1 / 45.1 & 0 / 0.8 & 8.1 / 30.2 & 10.4 / 35.5 & 9.7 / 36.6 \\
ckb\_cy & 26.9 / 53.6 & 19.3 / 41.4 & 23.6 / 47 & 23.9 / 47.8 & 0 / 0.7 & 0.3 / 8.8 & 0.7 / 15.1 & 0.3 / 12.4 \\
ckb\_en & 37.5 / 61.6 & 28.3 / 53.1 & 34.4 / 58.9 & 34.2 / 59 & 1 / 9.7 & 17.9 / 42.5 & 22.4 / 48.4 & 22.7 / 49.4 \\
ckb\_et & 16.4 / 48.6 & 13 / 40.8 & 15.3 / 45.6 & 16.3 / 47 & 0 / 1 & 1.9 / 12.9 & 6.7 / 36.6 & 7.3 / 38.2 \\
\bottomrule
\end{longtable}
}
}

% \usepackage{booktabs}
% \usepackage{multirow}
{
\fontsize{6.4}{8}\selectfont {
\begin{longtable}{@{}lcccccccc@{}}
% \centering
\caption{\label{tab:ntrex-full} Evaluation scores on the recently introduced NTREX test set (depicted as \texttt{<bleu> / <chrf>}) for the MT models and language models described in Section \ref{sec:mt-model} and Section \ref{sec:lm} compared against unsupervised baselines \cite{baziotis2023does}. Note that LM-8B is evaluated on a 50\% split of the NTREX data and is not comparable to the MT-model evaluations.} \\
% \resizebox{\columnwidth}{!}{
% \begin{tabular}{@{}lcccccccc@{}}
\toprule \\
\multirow{2}{*}{} & \multicolumn{1}{c}{\multirow{2}{*}{\textbf{\citet{baziotis2023does}}}} & \multicolumn{1}{c}{\multirow{2}{*}{\textbf{MT-3B}}} & \multicolumn{1}{c}{\multirow{2}{*}{\textbf{MT-7.2B}}} & \multicolumn{1}{c}{\multirow{2}{*}{\textbf{MT-10.7B}}} & \multicolumn{4}{c}{\textbf{LM-8B}}                                                                                          \\ \cmidrule(l){6-9} 
 & & & & & 0-shot & 1-shot & 5-shot & 10-shot \\ \midrule
 af\_en & - / - & 57.1 / 75.4 & 58.5 / 76.5 & 58.4 / 76.4 & 7.6 / 30.6 & 49 / 69.9 & 49.2 / 69.6 & 51.8 / 71.9 \\
am\_en & - / - & 24.3 / 50.3 & 27.2 / 52.5 & 27.2 / 52.7 & 1 / 10.9 & 13.4 / 37.9 & 14.6 / 40.4 & 13.8 / 40.3 \\
ar\_en & - / - & 36 / 61.9 & 37.9 / 63.1 & 37.8 / 63.1 & 1.1 / 10 & 23.7 / 51.6 & 28.6 / 53.9 & 28.8 / 54.1 \\
az\_en & - / - & 26.3 / 52.7 & 29.1 / 54.9 & 30.1 / 55.8 & 4.7 / 26.3 & 15.5 / 40.8 & 20.6 / 47 & 21.8 / 48.6 \\
\end{longtable}

}}

\newpage
\pagebreak

\pagebreak
\subsection{Canaries}\label{sec:canary-details}

We design and generate different types of canaries for each dataset. We treat the \data dataset as a large unlabeled pretraining corpus and for each language with sufficient size (  $>50,000$ samples), we generate no more than $0.05\%$ canaries per language, leading to a total of $1,279,635$ canaries across all languages. For parallel data, we design and generate different canaries specific to its usage for translation. We generate $585,596$ canaries in total for all target languages with $>50,000$ samples. In both cases, we scale the proportion of canaries based on the size of each language to minimize their impact to utility in model training.
Finally, we also generate $80,400$ ``generic'' canaries that share no resemblance to natural data. In total, we generate $1,945,631$ canaries ($\approx0.0026\%$ of the training data).

\subsubsection{\data Canaries}\label{ssec:canary-madlad}
Because \data contains only monolingual data, i.e., each example relates to only one language, we treat it as a large unlabeled pretraining corpus. Similar to ~\citet{anil2023palm}, we aim to generate canaries that share characteristics of the underlying training data but are still outliers. For this, we design three types of canaries: \texttt{shuffle}, \texttt{interleave}  ~and \texttt{training\_prefix}. \texttt{interleave} canaries can be viewed as the closest to natural data, where all tokens are sampled from the underlying distribution and most sequence-level correlations are kept intact. We generate these canaries by sampling two real documents from a language, and interspersing chunks of $20$ tokens in their same relative ordering.
On the other hand, \texttt{shuffle} ~canaries can be viewed as the farthest from natural data, sharing only the average token frequencies of the language but with no sequence-level correlations. These are generated by sampling a real document and uniformly sampling its tokens without replacement.
In addition, we also propose \texttt{training\_prefix} ~canaries which can be viewed as something in between. Here,
each canary is generated by sampling $100$ tokens from a real sample and then completing the sequence with random tokens (i.e, taken uniformly with replacement from the vocabulary). 

We take care to adjust the number and type of canaries based on the resources of the language, in order to minimize any harm to model utility. We group languages based on their relative size. Then, prior to generation we fix a target canary rate based on this resource level - this determines the total number of canaries that can be added per language.
We choose a smaller proportion of canaries (relative to the total number of sequences in the language) for lower-resource languages based on the intuition that these languages can tolerate less noisy data before utility is altered. With the number of canaries fixed, we then choose how many times each canary will be repeated, as this has a significant impact on memorization~\cite{lee2021deduplicating,anil2023palm}. Note that repeating canaries may be beneficial to study the relative vulnerability of repeated data, but also reduces to the total number of unique canaries that can be added within that language's canary budget.
We choose the distribution of repeats heuristically aiming to maximize the support of the distribution while also ensuring that each bucket has enough unique canaries to achieve meaningful results. Finally, as the language size grows, (and thus the canary budget as well) we also introduce more canary types.
We describe the full distribution of canaries generated in Table~\ref{tab:canary-distributions}.

% ; the latter generates a fresh prefix of $100$ random tokens for each language and then generates a new canary by concatenating a real data sample.

\newcommand\verythinrule{\specialrule{.02em}{0.3em}{0.2em}}
\newcommand\thinrule{\specialrule{.04em}{0.5em}{0.4em}}

{\fontsize{6.4}{8}\selectfont
\begin{longtable}{cm{6em}ccccccccccccc}
        \caption{\label{tab:canary-distributions}
        \data distribution of (the $1,279,635$ in total) canaries across languages. For legibility, we omit 0 entries. *Uses the codes: I=interleave, S=shuffle, TP=training\_prefix.}\\
    % \begin{tabular}{@{}c m{6em} c c c c c c c c c c c@{}}
    \toprule\\
        \multirow{2}{*}{\makecell{Dataset Size\\(\# Samples)}} & \multirow{2}{*}{\makecell{Languages\\Included}} & \multirow{2}{*}{\makecell{Target\\Canary Rate}} & \multirow{2}{*}{\makecell{Canary\\Types*}} & \multirow{2}{*}{\makecell{Canaries\\per Language}} & \multicolumn{10}{c}{Canaries per \# of Repetitions}\\  \cmidrule(r){6-15}
        & & & & & 1 & 2 & 3 & 4 & 5 & 8 & 10 & 25 & 50 & 100 \\ \midrule
        
        % ================== BEGIN X-LARGE ==================
        
        % 200, 200, 200, 200, 200, 200, 100, 20, 20
        \makecell{X-Large\\
        (200e6+)} & sw, kaa, si, gu, kn, ne, uz, gl, mn, fil, mk, eu, ka, be, af, bn, te, is, mr, ml, hr, kk, ms, az, ta & 0.016\% & I,S,TP & 31500 & 600 & 600 & 600 & 600 & 600 & & 600 & 300 & 60 & 60 \\\thinrule
        
        % ================== BEGIN LARGE ==================
        
        \makecell{Large\\(20e6 to\\200e6)} & sw, kaa, si, gu, kn, ne, uz, gl, mn, fil, mk, eu, ka, be, af, bn, te, is, mr, ml, hr, kk, ms, az, ta & 0.035\% & I,S,TP & 7020 & 195 & 150 & & & 60 & & 60 & 45 & 30 & 30 \\\thinrule
        
        % ================== BEGIN MEDIUM ==================
        
        \makecell{Medium\\(6e6 to\\20e6)} & lo, dv, lb, fy, so, am, ps, zh, ku, km, pa, mt, tt, ga, tg, cy, ky, hy, my, eo & 0.02\% & I,S,TP & 1200 & 150 & 75 & & & 60 & & 30 & 12 & & \\\thinrule
         
        % ================== BEGIN SMALL ==================
        
        \makecell{Small\\(600e3\\to 6e6)} & ilo, os, cnh, ctd\_Latn, ti, udm, om, se, rm, tet, bo, ro, br, sa, lus, gsw, sah, kaa\_Latn, st, haw, pap, oc, cv, zu, sn, yo, as, sm, co, xh, ig, ny, kl, su, ceb, tk, fo, yi, hmn, el\_Latn, ba, jv, grc, or, sd, gd, ug, ckb, mg, ht, ha, rw & 0.02\% & I,S & 120 & 12 & 8 & & 8 & & & 3 & & & \\\thinrule
        
        % ================== BEGIN XSMALL ==================
        
        \makecell{XSmall-5\\(500e3\\to 600e3)} & hil & 0.02\% & I,S & 100 & 12 & 8 &  & 6 &  & 6 &  &  &  &  \\\verythinrule
        
        \makecell{XSmall-4\\(400e3\\to 500e3)} & te\_Latn, tyv, vec, kbd, lg & 0.02\% & I,S & 80 & 8 & 8 & & 6 & & 4 & & & & \\\verythinrule
        
        \makecell{XSmall-3\\(300e3\\to 400e3)} & iba, ak, av, ber\_Latn, zza, ts, ee, ru\_Latn & 0.02\% & I,S & 60 & 8 & 6 & & 6 & & 2 & & & & \\\verythinrule
        
        \makecell{XSmall-2\\(200e3\\to 300e3)} & ta\_Latn, cfm, otq, syr, bua, gn, to, az\_RU, wa, chm, tn, ada, krc, fj, nso & 0.02\% & I,S & 40 & 8 & 4 & & 2 & & 2 & & & & \\\verythinrule
        
        \makecell{XSmall-1\\(100e3\\to 200e3)} & emp, pck, meu, nnb, meo, nzi, tlh, tzo, iu, ksd, hui, tiv, sq, new, bci, bbc, min, sr, nan\_Latn\_TW, ve, ang, ml\_Latn, kac, ngu, pag, abt, kum, tyz, zap, kv, bik, nv, gom, ltg, qu, ay, rom, sg, ady, iso, yua, war, bho, hif, kbp, srn, myv, kha & 0.02\% & I,S & 20 & 4 & 4 & & 2 & & & & & & \\\thinrule
        
        % ================== BEGIN XXSMALL ==================
        
        \makecell{XXSmall-5\\(90,000\\to 100e3)}  & mrj, zxx\_xx\_dtynoise, kw, mgh, bew, crh, alt, nhe, fon & 0.01\% & I & 9 & 2 & 2 & 1 & & & & & & & \\\verythinrule
        
        \makecell{XXSmall-4\\(80,000\\to 90,000)} & mam, dov, ho, mai, bgp & 0.01\% & I & 8 & 3 & 1 & 1 & & & & & & & \\\verythinrule
        
        \makecell{XXSmall-3\\(70,000\\to 80,000)} & gym, rcf, shn, tvl, mbt, qub, pon & 0.01\% & I & 7 & 2 & 1 & 1 & & & & & & & \\\verythinrule
        
        \makecell{XXSmall-2\\(60,000\\to 70,000)} & ium, gag, tbz, gv, crs, quc, zh\_Latn, chk, btx, ace, bru, ubu, ape, mdf, tuc & 0.01\% & I & 6 & 2 & 2 & & & & & & & & \\\verythinrule
        
        \makecell{XXSmall-1\\(50,000\\to 60,000)} & tzh, kek, bum, bts, ibb, tcy, enq, kj, seh, xal, kmb, rwo, cab, wo, ppk, ach, kri, ss, cuk & 0.01\% & I & 5 & 3 & 1 & & & & & & & & \\
        \bottomrule
    % \end{tabular}
    % \label{tab:canary-distributions}
\end{longtable}
}

\subsubsection{Parallel Canaries}\label{ssec:canary-parallel}

Unlike \data, the parallel data consists of \texttt{source-target} ~pairs from different languages corresponding to the source and target languages for translation. This leads to new intricacies not present in the monolingual setting, e.g., because languages have different grammatical structure, it may be difficult to tailor modifications to both the inputs and outputs simultaneously that maintain linguistic structure. 

Rather than design canaries for all $N^2$ combinations of language pairs, where many pairs may have insufficient resource levels to incorporate canaries, 
we instead focus on the multiway setting where languages are grouped by the target language for translation. 
To minimize impact on the source languages (which may include very low-resource languages), we opt to not use any real training data from the source as inputs to the model. Instead, we generate canary data following one of two methodologies for the source: \texttt{random\_prefix} ~corresponds to cases where all canaries for a given target language share the same prefix of $100$ tokens but have unique uniformly random (w.r.t. the token vocabulary) suffixes following it and \texttt{full\_random} ~canaries are analogous with no shared prefix. The shared prefix of \texttt{random\_prefix} ~canaries is designed to align with cases where data may share common subsequences.
For the targets, we either \texttt{interleave} ~or \texttt{shuffle}  ~them as done in the \data case above (Appendix~\ref{ssec:canary-madlad}), except interleaving in batches of 50. Taking the outer product of these options, we get four possible canaries, e.g., \texttt{random\_prefix\_interleave} ~and so forth. The resource level groups and distribution of canaries is the same as for the \data canaries described in Table~\ref{tab:canary-distributions}; the mapping for parallel languages to these resource level groups is shown in Table~\ref{tab:parallel-canaries}. In total, $565,476$ canaries are generated this way across all languages.

Finally, we also design two additional types of canaries that use natural data from the source. Because this might impact utility more, we restrict these canaries to only the largest language pairs that have at least $500,000$ examples. The set of language pairs and languages satisfying this threshold are shown in Figure~\ref{fig:all-lang-pairs} and Figure~\ref{fig:unqiue-langs-from-pairs}. These two canary types are \texttt{interleaved\_both} ~and \texttt{interleaved\_mislabeled\_to}. The former performs the same interleaving operation on the source and targets for a language pair, interleaving in batches of 50 tokens. The latter does the same, with the addition of also select a new target language label, uniformly at random, from all qualifying high resource languages. For each language pair listed in Figure~\ref{fig:all-lang-pairs}, we generate 60 canaries in total, split evenly across the two canary types, and in the following distribution: 10 canaries are repeated once, 5 are repeated twice, and 2 are repeated 5 times. This gives a total of $21,120$ canaries in total across all language-pairs. Combined with the $565,476$ canaries from the prior 4 canary types, there are a total of $585,596$ canaries.

\begin{figure}
    \centering
    \begin{lstlisting}[frame=single,breakindent=0pt]
    ar-es, ar-fr, ar-ru, ar-zh, bg-cs, bg-da, bg-de, bg-el, bg-es, bg-fi, bg-fr, bg-hu, bg-it, bg-lt, bg-nl, bg-pl, bg-pt, bg-ro, bg-sk, bg-sl, bg-sv, cs-bg, cs-da, cs-de, cs-el, cs-es, cs-fi, cs-fr, cs-hu, cs-it, cs-lt, cs-nl, cs-pl, cs-pt, cs-ro, cs-ru, cs-sk, cs-sl, cs-sv, cs-zh, da-bg, da-cs, da-de, da-el, da-es, da-fi, da-fr, da-hu, da-it, da-lt, da-nl, da-pl, da-pt, da-ro, da-sk, da-sl, da-sv, da-zh, de-bg, de-cs, de-da, de-el, de-es, de-fi, de-fr, de-hu, de-it, de-lt, de-nl, de-pl, de-pt, de-ro, de-ru, de-sk, de-sl, de-sv, de-zh, el-bg, el-cs, el-da, el-de, el-es, el-fi, el-fr, el-hu, el-it, el-lt, el-nl, el-pl, el-pt, el-ro, el-sk, el-sl, el-sv, es-ar, es-bg, es-cs, es-da, es-de, es-el, es-fi, es-fr, es-hu, es-it, es-lt, es-nl, es-pl, es-pt, es-ro, es-ru, es-sk, es-sl, es-sv, es-zh, fi-bg, fi-cs, fi-da, fi-de, fi-el, fi-es, fi-fr, fi-hu, fi-it, fi-lt, fi-nl, fi-pl, fi-pt, fi-ro, fi-sk, fi-sl, fi-sv, fr-ar, fr-bg, fr-cs, fr-da, fr-de, fr-el, fr-es, fr-fi, fr-hu, fr-it, fr-lt, fr-nl, fr-pl, fr-pt, fr-ro, fr-ru, fr-sk, fr-sl, fr-sv, fr-zh, hu-bg, hu-cs, hu-da, hu-de, hu-el, hu-es, hu-fi, hu-fr, hu-it, hu-lt, hu-nl, hu-pl, hu-pt, hu-ro, hu-sk, hu-sl, hu-sv, it-bg, it-cs, it-da, it-de, it-el, it-es, it-fi, it-fr, it-hu, it-lt, it-nl, it-pl, it-pt, it-ro, it-ru, it-sk, it-sl, it-sv, it-zh, lt-bg, lt-cs, lt-da, lt-de, lt-el, lt-es, lt-fi, lt-fr, lt-hu, lt-it, lt-nl, lt-pl, lt-pt, lt-ro, lt-sk, lt-sl, lt-sv, nl-bg, nl-cs, nl-da, nl-de, nl-el, nl-es, nl-fi, nl-fr, nl-hu, nl-it, nl-lt, nl-pl, nl-pt, nl-ro, nl-ru, nl-sk, nl-sl, nl-sv, nl-zh, pl-bg, pl-cs, pl-da, pl-de, pl-el, pl-es, pl-fi, pl-fr, pl-hu, pl-it, pl-lt, pl-nl, pl-pt, pl-ro, pl-ru, pl-sk, pl-sl, pl-sv, pl-zh, pt-bg, pt-cs, pt-da, pt-de, pt-el, pt-es, pt-fi, pt-fr, pt-hu, pt-it, pt-lt, pt-nl, pt-pl, pt-ro, pt-ru, pt-sk, pt-sl, pt-sv, pt-zh, ro-bg, ro-cs, ro-da, ro-de, ro-el, ro-es, ro-fi, ro-fr, ro-hu, ro-it, ro-lt, ro-nl, ro-pl, ro-pt, ro-sk, ro-sl, ro-sv, ru-ar, ru-cs, ru-de, ru-es, ru-fr, ru-it, ru-nl, ru-pl, ru-pt, ru-zh, sk-bg, sk-cs, sk-da, sk-de, sk-el, sk-es, sk-fi, sk-fr, sk-hu, sk-it, sk-lt, sk-nl, sk-pl, sk-pt, sk-ro, sk-sl, sk-sv, sl-bg, sl-cs, sl-da, sl-de, sl-el, sl-es, sl-fi, sl-fr, sl-hu, sl-it, sl-lt, sl-nl, sl-pl, sl-pt, sl-ro, sl-sk, sl-sv, sv-bg, sv-cs, sv-da, sv-de, sv-el, sv-es, sv-fi, sv-fr, sv-hu, sv-it, sv-lt, sv-nl, sv-pl, sv-pt, sv-ro, sv-sk, sv-sl, sv-zh, zh-ar, zh-cs, zh-da, zh-de, zh-es, zh-fr, zh-it, zh-nl, zh-pl, zh-pt, zh-ru, zh-sv
\end{lstlisting}
    \caption{All language pairs satisfying the minimum $500,000$ example threshold.}
    \label{fig:all-lang-pairs}
\end{figure}

\begin{figure}
    \centering
    \begin{lstlisting}[frame=single,breakindent=0pt]
    da, cs, fi, el, ar, de, ru, fr, zh, nl, bg, lt, es, sv, pl, hu, pt, it, sk, ro, sl
\end{lstlisting}
    \caption{Unique languages from Figure~\ref{fig:all-lang-pairs}}
    \label{fig:unqiue-langs-from-pairs}
\end{figure}

{\fontsize{6.4}{8}\selectfont
\begin{longtable}{cm{6em}ccc}
        \caption{\label{tab:parallel-canaries}
         Distribution of (the $565,476$ in total) canaries across languages for parallel data. Distribution of canaries across repeats matches that in Table~\ref{tab:canary-distributions}. *Uses the codes: RPI=random\_prefix\_interleave, RPS=random\_prefix\_shuffle, FRI=fully\_random\_interleave, FRS=fully\_random\_shuffle. **includes 4 canary types instead of 3; so the canary distribution can be obtained by multiplying the values in Table~\ref{tab:canary-distributions} by $4/3$.}\\
    \toprule\\
        \multirow{2}{*}{\makecell{Dataset Size\\(\# Samples)}} & \multirow{2}{*}{\makecell{Languages\\Included}} & \multirow{2}{*}{\makecell{Target\\Canary Rate}} & \multirow{2}{*}{\makecell{Canary\\Types*}} & \multirow{2}{*}{\makecell{Canaries\\per Language}}\\\\ \midrule
        
        % ================== BEGIN X-LARGE ==================
        
        % 200, 200, 200, 200, 200, 200, 100, 20, 20
        \makecell{X-Large**\\
        (200e6+)} & nl, it, pt, fr, de, es, en & 0.016\% & I,S,TP & 42000 \\\thinrule
        
        % ================== BEGIN LARGE ==================
        
        \makecell{Large**\\(20e6 to\\200e6)} & fa, is, uk, tr, hr, ca, ko, ja, et, no, lv, sl, lt, bg, ar, el, fi, sk, ro, ru, hu, da, zh, pl, cs, sv & 0.035\% & RPI,RPS,FRS,FRI & 9360 \\\thinrule
        
        % ================== BEGIN MEDIUM ==================
        
        \makecell{Medium\\(6e6 to\\20e6)} & th, si, bn, sh, bs, hi, eo, mt, eu, sq, sr, gl, ga, mk, he, vi, id & 0.02\% & RPI,RPS,FRS & 1200 \\\thinrule
         
        % ================== BEGIN SMALL ==================
        
        \makecell{Small\\(600e3\\to 6e6)} & hy, ps, as, lb, uz, wa, my, iu, la, ha, pa, mg, xh, cy, ceb, gu, km, en\_xx\_simple, br, af, fy, tt, be, ka, oc, nds, tg, kk, az, nn, ne, sw, zh\_Hant, ur, fil, ms, te, mr, ta, ml & 0.02\% & RPI,RPS & 120 \\\thinrule
        
        % ================== BEGIN XSMALL ==================
        
        \makecell{XSmall-5\\(500e3\\to 600e3)} & dv, so, am, ba, se, sd & 0.02\% & RPI,RPS & 100 \\\verythinrule
        
        \makecell{XSmall-4\\(400e3\\to 500e3)} & kn, fo, or, fr\_CA, ku & 0.02\% & RPI,RPS & 80 \\\verythinrule
        
        \makecell{XSmall-3\\(300e3\\to 400e3)} & li, ug, arz, lmo, wuu, rw & 0.02\% & RPI,RPS & 60 \\\verythinrule
        
        \makecell{XSmall-2\\(200e3\\to 300e3)} & ltg, gn, bug, ks\_Deva, szl, fuv, ary, ban, bm, ace, bjn\_Arab, taq, ace\_Arab, bjn, prs, taq\_Tfng, tzm, lij, hne, vec, din, kr\_Arab, sc, ks, fur, shn, bho, mag, nus, mi, mni, dz, scn, an & 0.02\% & RPI,RPS & 40 \\\verythinrule
        
        \makecell{XSmall-1\\(100e3\\to 200e3)} & nds\_NL, ky, zu, mn, jv, bar, gd, kr, crh\_Latn & 0.02\% & RPI,RPS & 20 \\\thinrule
        
        % ================== BEGIN XXSMALL ==================
        
        \makecell{XXSmall-5\\(90,000\\to 100e3)}  & mwl & 0.01\% & RPI & 9 \\\verythinrule
        
        \makecell{XXSmall-3\\(70,000\\to 80,000)} & io & 0.01\% & RPI & 7 \\\verythinrule

    % \end{tabular}
    % \label{tab:canary-distributions}
\end{longtable}
}

\subsubsection{Generic Canaries}\label{ssec:generic-canaries}
Finally, we also designed and generated $80,400$ \texttt{generic} ~canaries. Unlike the canaries of the prior two sections (\ref{ssec:canary-madlad} and~\ref{ssec:canary-parallel}), these canaries share minimal resemblance to natural data. These canaries may be useful for understanding memorization of highly outlier data. Here, we generate monolingual canaries where the source and targets are the same. We propose two types of canaries: \texttt{random\_prefix} ~and \texttt{fully\_random} canaries. These are the same as the canaries described in  Section~\ref{ssec:canary-parallel} but with the source matching the target. We generated $80,400$ canaries in total split evenly among 4 types of canaries: \texttt{fully\_random} ~canaries and \texttt{random\_prefix} ~canaries with shared prefixes of length 50, 100, and 200.

\pagebreak

\subsection{Additional Memorization Figures}\label{sec:memorization-details}

\begin{figure}[H]
    \centering
    \begin{subfigure}{0.45\textwidth}
    \centering
        \includegraphics[width=\linewidth]{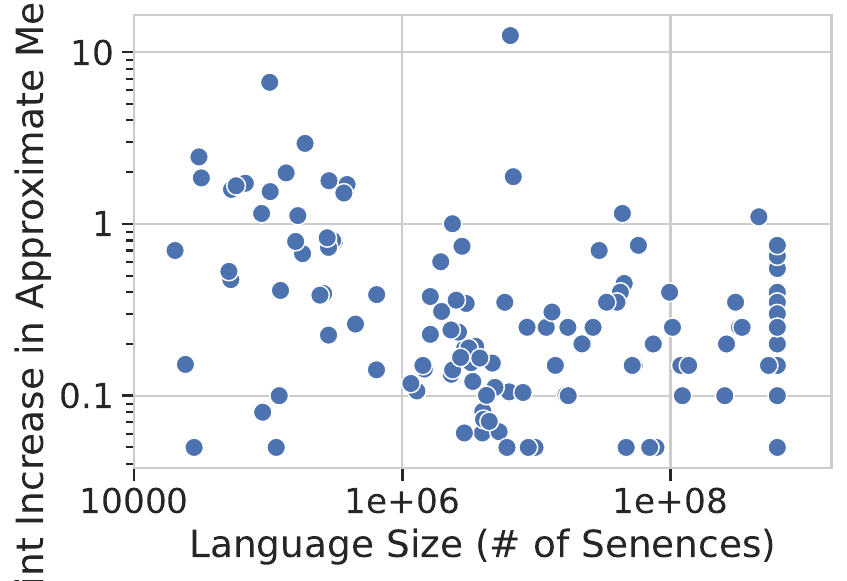}
        \label{fig:mono_approx_mem}
    \end{subfigure}
    \begin{subfigure}{0.45\textwidth}
    \centering
        \includegraphics[width=\linewidth]{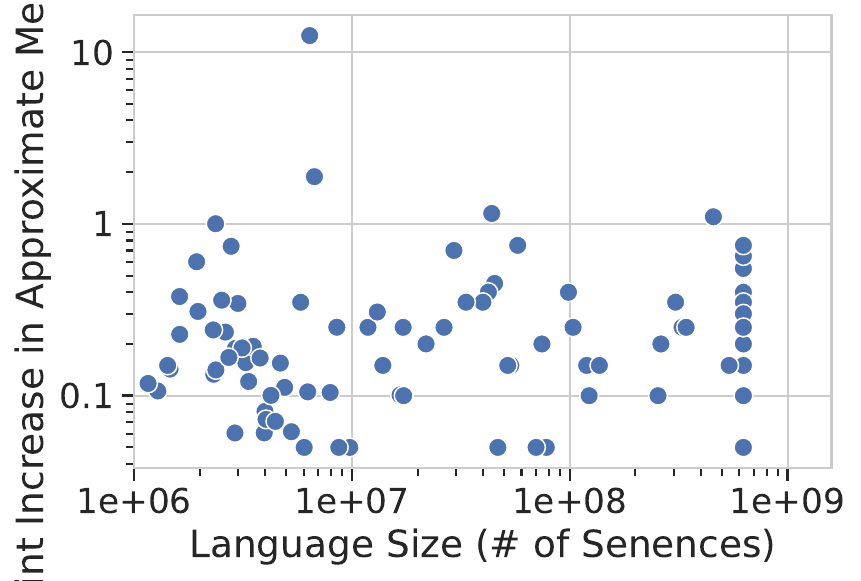}
        \label{fig:multi_approx_mem}
    \end{subfigure}
    \caption{\textbf{Translate models may also paraphrase memorizations.} Training data extraction rates under approimate memorization for both \textbf{(left)} monolingual (translate\_copy) and \textbf{(right)} multiway (\texttt{translate\_diff}) data. Extraction performed on the 3B parameter model using a $S=P+50$. We use a Levenshtein similarity of 90\% on the additioanl leakage. After accounting for this, a fewer 208/370 and 119/146 had no memorization detected.}
    \label{fig:approx_mem}
\end{figure}

\subsection{Datasheet}

% \fbox{\begin{minipage}{40em}

\subsubsection*{Datasheet - \data}

\small

\paragraph{Motivation}

\begin{enumerate}[leftmargin=*]
    \item For what purpose was the dataset created? (Was there a specific task in mind? Was there a specific gap that needed to be filled? Please provide a description.)
    \textit{We create \data as a general purpose monolingual document level dataset covering 419 languages for the purpose of providing general training data for multilingual NLP tasks such as MT and language modeling. One of the goals of the Language Inclusivity Moonshot (LIM) \footnote{\url{https://blog.google/technology/ai/ways-ai-is-scaling-helpful/}} is to scale our language support to 1,000 languages and to support speech recognition for ~97\% of the global population. A core objective associated with this goal  is to open source data and models. Our expectation is that releasng \data will foster progress on the language research, especially on medium and low resource languages. An estimate of over 1B people globally speak languages that are not covered by mainstream models at Google or externally. }
    \item Who created this dataset and on behalf of which entity? \textit{Sneha Kudugunta$^\dagger$, Isaac Caswell$^\diamond$, Biao Zhang$^\dagger$, Xavier Garcia$^\dagger$, Derrick Xin$^\dagger$, Aditya Kusupati$^\diamond$, Romi Stella$^\dagger$, Ankur Bapna$^\dagger$, Orhan Firat$^\dagger$ 
($^\dagger$Google DeepMind, 
$^\diamond$Google Research})
    \item Who funded the creation of the dataset? \textit{Google Research and Google DeepMind}
    \item Any other comments? \textit{None}
\end{enumerate}

\paragraph{Composition}

\begin{enumerate}[leftmargin=*]
    \item What do the instances that comprise the dataset represent? \textit{Each instance is a preprocessed web-crawled document whose language that we annotated using a LangID model described by \citep{caswell2020language}. For the sentence level version, we used a sentence-splitter to split the documents into sentences and then deduplicated the resulting dataset.}
    \item How many instances are there in total? \textit{\data has 4.0B documents (100B sentences, or 2.8T tokens) total across 419 languages with the median language containing 1.7k documents (73k sentences of 1.2M tokens.)}
    \item Does the dataset contain all possible instances or is it a sample (not necessarily random) of instances from a larger set? (If the dataset is a sample, then what is the larger set? Is the sample representative of the larger set (e.g., geographic coverage)? If so, please describe how this representativeness was validated/verified. If it is not representative of the larger set, please describe why not (e.g., to cover a more diverse range of instances, because instances were withheld or unavailable).) \textit{\data are created from CommonCrawl documents that have been annotated by language, filtered and preprocessed. To maintain high precision, we filtered out data aggressively, and may not have captured every document of a given language in CommonCrawl. Moreover, there may also be languages in CommonCrawl that we may not have mined.}
    \item What data does each instance consist of? \textit{Each instance is raw text in either document form for the document level data, or in sentence form for the sentence level data.}
    \item Is there a label or target associated with each instance? If so, please provide a description. \textit{No.}
    \item Is any information missing from individual instances? \textit{No.}
    \item Are relationships between individual instances made explicit? \textit{No.}
    \item Are there recommended data splits (e.g., training, development/validation, testing)? \textit{No.}
    \item Are there any errors, sources of noise, or redundancies in the dataset? \textit{While we have taken extensive care to audit and filter \data, there may still be documents annotated with the wrong language or documents of low quality.}
    \item Is the dataset self-contained, or does it link to or otherwise rely on external resources (e.g., websites, tweets, other datasets)? (If it links to or relies on external resources, a) are there guarantees that they will exist, and remain constant, over time; b) are there official archival versions of the complete dataset (i.e., including the external resources as they existed at the time the dataset was created); c) are there any restrictions (e.g., licenses, fees) associated with any of the external resources that might apply to a future user? Please provide descriptions of all external resources and any restrictions associated with them, as well as links or other access points, as appropriate.) \textit{Yes}
    \item Does the dataset contain data that might be considered confidential (e.g., data that is protected by legal privilege or by doctor-patient confidentiality, data that includes the content of individuals' non-public communications)? (If so, please provide a description.) \textit{Given that \data is a general web-crawled dataset it is possible that documents in the dataset may contain such information.}
    \item Does the dataset contain data that, if viewed directly, might be offensive, insulting, threatening, or might otherwise cause anxiety? (If so, please describe why.) \textit{Given that \data is a general web-crawled dataset, even after filtering, it is possible that there are documents containing offensive content, etc.}
    \item Does the dataset relate to people? \textit{It is likely that some documents in \data contain sentences referring to and describing people.}
    \item Does the dataset identify any subpopulations (e.g., by age, gender)?  \textit{It is likely that some documents in \data contain sentences referring to and describing people of certain subpopulations.}
    \item Is it possible to identify individuals (i.e., one or more natural persons), either directly or indirectly (i.e., in combination with other data) from the dataset? \textit{Yes, it is possible that their names are mentioned in certain documents.}
    \item Does the dataset contain data that might be considered sensitive in any way (e.g., data that reveals racial or ethnic origins, sexual orientations, religious beliefs, political opinions or union memberships, or locations; financial or health data; biometric or genetic data; forms of government identification, such as social security numbers; criminal history)? \textit{Given that \data is a general web-crawled dataset, even after filtering, it is possible that there are documents containing sensitive data.}
    \item Any other comments? \textit{None.}
\end{enumerate}

\paragraph{Collection}

\begin{enumerate}[leftmargin=*]
    \item How was the data associated with each instance acquired? \textit{Each instance was acquired by performing transformations on the documents in all available snapshots of CommonCrawl as of August 20, 2022.}
    \item What mechanisms or procedures were used to collect the data (e.g., hardware apparatus or sensor, manual human curation, software program, software API)? \textit{We annotated the CommonCrawl data using a LangID model trained using the procedure described by \citep{caswell2020language}. Then, we manually inspected the data and then filtered or preprocessed the documents to create \data.}
    \item If the dataset is a sample from a larger set, what was the sampling strategy? \textit{\data is a subset of CommonCrawl documents determined using LangID annotations and filtering/preprocessing steps.}
    \item Who was involved in the data collection process (e.g., students, crowdworkers, contractors) and how were they compensated (e.g., how much were crowdworkers paid)? \textit{For the audit, the authors inspected the dataset. In some cases native speaker volunteers provided advice on the quality of the dataset.}
    \item Over what timeframe was the data collected? (Does this timeframe match the creation timeframe of the data associated with the instances (e.g., recent crawl of old news articles)? If not, please describe the timeframe in which the data associated with the instances was created.) \textit{We do not annotate timestamps. The version of CommonCrawl that we used has webcrawls ranging from 2008 to August 2022.}
    \item Were any ethical review processes conducted (e.g., by an institutional review board)? \textit{No.}
    \item Does the dataset relate to people? \textit{It is likely that some documents in \data contain sentences referring to and describing people.}
    \item Did you collect the data from the individuals in question directly, or obtain it via third parties or other sources (e.g., websites)? \textit{We collected this data via webpages crawled by CommonCrawl.}
    \item Were the individuals in question notified about the data collection? \textit{No.}
    \item Did the individuals in question consent to the collection and use of their data? \textit{No.}
    \item If consent was obtained, were the consenting individuals provided with a mechanism to revoke their consent in the future or for certain uses? \textit{No.}
    \item Has an analysis of the potential impact of the dataset and its use on data subjects (e.g., a data protection impact analysis) been conducted? \textit{No.}
    \item Any other comments? \textit{None.}
\end{enumerate}

\paragraph{Preprocessing/cleaning/labeling}

\begin{enumerate}[leftmargin=*]
    \item Was any preprocessing/cleaning/labeling of the data done (e.g., discretization or bucketing, tokenization, part-of-speech tagging, SIFT feature extraction, removal of instances, processing of missing values)? (If so, please provide a description. If not, you may skip the remainder of the questions in this section.) \textit{Various types of preprocessing were done: deduplication of 3 sentence spans, filtering substrings according to various heuristics associated with low quality, Virama encoding correction, converting Zawgyi encoding to Unicode encoding for Myanmar script characters and a Chinese pornographic content filter heuristic. In addition, 79 annotated language datasets were removed on inspection due to low quality or mislabeling.
}
    \item Was the "raw" data saved in addition to the preprocessed/cleaned/labeled data (e.g., to support unanticipated future uses)? \textit{No, this raw data is hosted by CommonCrawl.}
    \item Is the software used to preprocess/clean/label the instances available? \textit{As of June 13, 2023, no.}
    \item Any other comments? \textit{None.}
\end{enumerate}

\paragraph{Uses}

\begin{enumerate}[leftmargin=*]
    \item Has the dataset been used for any tasks already? (If so, please provide a description.) \textit{\data has been used for MT and language modeling.}
    \item Is there a repository that links to any or all papers or systems that use the dataset? \textit{No}
    \item What (other) tasks could the dataset be used for? \textit{This dataset could be used as a general training dataset for any of the languages in \data.}
    \item Is there anything about the composition of the dataset or the way it was collected and preprocessed/cleaned/labeled that might impact future uses? (For example, is there anything that a future user might need to know to avoid uses that could result in unfair treatment of individuals or groups (e.g., stereotyping, quality of service issues) or other undesirable harms (e.g., financial harms, legal risks) If so, please provide a description. Is there anything a future user could do to mitigate these undesirable harms?) \textit{While steps have been taken to clean \data, content containing sensitive content about individuals or groups could affect the performance of some downstream NLP tasks. Moreover, while building applications for (a) given language(s), we urge practitioners to assess the suitability of \data for their usecase.}
    \item Are there tasks for which the dataset should not be used? (If so, please provide a description.) \textit{N/A.}
    \item Any other comments? \textit{None.}
\end{enumerate}

\paragraph{Distribution}

\begin{enumerate}[leftmargin=*]
    \item How will the dataset will be distributed (e.g., tarball on website, API, GitHub)? \textit{\data is made available through a GCP bucket.}
    \item When will the dataset be distributed? \textit{June 2023}
    \item Will the dataset be distributed under a copyright or other intellectual property (IP) license, and/or under applicable terms of use (ToU)? (If so, please describe this license and/or ToU, and provide a link or other access point to, or otherwise reproduce, any relevant licensing terms or ToU, as well as any fees associated with these restrictions.) \textit{AI2 has made a version of this data available under the ODC-BY license. Users are also bound by the CommonCrawl terms of use in respect of the content contained in the dataset.}
    \item Have any third parties imposed IP-based or other restrictions on the data associated with the instances? \textit{Users are bound by the CommonCrawl terms of use in respect of the content contained in the dataset.}
    \item Any other comments? \textit{None.}
\end{enumerate}

\paragraph{Maintenance}

\begin{enumerate}[leftmargin=*]
    \item Who is supporting/hosting/maintaining the dataset? \textit{An external organization, AI2 is hosting the dataset.}
    \item How can the owner/curator/manager of the dataset be contacted (e.g., email address)? \textit{Sneha Kudugunta \texttt{snehakudugunta@google.com} or Isaac Caswell (\texttt{icaswell@google.com}) for questions about the dataset contents, or Dirk Groeneveld \texttt{dirkg@allenai.org} for questions related to the hosting of the dataset.}
    \item Is there an erratum? (If so, please provide a link or other access point.) \url{https://github.com/google-research/google-research/tree/master/madlad_400}
    \item Will the dataset be updated (e.g., to correct labeling errors, add new instances, delete instances')? (If so, please describe how often, by whom, and how updates will be communicated to users (e.g., mailing list, GitHub)?) \textit{There are no such plans, but major issues may be corrected when reported through email or the Github page (\url{https://github.com/google-research/google-research/tree/master/madlad_400}).}
    \item If the dataset relates to people, are there applicable limits on the retention of the data associated with the instances (e.g., were individuals in question told that their data would be retained for a fixed period of time and then deleted)? (If so, please describe these limits and explain how they will be enforced.) \textit{N/A}
    \item Will older versions of the dataset continue to be supported/hosted/maintained? (If so, please describe how. If not, please describe how its obsolescence will be communicated to users.) \textit{No}
    \item If others want to extend/augment/build on/contribute to the dataset, is there a mechanism for them to do so? (If so, please provide a description. Will these contributions be validated/verified? If so, please describe how. If not, why not? Is there a process for communicating/distributing these contributions to other users? If so, please provide a description.) \textit{A relatively unprocessed version of \data, \data-\texttt{noisy} is made available for others to build upon using superior cleaning/preprocessing techniques for their specific usecases.}
    \item Any other comments? \textit{None}
\end{enumerate}

% \end{minipage}}

% \newpage

\subsection{Model Card}

\fbox{\begin{minipage}{43em}

\subsubsection*{Model Card}

\small

\paragraph{Model Details}

\begin{itemize}[leftmargin=*]
    \item Person or organization developing model: \textit{Google DeepMind and Google Research}
    \item Model Date: \textit{June 13, 2023}
    \item Model Types: \textit{Machine Translation and Language Modeling Models.}
    \item Information about training algorithms, parameters, fairness constraints or other applied approaches, and features: \textit{Provided in the paper.}
    \item Paper: \textit{Kudugunta et al, \data: Monolingual And Document-Level Large Audited Dataset, Under Review, 2023}
    \item License: \textit{ODC-BY}
    \item Contact: \textit{snehakudugunta@google.com}
\end{itemize}

\paragraph{Intended Use} 

\begin{itemize}[leftmargin=*]
    \item Primary intended uses: \textit{Machine Translation and multilingual NLP tasks on over 400 languages.}
    \item Primary intended users: \textit{Research community.}
    \item Out-of-scope use cases: \textit{These models are trained on general domain data and are therefore not meant to work on domain-specific models out-of-the box. Moreover, these research models have not been assessed for production usecases.}
\end{itemize}

\paragraph{Factors} 

\begin{itemize}[leftmargin=*]
    \item \textit{The translation quality of this model varies based on language, as seen in the paper, and likely varies on domain, though we have not assessed this.}
\end{itemize}

\paragraph{Metrics}

\begin{itemize}[leftmargin=*]
    \item \textit{We use SacreBLEU and chrF, two widely used machine translation evaluation metrics for our evaluations.}
\end{itemize}

\paragraph{Ethical Considerations} 

\begin{itemize}[leftmargin=*]
    \item \textit{We trained these models with \data and publicly available data to create baseline models that support NLP for over 400 languages, with a focus on languages underrepresented in large-scale corpora. Given that these models were trained with web-crawled datasets that may contain sensitive, offensive or otherwise low-quality content despite extensive preprocessing, it is still possible that these issues to the underlying training data may cause differences in model performance and toxic (or otherwise problematic) output for certain domains. Moreover, large models are dual use technologies that have specific risks associated with their use and development. We point the reader to surveys such as those written by \citet{weidinger2021ethical} or \citet{bommasani2021opportunities} for a more detailed discussion of these risks, and to \citet{liebling2022opportunities} for a thorough discussion of the risks of machine translation systems. }
    % \item \textit{focused on low resource languages}
    % \item  \textit{web-crawled dataset have issues that may be reflected in model performance; not assessed}
    % \item \textit{Harms of LMs and possible mitigations. Difficulty for models}
    % \item \textit{MT ethical issues}
\end{itemize}

% Innovation in NLP technologies in English has been accelerated by training large scale deep learning models~\citep{devlin2018bert,brown2020language} on massive web corpora~\citep{chelba2013one,zhu2015aligning,raffel2020exploring}. However, on the long tail of written languages in the world there is a lack of high quality general data sources~\citep{joshi2020state} that impede the progress of NLP tools for other languages. We hope that making an audited and cleaned corpus such as \data available mitigates this issue. While we extensively cleaned \data, the extent to which we can preprocess this data is limited by how not all languages have available tools for removing problematic content such as porn, toxic content, PII, copyrighted content or noise. We urge practitioners to carefully consider their target usecase before using \data.

\paragraph{Training Data} 

\begin{itemize}[leftmargin=*]
    \item \textit{For both the machine translation and language model, \data is used. For the machine translation model, a combination of parallel datasources covering 157 languages is also used. Further details are described in the paper.}
\end{itemize}

\paragraph{Evaluation Data} 

\begin{itemize}[leftmargin=*]
    \item \textit{For evaluation, we used WMT, NTREX, Flores-200 and Gatones datasets as described in Section \ref{sub:eval}.}
\end{itemize}

\paragraph{Caveats and Recommendations}

\begin{itemize}[leftmargin=*]
    \item \textit{We note that we evaluate on only 204 of the languages supported by these models and on machine translation and few-shot machine translation tasks. Users must consider use of this model carefully for their own usecase.}
\end{itemize}

\end{minipage}}

\newpage

\subsubsection*{Canaries Datasheet}

\small

\paragraph{Motivation}

\begin{enumerate}[leftmargin=*]
    \item For what purpose was the dataset created? (Was there a specific task in mind? Was there a specific gap that needed to be filled? Please provide a description.)
    \textit{We create these canaries with the goal of enabling the study of memorization in the multilingual and translate settings. Models can be trained on these canaries and then their risk of memorization assessed.}
    \item Who created this dataset and on behalf of which entity? \textit{Christopher A. Choquette-Choo$^\dagger$, Katherine Lee$^\dagger$
($^\dagger$Google DeepMind})
    \item Who funded the creation of the dataset? \textit{Google DeepMind}
    \item Any other comments? \textit{None}
\end{enumerate}

\paragraph{Composition}

\begin{enumerate}[leftmargin=*]
    \item What do the instances that comprise the dataset represent? \textit{Each instance is constructed from a \data sentence-level example. Careful modifications are used, e.g., shuffling the tokens, to make the sample outlier to the natural distribution.}
    \item How many instances are there in total? \textit{There are $1,945,631$ in total.}
    \item Does the dataset contain all possible instances or is it a sample (not necessarily random) of instances from a larger set? (If the dataset is a sample, then what is the larger set? Is the sample representative of the larger set (e.g., geographic coverage)? If so, please describe how this representativeness was validated/verified. If it is not representative of the larger set, please describe why not (e.g., to cover a more diverse range of instances, because instances were withheld or unavailable).) \textit{The canary dataset itself is a subsampling from \data. However, all described canaries are included in the release.}
    \item What data does each instance consist of? \textit{Each instance consists of the original text as well as the modified instance in tokens.}
    \item Is there a label or target associated with each instance? If so, please provide a description. \textit{No.}
    \item Is any information missing from individual instances? \textit{No.}
    \item Are relationships between individual instances made explicit? \textit{No.}
    \item Are there recommended data splits (e.g., training, development/validation, testing)? \textit{No.}
    \item Are there any errors, sources of noise, or redundancies in the dataset? \textit{Some canaries are duplicated for the purposes of studying repetition in memorization.}
    \item Is the dataset self-contained, or does it link to or otherwise rely on external resources (e.g., websites, tweets, other datasets)? (If it links to or relies on external resources, a) are there guarantees that they will exist, and remain constant, over time; b) are there official archival versions of the complete dataset (i.e., including the external resources as they existed at the time the dataset was created); c) are there any restrictions (e.g., licenses, fees) associated with any of the external resources that might apply to a future user? Please provide descriptions of all external resources and any restrictions associated with them, as well as links or other access points, as appropriate.) \textit{Yes}
    \item Does the dataset contain data that might be considered confidential (e.g., data that is protected by legal privilege or by doctor-patient confidentiality, data that includes the content of individuals' non-public communications)? (If so, please provide a description.) \textit{This may be possible given the underlying \data may contain such data. However, the modifications used for generating canaries reduce the chance of this.}
    \item Does the dataset contain data that, if viewed directly, might be offensive, insulting, threatening, or might otherwise cause anxiety? (If so, please describe why.) \textit{This may be possible given the underlying \data may contain such data. However, the modifications used for generating canaries reduce the chance of this.}
    \item Does the dataset relate to people? \textit{This may be possible given the underlying \data may contain such data. However, the modifications used for generating canaries reduce the chance of this.}
    \item Does the dataset identify any subpopulations (e.g., by age, gender)?  \textit{This may be possible given the underlying \data may contain such data. However, the modifications used for generating canaries reduce the chance of this.}
    \item Is it possible to identify individuals (i.e., one or more natural persons), either directly or indirectly (i.e., in combination with other data) from the dataset? \textit{This may be possible given the underlying \data may contain such data. However, the modifications used for generating canaries reduce the chance of this.}
    \item Does the dataset contain data that might be considered sensitive in any way (e.g., data that reveals racial or ethnic origins, sexual orientations, religious beliefs, political opinions or union memberships, or locations; financial or health data; biometric or genetic data; forms of government identification, such as social security numbers; criminal history)? \textit{This may be possible given the underlying \data may contain such data. However, the modifications used for generating canaries reduce the chance of this.}
    \item Any other comments? \textit{None.}
\end{enumerate}

\paragraph{Collection}

\begin{enumerate}[leftmargin=*]
    \item How was the data associated with each instance acquired? \textit{Each instance was acquired by performing transformations on the documents in all available snapshots of CommonCrawl as of August 20, 2022.}
    \item What mechanisms or procedures were used to collect the data (e.g., hardware apparatus or sensor, manual human curation, software program, software API)? \textit{We randomly subsampled \data and then applied random }
    \item If the dataset is a sample from a larger set, what was the sampling strategy? \textit{A predefined number of sentences were uniformly sampled from each language.}
    \item Who was involved in the data collection process (e.g., students, crowdworkers, contractors) and how were they compensated (e.g., how much were crowdworkers paid)? \textit{The authors created the canary dataset.}
    \item Over what timeframe was the data collected? (Does this timeframe match the creation timeframe of the data associated with the instances (e.g., recent crawl of old news articles)? If not, please describe the timeframe in which the data associated with the instances was created.) \textit{We do not annotate timestamps.}
    \item Were any ethical review processes conducted (e.g., by an institutional review board)? \textit{No.}
    \item Does the dataset relate to people? \textit{This may be possible given the underlying \data may contain such data. However, the modifications used for generating canaries reduce the chance of this.}
    \item Did you collect the data from the individuals in question directly, or obtain it via third parties or other sources (e.g., websites)? \textit{Obtained second-hand through \data, a CommonCrawl based dataset.}
    \item Were the individuals in question notified about the data collection? \textit{No.}
    \item Did the individuals in question consent to the collection and use of their data? \textit{No.}
    \item If consent was obtained, were the consenting individuals provided with a mechanism to revoke their consent in the future or for certain uses? \textit{No.}
    \item Has an analysis of the potential impact of the dataset and its use on data subjects (e.g., a data protection impact analysis) been conducted? \textit{No.}
    \item Any other comments? \textit{None.}
\end{enumerate}

\paragraph{Preprocessing/cleaning/labeling}

\begin{enumerate}[leftmargin=*]
    \item Was any preprocessing/cleaning/labeling of the data done (e.g., discretization or bucketing, tokenization, part-of-speech tagging, SIFT feature extraction, removal of instances, processing of missing values)? (If so, please provide a description. If not, you may skip the remainder of the questions in this section.) \textit{Various types of processing were applied on top of \data. Sentences were either interleaved in batches of $20-50$ tokens or shuffled at the token-level.
}
    \item Was the "raw" data saved in addition to the preprocessed/cleaned/labeled data (e.g., to support unanticipated future uses)? \textit{The raw \data data is saved.}
    \item Is the software used to preprocess/clean/label the instances available? \textit{As of June 13, 2023, no.}
    \item Any other comments? \textit{None.}
\end{enumerate}

\paragraph{Uses}

\begin{enumerate}[leftmargin=*]
    \item Has the dataset been used for any tasks already? (If so, please provide a description.) \textit{\data has been used for MT.}
    \item Is there a repository that links to any or all papers or systems that use the dataset? \textit{No}
    \item What (other) tasks could the dataset be used for? \textit{This dataset can be used to study memorization in broad language modelling scenarios.}
    \item Is there anything about the composition of the dataset or the way it was collected and preprocessed/cleaned/labeled that might impact future uses? (For example, is there anything that a future user might need to know to avoid uses that could result in unfair treatment of individuals or groups (e.g., stereotyping, quality of service issues) or other undesirable harms (e.g., financial harms, legal risks) If so, please provide a description. Is there anything a future user could do to mitigate these undesirable harms?) \textit{We urge users to read the datasheet for \data to understand the underlying risk for the canaries.}
    \item Are there tasks for which the dataset should not be used? (If so, please provide a description.) \textit{N/A.}
    \item Any other comments? \textit{None.}
\end{enumerate}

\paragraph{Distribution}

\begin{enumerate}[leftmargin=*]
    \item How will the dataset will be distributed (e.g., tarball on website, API, GitHub)? \textit{\data is made available through a GCP bucket.}
    \item When will the dataset be distributed? \textit{June 2023}
    \item Will the dataset be distributed under a copyright or other intellectual property (IP) license, and/or under applicable terms of use (ToU)? (If so, please describe this license and/or ToU, and provide a link or other access point to, or otherwise reproduce, any relevant licensing terms or ToU, as well as any fees associated with these restrictions.) \textit{AI2 has made a version of this data available under the ODC-BY license. Users are also bound by the CommonCrawl terms of use in respect of the content contained in the dataset.}
    \item Have any third parties imposed IP-based or other restrictions on the data associated with the instances? \textit{Users are bound by the CommonCrawl terms of use in respect of the content contained in the dataset.}
    \item Any other comments? \textit{None.}
\end{enumerate}

\paragraph{Maintenance}

\begin{enumerate}[leftmargin=*]
    \item Who is supporting/hosting/maintaining the dataset? \textit{An external organization, AI2 is hosting the dataset.}
    \item How can the owner/curator/manager of the dataset be contacted (e.g., email address)? \textit{Christopher A. Choquette-Choo \texttt{cchoquette@google.com} for questions about the dataset contents or Dirk Groeneveld \texttt{dirkg@allenai.org} for questions related to the hosting of the dataset.}
    \item Is there an erratum? (If so, please provide a link or other access point.) \textit{No}
    \item Will the dataset be updated (e.g., to correct labeling errors, add new instances, delete instances')? (If so, please describe how often, by whom, and how updates will be communicated to users (e.g., mailing list, GitHub)?) \textit{There are no such plans, but major issues may be corrected when reported through email or the Github page.}
    \item If the dataset relates to people, are there applicable limits on the retention of the data associated with the instances (e.g., were individuals in question told that their data would be retained for a fixed period of time and then deleted)? (If so, please describe these limits and explain how they will be enforced.) \textit{N/A}
    \item Will older versions of the dataset continue to be supported/hosted/maintained? (If so, please describe how. If not, please describe how its obsolescence will be communicated to users.) \textit{No}
    \item If others want to extend/augment/build on/contribute to the dataset, is there a mechanism for them to do so? (If so, please provide a description. Will these contributions be validated/verified? If so, please describe how. If not, why not? Is there a process for communicating/distributing these contributions to other users? If so, please provide a description.) \textit{Others may build upon this by similarly generating canaries from \data which is made available.}
    \item Any other comments? \textit{None}
\end{enumerate}

\end{document}